%% file: main.tex
\documentclass[12pt,a4paper, oneside]{book}
\include{misc/preamble}

\include{misc/glossary}

\begin{document}
	\pagenumbering{gobble}
	\pagestyle{plain}

	\include{misc/frontpage}

	\frontmatter
	\include{misc/dedication}

	\include{misc/acknowledgement}

	\input{chapters/abstract.tex}
	\tableofcontents
	\clearpage
	\listoffigures
	\clearpage
	\listoftables
	\clearpage
	\printglossary[type=\acronymtype,nonumberlist]
	\printglossary
	\input{misc/epigraph}
	\mainmatter
	\pagestyle{fancy}
	\include{chapters/introduction}

	\input{chapters/chapter01}
	\input{chapters/chapter02}
	\input{chapters/chapter03}
	\input{chapters/chapter04}
	\input{chapters/chapter05}
	\input{chapters/conclusion}

	\newpage

	\bibliography{misc/main} 
	\bibliographystyle{ieeetr}

\end{document}

%% file: misc/preamble.tex
\usepackage[top=1in, bottom=1in, right=1in, left=1.2in]{geometry}
\usepackage[acronym,automake]{glossaries}
\usepackage{epigraph}
\usepackage{fancyhdr}
\usepackage{hyperref}
\hypersetup{colorlinks=true,citecolor=red,linkcolor=black}
\usepackage[Lenny]{fncychap}
\usepackage{pgfplots}
\usepackage{wrapfig}
\usepackage[center]{caption}
\usepackage{subcaption}
\usepackage{makecell}
\usepackage{enumitem}
\usepackage{verbatim}
\usepackage{float}
\usepackage{graphicx}
\usepackage{adjustbox}
\usepackage{tabu}
\usepackage{tikz}
\usepackage[graphicx]{realboxes}
\usepackage{pgfornament}

\usepgfplotslibrary{groupplots,dateplot}
\pgfplotsset{compat=newest}
\usetikzlibrary{positioning, patterns,calc, arrows}

\definecolor{navyblue}{rgb}{0.0, 0.0, 0.5}

\newcommand{\chapterintrobox}[1]{\begin{center}\fbox{\begin{minipage}[t]{0.9\linewidth}\centering\vspace{0.04\linewidth}\begin{minipage}[t]{0.86\linewidth} #1 \vspace{0.04\linewidth}\end{minipage}\end{minipage}}\vspace{1cm}\end{center}}

\usetikzlibrary{positioning,arrows,calc}
\def\nodedist{35pt}
\def\layerdist{80pt}

%% file: misc/glossary.tex
\makeglossaries
\newacronym{pm}{PM}{Preventive Maintenance}
\newacronym{cbm}{CBM}{Condition-Based Maintenance}
\newacronym{phm}{PHM}{Prognostics and Health Management}
\newacronym{lcm}{LCM}{Life Cycle Management}
\newacronym{cm}{CM}{Condition Monitoring}
\newacronym{rul}{RUL}{Remaining Useful Life}
\newacronym{dd}{DD}{Data-Driven}

\newacronym{rnn}{RNN}{Recurrent Neural Networks}
\newacronym{cnn}{CNN}{Convolutional Neural Networks}
\newacronym{lstm}{LSTM}{Long Short-Term Memory}
\newacronym{roc}{ROC}{Receiver Operating Characteristics}
\newacronym{auc}{AUC}{Area Under Curve}
\newacronym{dft}{DFT}{Discrete Fourier Transform}
\newacronym{fft}{FFT}{Fast Fourier Transform}
\newacronym{cwt}{CWT}{Continuous Wavelet Transform}
\newacronym{dwt}{DWT}{Discrete Wavelet Transform}
\makeglossaries

%% file: misc/frontpage.tex
\begin{tikzpicture}[overlay,remember picture]
    \draw [line width=5pt, color=navyblue]
        ($ (current page.north west) + (1cm,-1cm) $)
        rectangle
        ($ (current page.south east) + (-1cm,1cm) $);
\end{tikzpicture}
{\small N$^o$ d'ordre: \ldots\ldots\ldots/Faculté/UMBB/2020}
\begin{center}
	{\large People's Democratic Republic of Algeria}

	{\large Ministry of Higher Education and Scientific Research}

	\vspace{.6cm}
\textbf{University of M'hamed Bougara}\\ 
\textbf{Faculty of Hydrocarbons \& Chemistry} 

\includegraphics[width=3.5cm]{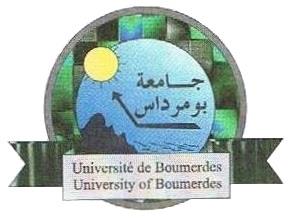}\hfill
\includegraphics[width=3.5cm]{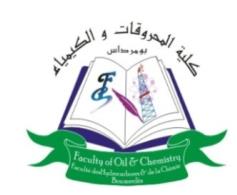}

\textbf{Department of Hydrocarbons Transportation \& Equipment} 
\end{center}

\vspace{1cm}

\textbf{Filière:} Hydrocarbons\\
\textbf{Spécialité:} Mechanical Engineering\\
\textbf{Option:} Upstream Mechanical Engineering (MACP-15) 

\vspace{1cm}

\begin{center}
	\textbf{Master Thesis} 

	Présenté par: LATRACH Abdeldjalil

	\vspace{.5cm}

	Theme:

	\vspace{.5cm}
\begin{tikzpicture}
\draw [very thick, rounded corners=9pt] (-7,-1.5) rectangle (7,1.5);
\node[text width=14cm, align=center] at (0,0) {\Large Application of Deep Learning for\\ Predictive Maintenance of\\ Oilfield Equipment};
\end{tikzpicture}

\end{center}

Devant le jury:

HALIMI Djamel\hspace{2cm} Docteur\hspace{2cm} UMBB\hspace{2cm} Encadreur

\vfill

\begin{center}
	Boumerdès : 2020
\end{center}

%% file: misc/dedication.tex
\begin{center}
	\pgfornament[width=.8cm]{65}
                \vspace{.3cm}\\
                \textsc{\Large Dedication}\\
                \pgfornament[width = 4cm]{88}

		\vspace{2cm}
		\textit{I dedicate this humble work to my dear parents, to everyone who stood by my side and supported me through my academic journey, to everyone who ever taught me a letter.}

		\newpage
\end{center}

%% file: misc/acknowledgement.tex
\begin{center}
                \textsc{\Large Acknowledgement}\\

		\vspace{2cm}
		I would like to thank EXPRO family for the host during my internship period. I acquired during those few weeks of my stay there—which was suddenly and sadly terminated due to COVID-19 lockdown—more valuable knowledge than what I was taught during whole semesters of university. A special thank for Mr. Aziz Menai and Mr. Zohir Meksen and all family of production department.

I would like to thank the whole opensource software community for providing amazing set of tools and voluntary technical support which without it this work wouldn’t have been brought out in this form.

I would like to thank the whole family of Petroleum Club at University of Boumerdes. These past few years were full of memories and valuable experiences which aren’t and couldn’t be taught in a classroom.
	\newpage	
\end{center}

%% file: chapters/abstract.tex
\chapter*{Abstract}
\vspace{-2cm}
This thesis explored applications of the new emerging techniques of artificial intelligence and deep learning (neural networks in particular) for predictive maintenance, diagnostics and prognostics. Many neural architectures such as fully-connected, convolutional and recurrent neural networks were developed and tested on public datasets such as NASA C-MAPSS, Case Western Reserve University Bearings and FEMTO Bearings datasets to diagnose equipment health state and/or predict the remaining useful life (RUL) before breakdown. Many data processing and feature extraction procedures were used in combination with deep learning techniques such as dimensionality reduction (Principal Component Analysis) and signal processing (Fourier and Wavelet analyses) in order to create more meaningful and robust features to use as an input for neural networks architectures. This thesis also explored the potential use of these techniques in predictive maintenance within oil rigs for monitoring oilfield critical equipment in order to reduce unpredicted downtime and maintenance costs.

Keywords: predictive maintenance, prognostics, deep learning, neural networks, signal processing

%% file: misc/epigraph.tex
\clearpage\mbox{}\vfill
\begin{epigraphs}
	\qitem{``All models are wrong, but some are useful''.}{George E. P. Box}
\end{epigraphs}
\par\vfill\clearpage

%% file: chapters/introduction.tex
\chapter*{Introduction}
\addcontentsline{toc}{chapter}{Introduction}
\markboth{Introduction}{Introduction}

All machines in general, and mechanical machines in particular, are subject to degradation of their condition and deterioration of their performance over time, possibly leading to their failure. These failures can have negative impacts on economic, human, environmental aspects.
Machine failure is an intrinsic property of these systems (due to their inherent physical properties), it can be - in order to avoid negative results - partially prevented, delayed and even foreseen, but it can never be totally prevented or stopped, this is mainly achieved through maintenance.

Maintenance is defined as a set of activities intended to maintain or restore a utility unit in a state in which it can perform a required function \cite{ISO2015}.

The way in which industrial maintenance is performed has evolved with advances in technology from the beginning of the industrial revolution to the present day. Its most basic form is the unplanned corrective maintenance that is carried out after a failure has occurred, in order to restore a functional unit to a state where it can perform a required function\cite{ISO2015}. Prior to World War II, machinery was relatively simple and production demand was moderate, so that it could be maintained after a breakdown. After the war and with the rebuilding of the industry, the market became more competitive and less tolerant of downtime, so the industry turned to preventive maintenance (\acrlong{pm}: \acrshort{pm}) which is carried out at predetermined intervals or according to prescribed criteria to reduce the probability of failure of a functional unit \cite{ISO2015}. Corrective and preventive maintenance are both obsolete. The goal is to replace "post-mortem" and "blind" maintenance with "just-in-time" maintenance. 
It was therefore only a matter of time before the idea of predictive or condition-based maintenance (\acrlong{cbm}: \acrshort{cbm}) emerged, which differs from other forms by basing the need for intervention on the actual state of the machine rather than on a pre-set schedule\cite{Kadry2013}. There is disagreement in the literature on this classification (of \acrshort{cbm} as preventive maintenance or a separate form)\cite{Shin2015}. The difference in taxonomy is not of real interest to the current discussion.

A new field that has recently emerged is that of health prognosis and management (\acrlong{phm}: \acrshort{phm}). \acrshort{phm} has emerged as an essential approach to achieving competitive advantages in the global marketplace by improving reliability, maintainability, safety and affordability. Like \acrshort{cbm}, \acrshort{phm} is a discipline that has emerged in industry with a modest presence in the academic field, particularly in military applications\cite{Tinga2014}. (e.g. the development of the F-35 fighter plane \cite{Brown2007}).
The concepts and components of the \acrshort{phm} have been developed separately in many fields such as mechanical, electrical and statistical engineering, under various names\cite{Tsui2015}. While \acrshort{cbm} focuses on system monitoring, \acrshort{phm} is a more integrated approach that aims to provide guidelines for system health management. It is therefore a life cycle management philosophy (\acrlong{lcm}: \acrshort{lcm}) that emphasizes predictability (i.e., prognosis) of failures and maintenance. This is usually achieved through the adoption of a monitoring strategy, which can be a technique of \acrshort{cbm} \cite{Tinga2014}. The main goal of the prognosis (also \acrshort{cbm}) is the estimation of the remaining useful life (\acrlong{rul}: \acrshort{rul}) given the current state of the equipment and its historic operational profile\cite{Jardine2006}.

The \acrshort{phm} operates at a slightly higher level than the \acrshort{cbm}, since it has the clear ambition to enable health management. This is an activity related to \acrshort{lcm}, which means that an approach is followed to optimize all (maintenance) activities during the complete life-cycle of the equipment. This includes choosing an appropriate maintenance policy, defining the intervention schedule and deciding when a piece of equipment should be taken out of service. \acrshort{cbm} does not offer this extensive \acrshort{lcm} support. The \acrshort{phm} domain prescribes neither a specific maintenance concept nor a monitoring strategy. However, in typical \acrshort{phm} studies, \acrshort{cbm} or other maintenance policies are adopted and in many cases, condition monitoring techniques (\acrlong{cm}: \acrshort{cm}) are applied\cite{Tinga2014}.

As mentioned earlier, all these efforts to develop maintenance strategies have been mainly due to the high costs of production loss and reactive maintenance in capital-intensive industries such as Onshore and Offshore projects in the oil and gas industry are mainly investments, which can have serious financial and environmental consequences if a catastrophic failure occurs. Therefore, an effective approach to maintenance management is essential to continue production safely and reliably \cite{Telford2011}. Offshore organizations experience an average financial impact of \$38 million per year due to unplanned downtime (for the worst performing organizations, \$88 million). Less than 24\% of operators describe their maintenance approach as predictive and based on data and analysis. More than three-quarters take a reactive or time-based approach. Operators using a predictive data-based approach experience on average 36\% less unplanned downtime than those using a reactive approach. This can translate, on average, into a \$34 million per year decline in bottom line earnings\cite{Eriksen2016}.

In this thesis, a data-driven prognostic approach will be introduced, outlining the various steps required from data acquisition to the estimation of \acrshort{rul}. Due to the commercial and technical sensitivity of the data in the petroleum field (and all industries with high added value) and therefore the impossibility to obtain them, the public databases (NASA Ames Data Repository \cite{NASAAAmes}) that are usually used to evaluate prognostic algorithms in the literature will be used here as well. An approach will be proposed to transfer the knowledge obtained to concrete applications in the petroleum industry.
\newline
This thesis is divided as follows:
\begin{description}
	\item [Chapter 01] Towards a Data-Driven Prognostics Approach
	\item [Chapter 02] Steps of a Data-Driven Approach
	\item [Chapter 03] Introduction to Artificial Neural Networks
	\item [Chapter 04] Equipment Health Assessment using Artificial Neural Networks
	\item [Chapter 05] Bearings Faults Diagnostics and Prognostics
	\item [Conclusion] 
\end{description}

%% file: chapters/chapter01.tex
\chapter{Towards a Data-Driven Prognostics Approach}

\chapterintrobox{
The aim of this chapter is to present the different prognostic approaches with a detailed taxonomy, the different steps of any prognostic approach will be described and then the emphasis will shift towards data-driven methods.
}

\section{Prognostics of mechanical equipment}
Some complex systems, especially in the oil industry, operate under very severe conditions (offshore, desert, etc.), which can lead to the occurrence of failures and their degradation. Breakdowns and unplanned shutdowns inevitably cause production losses, which can have enormous economic consequences. With these economic constraints, maintenance programs must be developed to minimize the probability of failures, thereby reducing the cost. As discussed in the introduction to this thesis, these programs must be based on the principles of \acrlong{cbm} and \acrlong{phm}.

Prognostics and health management (\acrlong{phm}) has two major aspects \cite{Hess2008}:

\begin{enumerate}
    \item \textbf{Prognostics}: Predictive diagnostics, which includes determining the remaining useful life (service life) of a component or a full system.
    \item \textbf{Health Management}: the ability to make decisions regarding maintenance actions based on diagnostic/prognostic information, available resources and operational demand.
\end{enumerate}

\section{Estimating Remaining Useful Life}
\label{section:rul}
\label{section:rul-estimation}
The main objective of the prognostics is to estimate the remaining useful life (\acrlong{rul}) of the system.
\acrshort{rul} is defined according to the equation \ref{eq:rul}:

\begin{equation}
    RUL = t_f-t_c
    \label{eq:rul}
\end{equation}

Where $t_f$ is the predicted time for the occurrence of the failure and $t_c$ is the current time (the time when the prediction is made).

\section{Physical, Data-Driven and Hybrid Approaches}
\label{section:prognostics-approaches}
\begin{figure}[ht]
    \centering
	\input{figures/prognostic-approaches-venn.tex}
    \caption{Classification of prognostics approaches}
    \label{fig:prognostic-approaches-venn}
\end{figure}
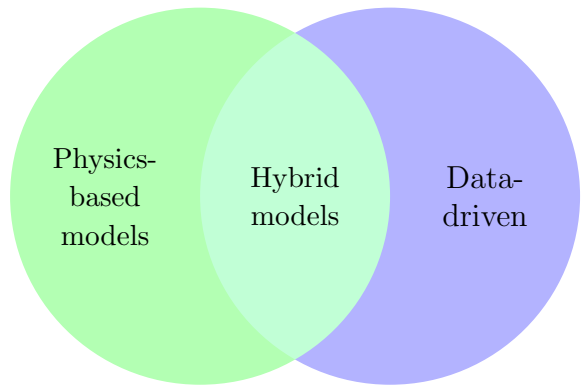

Any prognostic approach can be based on physical models, data-driven models or a hybrid combination of the two (Figure \ref{fig:prognostic-approaches-venn}).

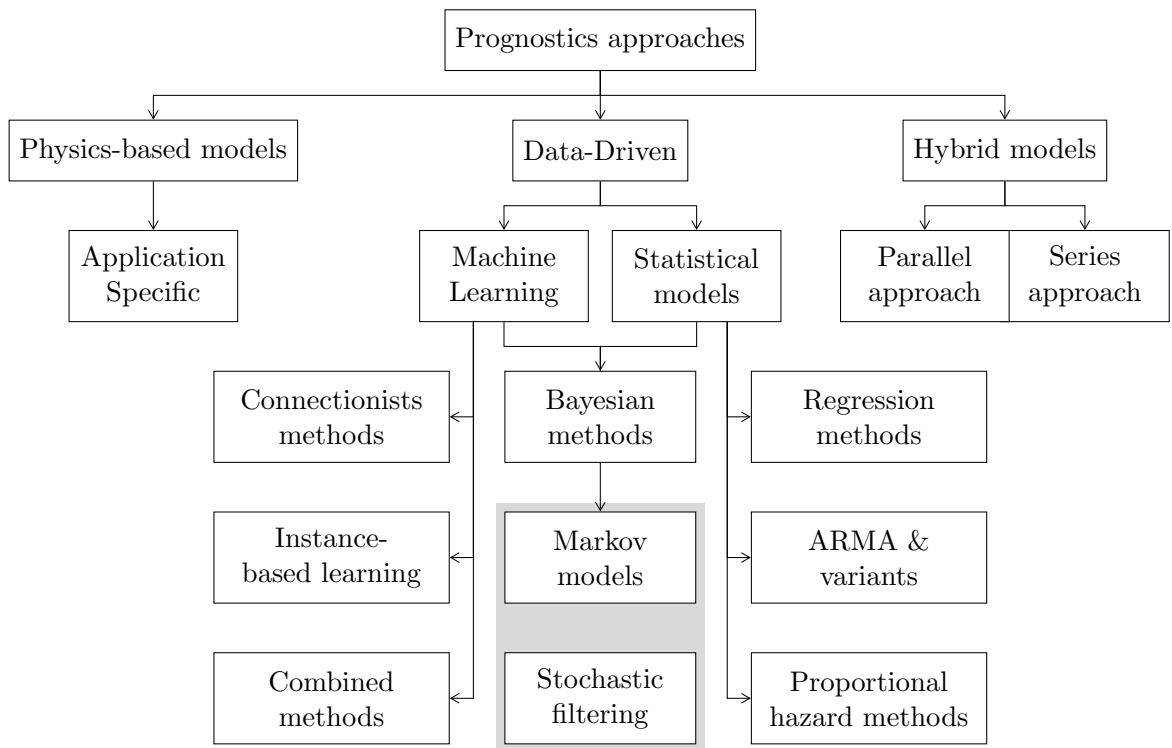
\begin{figure}[ht]
	\resizebox{\textwidth}{!}{\input{figures/prognostic-approaches.tex}}
    \caption{Taxonomy of prognostics approaches \cite{Javed2017}}
    \label{fig:prognostic-approaches-tree}
\end{figure}

These three categories constitute a general classification based on the approach followed, each of which can be subdivided into sub-categories. A detailed taxonomy is presented in Figure \ref{fig:prognostic-approaches-tree}.

\subsection{Physics-based models}
Physics-based models assess the health of the system using an explicit mathematical formulation (white boxes) developed on the basis of a scientific and technical understanding of its behavior. However, the main advantage of these physical models is the use of degradation models to predict long-term behavior \cite{Cubillo2016}. Physical approaches are capable of providing an accurate estimate of the health of the system if the physical model is developed with a complete understanding of the failure mechanisms and efficient estimation of the model parameters. However, for some complex mechanical systems, it is difficult to understand the physics of damage, which limits the application of these approaches \cite{Lei2018}.

\subsection{Data-Driven Models}
\acrlong{dd}  models rely on previously collected data (monitoring data, data on operational parameters, …) to establish a model capable of assessing the health of the system and predicting its behavior and degradation. Contrary to physical models, and as their name indicates, \acrlong{dd} models do not rely on human knowledge but mainly on historical data collected to model the degradation process. Usually, they are considered as black boxes.

\subsection{Statistical Models}
The statistical approach is based on the construction and fitting of a probabilistic model using historical data without depending on any physical or technical principles. 
Si et al. \cite{Si2011} presented a review of statistical approaches. According to this review, many models fall into this category such as regression models (e.g. linear regression), auto-regressive moving average and its variants, stochastic filtering techniques (e.g. Kalman filter, particles filter, …).

\subsection{Machine Learning}
Machine Learning is a field of Artificial Intelligence that has exploded in popularity during recent years and has made breakthroughs in many areas such as computer vision and natural language processing. Machine learning models are black box models that allow even very complex mappings from input to output to be discovered. Many types of algorithms fall into this category such as connectionist methods (e.g. artificial neural networks), context learning (e.g. support-vector machines). Different approaches can be combined together to create mixed models that may perform better than a single model.

\subsection{Hybrid Models}
\begin{figure}[H]
    \centering
    \input{figures/hybrid-approach-series.tex}
    \caption{Series hybrid configuration (Figure adapted from \cite{Mangili2013})}
    \label{fig:hybrid-approach-series}
\end{figure}
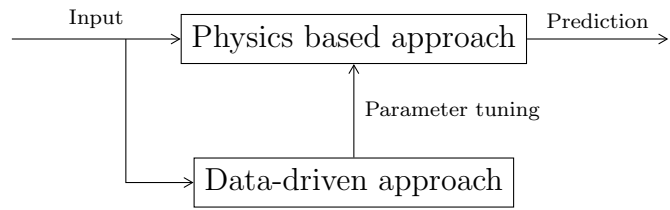

Hybrid models are a combination of a physical model and an \acrlong{dd} model. There are two types of hybrid models depending on how the two types of models are combined. The \acrlong{dd} model can be integrated into a physical model in a series configuration (Figure \ref{fig:hybrid-approach-series}) where it is used to adjust the parameters of the physical model which is then used to make predictions.

\begin{figure}[H]
    \centering
    \input{figures/hybrid-approach-parallel.tex}
    \caption{Parallel hybrid configuration (Figure adapted from \cite{Mangili2013})}
    \label{fig:hybrid-approach-parallel}
\end{figure}
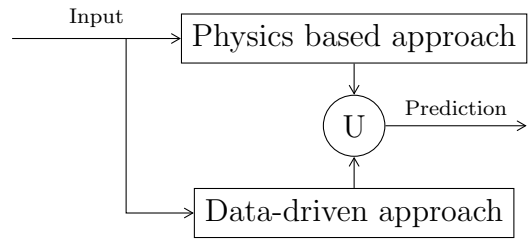

The two types of models can be combined in a parallel configuration (Figure \ref{fig:hybrid-approach-parallel}) where the two models make separate predictions that can be combined to obtain the final estimate.

\section{Why a Data-Driven Approach?}
As mentioned previously, understanding the degradation process for very complex systems is extremely difficult, which is why the development of physical models is very problematic for these systems.

The last 20 years have seen great progress in the development of new detection techniques, prognostics/diagnostics methods and in the application of computer-aided analysis methods. 

It is interesting to note that at the 2002 workshop on condition-based maintenance organized by the Advanced Technology Program of the United States National Institute of Standards and Technology (NIST), the following obstacles to the widespread application of CBMs were identified:
\begin{itemize}
    \item The impossibility of accurately and reliably predicting the remaining service life of a machine.
    \item The inability to continuously monitor a machine.
    \item The Inability of maintenance systems to learn and identify impending failures and recommend action.
\end{itemize} 

These barriers can be redefined as deficiencies in prognosis, detection and reasoning. These and other limitations in the current implementation of condition-based maintenance techniques have, of course, been recognized by others and have led to the development of programs (e.g. in the military field) to overcome them \cite{Hess2008}.

Today, sensors in industry have become inexpensive and ubiquitous, computing power has increased exponentially which has allowed the development of more advanced algorithms and computing tools: the artificial intelligence revolution and machine learning. The industry generates huge amounts of data in many areas, the majority of which is untapped. Exploiting this data using the latest technological advances can increase profits, drastically reduce costs and prove to be a huge economic advantage.

\section{Conclusion}
Adoption of data-driven prognostics models can be helpful especially when behavior of degradation process is ambiguous and developing physics-based models to quantify the health state of complex systems is complicated and their results become unreliable. Recent developments in sensoring techniques, increase in available computation power (i.e. faster and cheaper processing units) and also abundance of—unexploited—monitoring data has provided the needed framework for adopting these data-driven models which are easier to develop, deploy and automate than their physics-based counterparts.

%% file: figures/prognostic-approaches-venn.tex
\begin{tikzpicture}
\begin{scope}
[blend group = soft light]
\fill [green!30!white] (0,0) circle (2.5);
\fill [blue!30!white] (2.5,0) circle (2.5);
\end{scope}

\node[text width=2cm, align=center] at (-1.25,0) {\small Physics-based models};
\node[text width=1.5cm, align=center] at (3.75,0) {Data-driven};
\node[text width=1.6cm, align=center] at (1.25,0) {\small Hybrid models};
\end{tikzpicture}

%% file: figures/prognostic-approaches.tex
\begin{tikzpicture}
	[all/.style={draw, minimum height=2em, fill=white, font=\small}]

	\fill [gray!30!white] (-1.41,-6.3) rectangle (1.41,-9.7) ;
	
	\draw(0,0) node[all] (progApp)                       	{Prognostics approaches};
	\draw node[all, below = 1.6em of progApp] (dataDriv)		{Data-Driven};
	\draw node[all, left = 7em of dataDriv] (physBas)      	{Physics-based models};
	\draw node[all, right = 7em of dataDriv] (hybridApp)				{Hybrid models};

	\draw node[all,text width=2cm,align=center,minimum height=3em, below = 1.6em of physBas] (appSpec)		{Application Specific};
	
	\begin{scope}[node distance=1.6em and -3.5em]
		\draw node[all,text width=2cm,align=center,minimum height=3em, below right = of hybridApp] (serApp)		{Series approach};
		\draw node[all,text width=2cm,align=center,minimum height=3em, below left = of hybridApp] (parApp)		{Parallel approach};
	\end{scope}
	
	\begin{scope}[node distance=1.6em and -2.5em]
		\draw node[all,text width=2cm,align=center,minimum height=3em, below right = of dataDriv] (statMod)		{Statistical models};
		\draw node[all,text width=2cm,minimum height=3em, align=center, below left = of dataDriv] (ML)		{Machine Learning};
	\end{scope}

	\begin{scope}[node distance=1.6em and 0em]
		\draw node[xshift=1em,all,text width=2.9cm,align=center,minimum height=3em, below left = of ML] (connect) {Connectionists methods};
		\draw node[all,text width=2.9cm,align=center,minimum height=3em, below = of connect] (instance) {Instance-based learning};
		\draw node[ all,text width=2.9cm,align=center,minimum height=3em, below = of instance] (comb)		{Combined methods};
	\end{scope}

	\begin{scope}[node distance=1.6em and 0em]
		\draw node[xshift=-1em,all,text width=2.9cm,align=center,minimum height=3em, below right = of statMod] (reg)		{Regression methods};
		\draw node[all,text width=2.9cm,align=center,minimum height=3em, below = of reg] (arma)		{ARMA \& variants};
		\draw node[all,text width=2.9cm,align=center,minimum height=3em, below = of arma] (propor)		{Proportional hazard methods};
	\end{scope}

	\begin{scope}[node distance=1.6em and 0em]
	\path let \p1 = (connect) in node[all,text width=2.3cm,align=center,minimum height=3em] (bayes)	at (0,\y1)	{Bayesian methods};
	\draw node[all,text width=2.3cm,align=center,minimum height=3em, below  = of bayes] (hmm)		{Markov models};
	\draw node[all,text width=2.3cm,align=center,minimum height=3em, below  = of hmm] (sotch)		{Stochastic filtering};
	\end{scope}

	\draw[->, >=angle 60] (progApp.south)   -- ++(0,0) -- ++(0,-0.8em) -| (physBas.north);
	\draw[->, >=angle 60] (progApp.south)   -- ++(0,0) -- ++(0,-0.8em) -| (hybridApp.north);
	\draw[->, >=angle 60] (progApp.south)   --  (dataDriv.north);
	
	\draw[->, >=angle 60] (physBas.south)   --  (appSpec.north);
	
	\draw[->, >=angle 60] (hybridApp.south)   -- ++(0,0) -- ++(0,-0.8em) -| (parApp.north);
	\draw[->, >=angle 60] (hybridApp.south)   -- ++(0,0) -- ++(0,-0.8em) -| (serApp.north);
	
	\draw[->, >=angle 60] (dataDriv.south)   -- ++(0,0) -- ++(0,-0.8em) -| (statMod.north);
	\draw[->, >=angle 60] (dataDriv.south)   -- ++(0,0) -- ++(0,-0.8em) -| (ML.north);
	
	\draw[->, >=angle 60] ([xshift=-1em]ML.south)   |- (connect.east);
	\draw[->, >=angle 60] ([xshift=-1em]ML.south)   |- (instance.east);
	\draw[->, >=angle 60] ([xshift=-1em]ML.south)   |- (comb.east);
	
	\draw[->, >=angle 60] ([xshift=1em]statMod.south)   |- (reg.west);
	\draw[->, >=angle 60] ([xshift=1em]statMod.south)   |- (arma.west);
	\draw[->, >=angle 60] ([xshift=1em]statMod.south)   |- (propor.west);

	\draw[->, >=angle 60] (ML.south)   -- ++(0,0) -- ++(0,-0.8em) -| (bayes.north);
	\draw[-] (statMod.south)   -- ++(0,0) -- ++(0,-0.8em) -| (bayes.north);
	\draw[->, >=angle 60] (bayes.south)   -- (hmm.north);
\end{tikzpicture}

%% file: figures/hybrid-approach-series.tex
\begin{tikzpicture}
 	\node[draw, rectangle] (ph) {Physics based approach};
 	\node[draw, rectangle, below = 3em of ph] (dd) {Data-driven approach};
 	
 	\draw[->, >=angle 60] (dd.north) -- node[right] {\scriptsize  Parameter tuning} (ph.south);
 	\draw[->, >=angle 60] (ph.east) -- node[above] {\scriptsize Prediction} ([xshift=4.5em]ph.east);
 	\draw[->, >=angle 60] (-4.5,0) -- node[above] {\scriptsize Input} (ph.west);
	\draw[->, >=angle 60] (-3,0) |-  (dd.west);
\end{tikzpicture}

%% file: figures/hybrid-approach-parallel.tex
\begin{tikzpicture}
    \node[draw, rectangle] (ph) {Physics based approach};
    \node[draw, circle, below = 1em of ph] (u) {U};
    \node[draw, rectangle, below = 1em of u] (dd) {Data-driven approach};

    \draw[->, >=angle 60] (u.east)      --      node[above] {\scriptsize Prediction} ([xshift=4.5em]u.east);
    \draw[->, >=angle 60] (ph.south)    --      (u.north);
    \draw[->, >=angle 60] (dd.north)    --      (u.south);
    \draw[->, >=angle 60] (-4.5,0)      --      node[above] {\scriptsize Input} (ph.west);
    \draw[->, >=angle 60] (-3,0)        |-      (dd.west);
    \draw[->, draw=none] (ph.east) -- ([xshift=4.5em]ph.east);
\end{tikzpicture}

%% file: chapters/chapter02.tex
\chapter{Steps of a Data-Driven Approach}
\chapterintrobox{
A \acrlong{dd} prognostics approach (an approach that uses system's operation data, like historic monitoring data, to build a model that is used to predict remaining useful life) must go through multiple stages---from data acquisition to estimating remaining service life. In this chapter, these different steps will be discussed in detail.
}

\section{Data acquisition}
\label{sec:data-acquisition}
A signal is a function that conveys information about the behavior of a system or the attributes of a phenomenon. Signals occur naturally and are also synthesized. A signal is not necessarily an electrical quantity. However, to perform activities such as synthesizing, transporting, recording, analyzing and modifying signals, it is often convenient to use a signal in the form of an electrical quantity \cite{Priemer1990}.

Data acquisition is a process of capturing and storing different types of monitoring data (i.e. signals) from various sensors installed on the monitored equipment. It is the first machine prognostics process, which provides basic condition monitoring information for subsequent processes. A data acquisition system is composed of sensors, data transmission devices and data storage devices \cite{Lei2018}. Condition monitoring data is very versatile. It can be vibration data, acoustic data, oil analysis data, temperature, pressure, humidity, weather or environmental data, and so on. Different sensors, such as micro-sensors, ultrasonic sensors and acoustic emission sensors have been designed to collect different types of data. Wireless technologies, such as Bluetooth, have provided a cost-effective alternative to data communication \cite{Jardine2006}.

Although research on advanced concepts such as wireless sensor networks and energy harvesting to power autonomous sensors is ongoing, data acquisition (sensors) and manipulation are now fairly well established. Consequently, much of the research in this discipline focuses on analyzing the data obtained to extract information \cite{Tinga2014}. Because this discipline is well developed, many new data acquisition facilities and techniques have been designed and applied in modern industries. These powerful and versatile facilities have made data acquisition for \acrshort{phm} implementation more practical and feasible\cite{Lei2016}.

\section{Features Extraction}
The data-driven prognostics approach is mainly used when it is difficult to understand the physical behavior of a complex system. Understanding the behavior and interaction of the different elements that lead to machine degradation is the starting point for developing a physical model for prognostics.
On the other hand, this approach uses condition monitoring data to implicitly model its behavior. The models used in the data-driven approach use monitoring data to model complex behavior and capture complex models, but they are considered as black boxes: they do not necessarily provide insight into the process.
In general, the performance of these models depends on the quality of the input (i.e. data). The human part cannot perform the task that the model performs, but processing the input data can significantly increase the results. This processing is necessary because the sensor data on which the model is based is usually redundant, noisy and incomplete, and there are many reasons for these imperfections.

Feature extraction is an important pre-processing step in the development process of machine learning models and directly influences the performance of the model. Therefore, this step must be performed carefully in order to extract meaningful features from the raw data. Vibration data contains very useful information about the state of the system, but requires extensive pre-processing before it can be used as input data for a specific model. This chapter describes some of the signal processing techniques used in traditional vibration analysis, but in this context they will be used as feature extractors for a neural network architecture.

\subsection{Signal analysis}
Signal processing is the study and analysis of stored signals to reveal their properties-which may not be apparent at first glance-using a set of algorithms and techniques. In the context of condition monitoring, these properties revealed by signal processing may be indicative of the health of the machine.
Signal processing is a well-established and mature sub-domain of electrical engineering, with many techniques and algorithms proposed in the literature.

Signal processing can be divided into three categories: \textbf{Time-Domain Analysis}, \textbf{Frequency-Domain Analysis} and \textbf{Time--Frequency Analysis}.

\subsection{Temporal Analysis}.
The original measurements of signals that are usually repeatedly sampled between predefined time intervals are in the form of a time series. Thus, time domain analysis is directly based on the original measurement \cite{Lei2016}.

\subsection{Frequency-Domain Analysis}
Frequency domain analysis is based on signals transformed in the frequency domain. The advantage of frequency-domain analysis over time-domain analysis is its ability to decompose the original signals into a series of frequency components. The most commonly used frequency domain analysis is spectrum analysis using Fast Fourier Transform (FFT). The main idea of spectrum analysis is to isolate and locate certain frequency components of interest related to the fault characteristics of \cite{Lei2016a} machines.

\subsection{Time--Frequency Analysis}
The problem with time-domain analysis and frequency-domain analysis is that each has no information on the other (time-domain analysis has no information on the frequency-domain, and frequency-domain analysis has no information on the time-position).

Thus, time--frequency domain analysis, which
studies measurement signals in the time and frequency domains, has been applied to the analysis of non-stationary measurement signals. Time--frequency analysis describes the characteristics of measurement signals in two-dimensional functions of time and frequency to better reveal machine failure modes.

Figure \ref{fig:signal-processing} presents different techniques for each type of analysis:

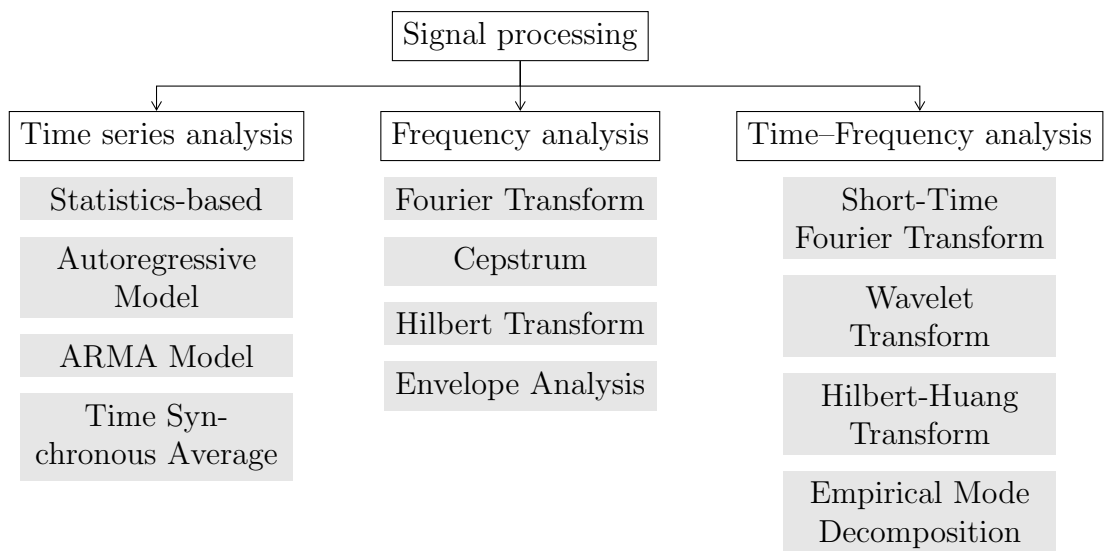
\begin{figure}[h]
    \centering
	\input{figures/feature-extraction.tex}
    \caption{Different signal processing techniques}
    \label{fig:signal-processing}
\end{figure}

\subsection{Fourier analysis}
Fourier analysis, also called harmonic analysis, of a periodic signal $x(t)$ is the decomposition of the series into summation of sinusoidal components, where each sinusoid has a specific amplitude and phase.

The Fourier transform (FT) of a signal $x(t)$ can be mathematically given by equation \ref{equation:fourier-transform}:

\begin{equation}
    X(w) = \int_{-\infty}^{\infty}x(n)e^{-jwt}dt
    \label{equation:fourier-transform}
\end{equation}

In practical applications of digital signal processing where signals are discrete in time rather than continuous (e.g. vibration analysis) a discretized version called discrete Fourier transform (DFT) is used instead, it is expressed mathematically by equation \ref{equation:discrete-fourier-transform}:

\begin{equation}
    X(w) = \sum_{-\infty}^{\infty}x(t)e^{-jwt}dt
    \label{equation:discrete-fourier-transform}
\end{equation}

Fast Fourier transform (FFT) is an effective algorithm used to implement DFT in computers. Figure \ref{figure:fft} shows a signal in its waveform (or time domain) and its corresponding spectrum (frequency domain) obtained using FFT algorithm. The spectrum shows the frequency components present in the signal:

\begin{figure}[H]
    \centering
    \includegraphics{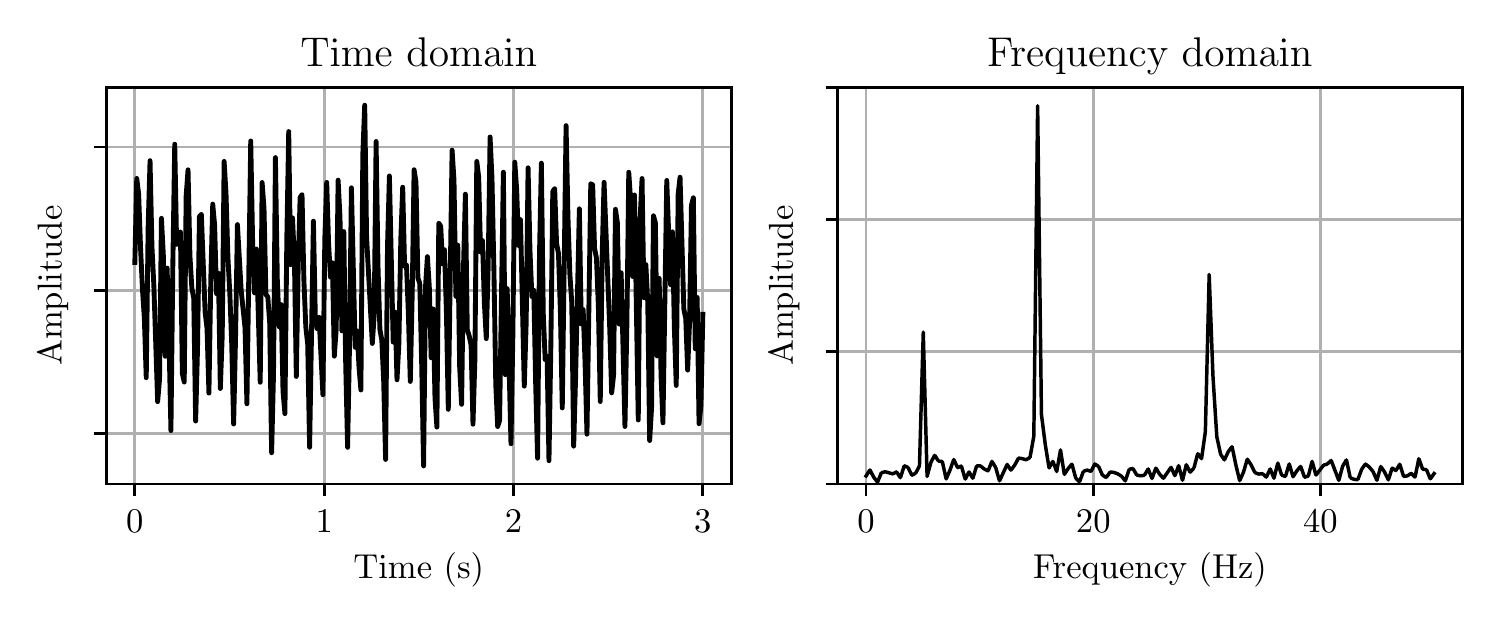}
    \caption{Signal in the time domain and its fast Fourier transform}
    \label{figure:fft}
\end{figure}

\subsection{Wavelet transform}
Wavelet transform is also a spectral analysis tool, like Fourier transform. The main difference is that Fourier transform decomposes the signal into sinusoidal components, but wavelet transform decomposes it into a set of oscillatory functions called \textbf{wavelets}. Unlike sinusoids, wavelets are localized in time, thus wavelet transform doesn't only provide information about the frequency present in a signal but also the time of their occurrence. Wavelet transform is a much better solution than Fourier transform when studying non-linear non-stationary signals (i.e. its frequency components vary with time).

Figure \ref{fig:time-frequency-plane} shows the difference in time and frequency resolutions between different methods. In the waveform, the signal has absolute resolution in time and zero resolution in frequency. Fourier transform on the contrary transforms the signal totally into the frequency domain, therefore it has absolute resolution in frequency but no resolution in time. Short-time Fourier transform is calculated identically to Fourier transform but it is performed on separate segments of the original signal to preserve some resolution in time. Wavelet transform on the other hand exhibits a high time resolution for high frequencies and high frequency resolution for low frequencies:

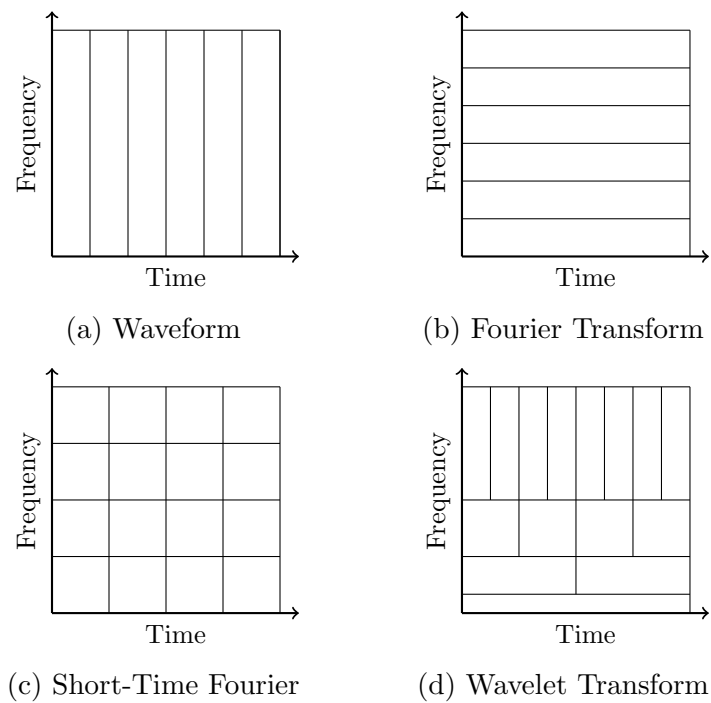
\begin{figure}[H]
    \centering
    \input{figures/time-frequency-resolution.tex}
    \caption{Time—frequency resolution plane}
    \label{fig:time-frequency-plane}
\end{figure}

There are a wide variety of wavelets that serve different purposes like Morlet wavelet, Daubechies wavelet and many others.

\begin{figure}[H]
    \centering
    \includegraphics{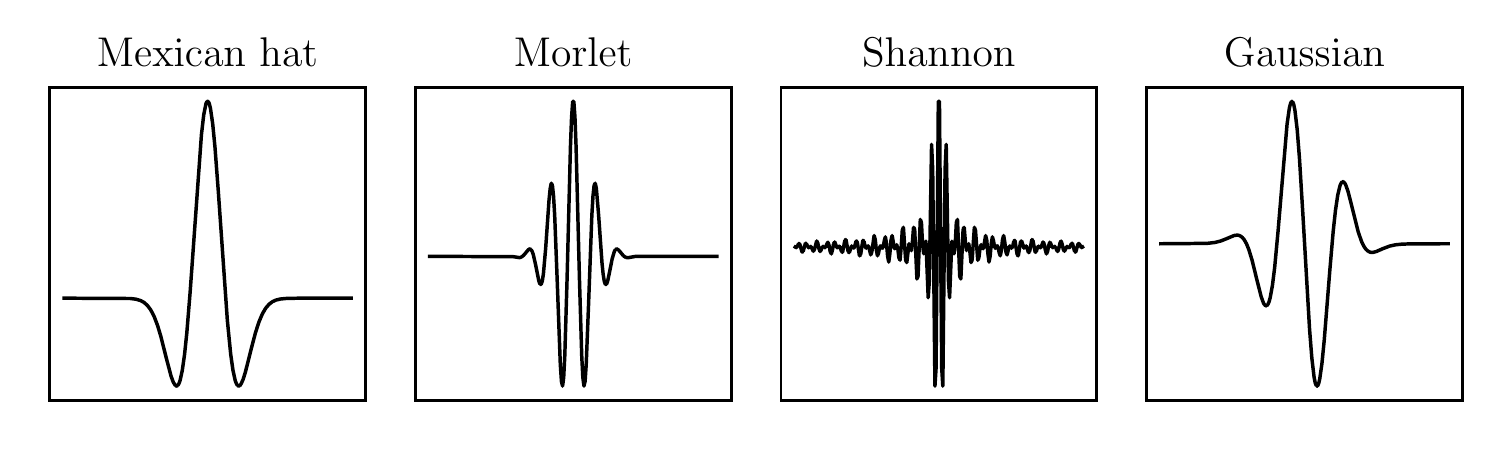}
    \caption{Different types of wavelets}
    \label{fig:wavelets}
\end{figure}

\subsubsection{Continuous wavelet transform}
Mathematically, continuous wavelet transform is defined by equation \ref{equation:cwt}:

\begin{equation}
    CWT_x^\psi(\tau, s)=\frac{1}{\sqrt{|s|}}\int_{-\infty}^{\infty}x(t)\psi^* \left(\frac{t-\tau}{s}\right)dt
    \label{equation:cwt}
\end{equation}

Where $x(t)$ is the original signal, $\psi^*$ is a function called the \textbf{mother wavelet}; $s$ and $\tau$ are the \textbf{scale} and \textbf{translation} parameters respectively. The original signal is multiplied by the mother wavelet which is scaled using different scales then translated over the signal.

The output of \acrshort{cwt} is a scaleogram like the one in figure \ref{fig:scaleogram} which is a scaleogram (filled contour plot) of vibrations data snapshot of 25ms:

\begin{figure}[H]
    \centering
    \includegraphics{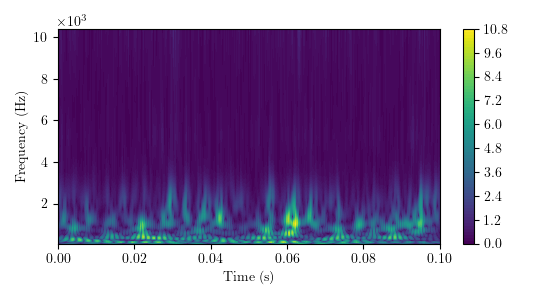}
    \caption{Scaleogram of vibration data snapshot}
    \label{fig:scaleogram}
\end{figure}

The x and y axes represent time and frequency respectively. Different colors indicate the power (i.e. amplitude) of each frequency (y-axis) during each instant of time (x-axis) which—unlike Fourier transform—provides information about the frequencies present in the signal and also the instances of time when these frequencies are present.

\subsubsection{Discrete wavelet transform}%
\label{subsub:discrete_wavelet_transform}
In practical applications, discrete wavelet transform (DWT) is implemented as a filter bank where the signal is passed through low- and high-pass filters to obtain \textbf{approximation} and \textbf{decomposition coefficients}. Figure \ref{fig:dwt} shows a DWT with 2 levels of decomposition which yields 2nd order approximation and decomposition coefficients:

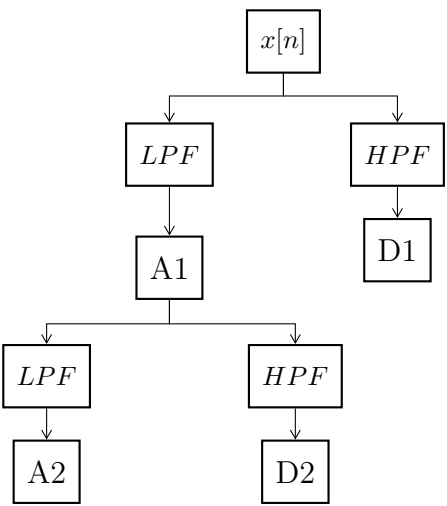
\begin{figure}[H]
    \centering
    \input{figures/dwt.tex}
    \caption{Discrete wavelet transform (DWT) as a filter bank}
    \label{fig:dwt}
\end{figure}

\begin{figure}[H]
    \centering
    \includegraphics{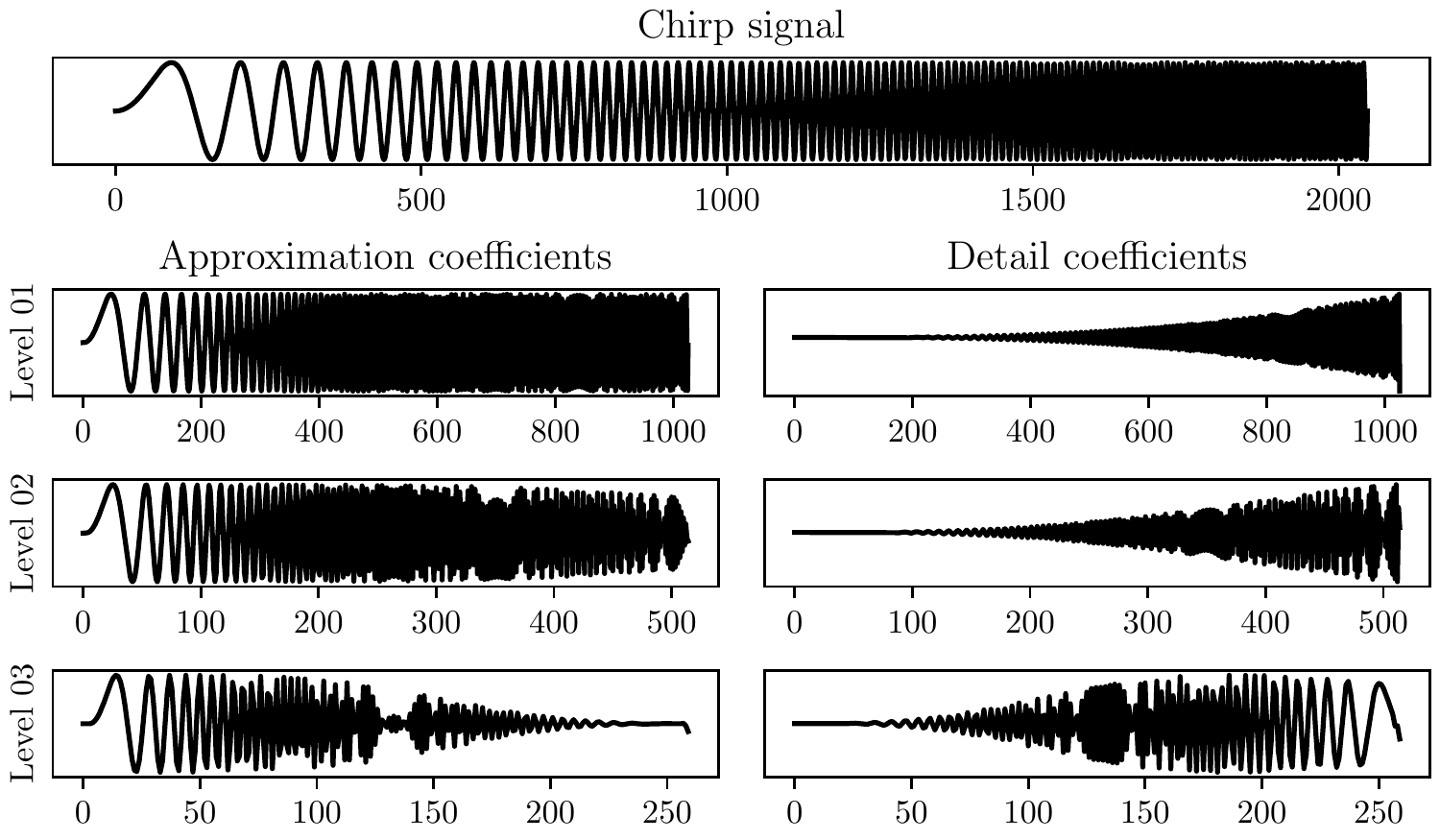}
    \caption{Level 3 signal decomposition using DWT}
    \label{fig:dwt-chirp-signal}
\end{figure}

DWT returns two sets of coefficients: \textbf{approximation coefficients} associated with the low pass filter and \textbf{detail coefficients} associated with the high pass filter of the DWT. By applying DWT again on the approximation coefficients the next level of decomposition can be obtained. At each level the original signal is downsampled by a factor of 2, this fact imposes a limitation on the possible number of decomposition levels for a given signal.

\subsection{Dimensionality reduction}
\label{section:dimensionality-reduction}
Dimensionality reduction refers to the process of taking high-dimensional data and finding a good representation of this data in a lower dimension while preserving its original characteristics. Dimensionality reduction is performed for feature extraction by finding the main components or for visualization (by reducing the number of dimensions to 2 or 3).
There are many dimensionality reduction techniques and algorithms such as:
\begin{itemize}
    \item Principal Component Analysis (PCA)
    \item Autoencoders
    \item t-Distributed Stochastic Neighbor Embedding (t-SNE)
\end{itemize}

It should be noted that for prognoses, where the data used consist of sensor inputs where each variable has a physical meaning and direct interpretation, performing a dimensionality reduction will give the main components, but the new variables \textit{will lose their physical interpretability} and \textit{shall be treated as abstractions}.

\section{Diagnostics}
Diagnostics is a process of identifying and determining the relationship between the information obtained in the measurement space and the failure modes of the machines in the failure space. Diagnostics consists of three main steps: fault detection, fault isolation and fault identification. Fault detection is the task of indicating whether a fault has already occurred in the monitored machines. Fault isolation consists of finding the component of the fault and the position of the fault. Fault identification is the final step in diagnosis, which attempts to determine the mode and severity of the failure. The three steps are interrelated. This last step is based on the results of the first one and therefore cannot be performed individually \cite{Lei2016b}.

\section{Prognostics}
Diagnosis is the analysis of the later event and prognosis is the analysis of the earlier event. Prognosis is much more effective than diagnosis in achieving performance without stopping production. However, diagnosis is necessary when the prediction of prognostic errors fails and an error occurs \cite{Jardine2006}.
Prognostics are the forecast or prediction of the future performance of a system, this prediction is based on its current state. The prediction is performed using a model, types of prognostic models are already discussed in details in section \ref{section:prognostics-approaches} but the focus of this current discussion is geared towards data-driven models. More concretely, what these models predict is the remaining useful life (\acrshort{rul}) defined in section \ref{section:rul-estimation}. Prognostics models---by estimating \acrshort{rul}---aim at scheduling maintenance actions according to the predictions and production constraints associated with the machine in order to achieve zero unscheduled equipment downtime.

\section{Maintenance decision}
The next and the last step of any prognostics approach is to use the output of the developed model (i.e. predictions, or estimations for the system's \acrshort{rul}) in making better and more accurate preventive maintenance actions. Maintenance decision makers should analyze the outputs of the model prior to taking any actions, since any model has an inherent error associated with its predictions. Usually, data-driven models provide confidence intervals associated with their predictions which must be considered carefully. Also it must be noted that those confidence intervals are reflection of the models certainty of its predictions based on the data provided for the training, in real applications a new unforeseen failure mode can occur which may not have happened before, thus it wasn't provided for the model as part of the data used to construct it. This is more or less associated with the ability of different types of models to generalize for new unforeseen data, but in general it is well-known fact that any data-driven model will be less reliable when performing predictions using data that the model didn't see before \cite{Chung2018}. That's why providing quality data that reflects different degradation patterns and fault modes is essential for the model to make more robust and reliable predictions.

\section{Conclusion}
Developing a (data-driven) prognostics model is a process that requires many steps: from data acquisition required to construct the model to employing the model predictions in making better and more accurate maintenance decisions which result in less (or even zero) unscheduled downtime this reduced costs and decreased production loss, the latter being the ultimate goal of all this discussion and prognostics/preventive maintenance literature in general. The focus of this discussion is data-driven models, one of the models that proved great ability in learning complex non-linear patterns in data are neural networks. These models have different architecture depending on the structure of data. They will be presented and explained in details in the next chapter.

\section{Conclusion}
Vibration data are discrete signals sampled at a certain frequency in time. Although they hold so many valuable information about equipment performance, these information are usually not directly observable in the time domain. Digital signal processing techniques offer a way to gain more insights from raw vibration data by converting it to frequency or time–frequency domains where unusual frequency components can indicate development of certain degradation pattern. This chapter introduced several of these techniques like Fourier and Wavelet transforms. A following chapter will present the use of these techniques for extracting features that serve as an input for a neural network that can estimate the remaining useful life.

%% file: figures/feature-extraction.tex
\begin{tikzpicture}
\tikzstyle{element}=[draw,rectangle]
\tikzstyle{entity}=[fill=gray!20,align=center, text width=8em]
\node[element] (fe) {Signal processing};

\node[element,below = 1.6em of fe] (fd) {Frequency analysis};
\node[element,left = of fd] (td) {Time series analysis};
\node[element,right = of fd] (tfd) {Time--Frequency analysis};

\node[entity,below =.5em of td] (stat) {Statistics-based};
\node[entity,below =.5em of stat] (arm) {Autoregressive Model};
\node[entity,below =.5em of arm] (arma) {ARMA Model};
\node[entity,below =.5em of arma] (tsa) {Time Synchronous Average};

\node[entity,below =.5em of fd] (fft) {Fourier Transform};
\node[entity,below =.5em of fft] (cep) {Cepstrum};
\node[entity,below =.5em of cep] (hil) {Hilbert Transform};\node[entity,below =.5em of hil] (env) {Envelope Analysis};

\node[entity,below =.5em of tfd] (stft) {Short-Time Fourier Transform};
\node[entity,below =.5em of stft] (wt) {Wavelet Transform};
\node[entity,below =.5em of wt] (hht) {Hilbert-Huang Transform};
\node[entity,below =.5em of hht,align=center, text width=8em] (emd) {Empirical Mode Decomposition};

\draw[->,>=angle 60] (fe.south) -- ++(0,0) -- ++(0,-.8em) -| (td);
\draw[->,>=angle 60] (fe.south) -| (fd);
\draw[->,>=angle 60] (fe.south) -- ++(0,0) -- ++(0,-.8em) -| (tfd);
\end{tikzpicture}

%% file: figures/time-frequency-resolution.tex
\begin{subfigure}{.35\textwidth}
	\centering
	\begin{tikzpicture}
	\draw (0,0) -- (3,0);
	\draw (3,0) -- (3,-3);
	\path[thick, ->]  (0,-3) edge node[below] {\footnotesize Time} (3.25,-3)  ;
	\path[thick, ->] (0,-3) edge node[above, rotate=90] {\footnotesize Frequency} (0,.25)  ;
	
	\draw (.5,0) -- (.5,-3);
	\draw (1,0) -- (1,-3);
	\draw (1.5,0) -- (1.5,-3);
	\draw (2,0) -- (2,-3);
	\draw (2.5,0) -- (2.5,-3);
	
	\draw (3,0) -- (3,-3);
	\end{tikzpicture}
	\caption{Waveform}
\end{subfigure}%
\begin{subfigure}{.35\textwidth}
	\centering
	\begin{tikzpicture}
	\draw (0,0) -- (3,0);
	\draw (3,0) -- (3,-3);
	\path[thick, ->]  (0,-3) edge node[below] {\footnotesize Time} (3.25,-3)  ;
	\path[thick, ->] (0,-3) edge node[above, rotate=90] {\footnotesize Frequency} (0,.25)  ;
	
	\draw (0,-.5) -- (3,-.5);
	\draw (0,-1) -- (3,-1);
	\draw (0,-1.5) -- (3,-1.5);
	\draw (0,-2) -- (3,-2);
	\draw (0,-2.5) -- (3,-2.5);
	
	\end{tikzpicture}
	\caption{Fourier Transform}
\end{subfigure}

\medskip

\begin{subfigure}{.35\textwidth}
	\centering
	\begin{tikzpicture}
	
	\draw (0,0) -- (3,0);
	\draw (3,0) -- (3,-3);
	\path[thick, ->]  (0,-3) edge node[below] {\footnotesize Time} (3.25,-3)  ;
	\path[thick, ->] (0,-3) edge node[above, rotate=90] {\footnotesize Frequency} (0,.25)  ;
	
	\draw (0.75,0) -- (0.75,-3);
	\draw (0,-0.75) -- (3,-0.75);
	\draw (1.5,0) -- (1.5,-3);
	\draw (2.25,0) -- (2.25,-3);
	\draw (0,-1.5) -- (3,-1.5);
	\draw (0,-2.25) -- (3,-2.25);
	
	\end{tikzpicture}
	\caption{Short-Time Fourier}
\end{subfigure}%
\begin{subfigure}{.35\textwidth}
	\centering
	\begin{tikzpicture}
	\draw (0,0) -- (3,0);
	\draw (3,0) -- (3,-3);
	\path[thick, ->]  (0,-3) edge node[below] {\footnotesize Time} (3.25,-3)  ;
	\path[thick, ->] (0,-3) edge node[above, rotate=90] {\footnotesize Frequency} (0,.25)  ;
	\draw (0,-1.5) -- (3,-1.5);
	\draw (0,-2.25) -- (3,-2.25);
	
	\draw (0,-2.75) -- (3,-2.75);
	\draw (1.5,0) -- (1.5,-2.75);
	
	\draw (0.75,0) -- (0.75,-2.25);
	\draw (2.25,0) -- (2.25,-2.25);
	
	\draw (0.375,0) -- (0.375,-1.5);
	\draw (1.125,0) -- (1.125,-1.5);
	\draw (1.875,0) -- (1.875,-1.5);
	\draw (2.625,0) -- (2.625,-1.5);
	\end{tikzpicture}
	\caption{Wavelet Transform}
\end{subfigure}

%% file: figures/dwt.tex
\begin{tikzpicture}[cell/.style={rectangle,draw, thick,align=center, minimum size=2em,inner sep=5pt}, input/.style={->}]

\node[cell] at (0,0) (xn) {\footnotesize $x[n]$};

\node[cell] at (-1.5,-1.5) (lpf1) {\footnotesize $LPF$};
\node[cell] at (1.5,-1.5) (hpf1){\footnotesize $HPF$};

\node[cell] at (-1.5,-3) (A1) {A1};
\node[cell, below = 1em of hpf1]  (D1) {D1};

\node[cell, below left = 2em of A1] (lpf2) {\footnotesize $LPF$};
\node[cell, below right = 2em of A1] (hpf2){\footnotesize $HPF$};

\node[cell, below = 1em of lpf2]  (A2) {A2};
\node[cell, below = 1em of hpf2]  (D2) {D2};

\draw[->, >=angle 60] (xn) -- ++(0,-1em) -- ++(0,-0.75em) -| (lpf1);
\draw[->, >=angle 60] (xn) -- ++(0,-1em) -- ++(0,-0.75em) -| (hpf1);

\draw[->, >=angle 60] (lpf1) -- (A1);
\draw[->, >=angle 60] (hpf1) -- (D1);

\draw[->, >=angle 60] (A1) -- ++(0,-1em) -- ++(0,-0.8em) -| (lpf2);
\draw[->, >=angle 60] (A1) -- ++(0,-1em) -- ++(0,-0.8em) -| (hpf2);

\draw[->, >=angle 60] (lpf2) -- (A2);
\draw[->, >=angle 60] (hpf2) -- (D2);

\node[] at (3,-0.9) (coord1) {};
\node[] at (3,-3.25) (coord2) {};

\end{tikzpicture}

%% file: chapters/chapter03.tex
\chapter{Introduction to Artificial Neural Networks}
\chapterintrobox{Artificial neural networks are computational systems capable of finding complex functions that link an input to an output. These systems can be used for a variety of tasks such as regression and classification. They can be used in predictive maintenance and prognostics to estimate the health of equipment and predict with some uncertainty its remaining useful life. This chapter discusses neural networks, their topology, training and mathematical formulation.}

\section{Structure of artificial neural networks}
Artificial neural networks are computational systems used to find the mapping between an input and an output, they consist of several layers (input layer, output layer and an arbitrary number of hidden layers between the input and the output) and each layer contains a number of neurons where each neuron in each layer is connected to all neurons in the previous and the next layer (except for the input and output layers which are connected only to the next and the previous layer respectively).
Each neuron of each layer receives an input from the neurons of the previous layer (in the form of a vector), multiplies the vector by a few weights and sums the result and then applies a linear activation function. Each neuron ends up with a single numerical value called activation, which will be passed on to the neurons in the next layer.

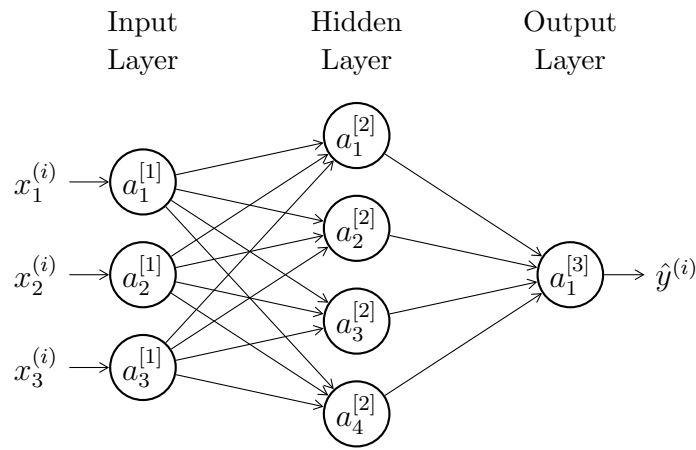
\begin{figure}[h]
    \centering
	\input{figures/mlp-structure.tex}
    \caption{Structure of an artificial neural network}
    \label{fig:neural-network-structure}
\end{figure}

Figure \ref{fig:neural-network-structure} shows a neural network with the following structure:

\begin{itemize}
    \item Input layer with 3 inputs
    \item Single hidden layer with 4 neurons
    \item Output layer with 1 neuron
\end{itemize}

\section{Feedforwad networks: from input to output}
\label{section:feedforward-neural-network}
The architecture of figure \ref{fig:neural-network-structure} is called Feedforward Neural Network, it is an acyclic architecture where information flows from the first to the last layer without any internal loop, unlike other architectures such as recurrent neural networks. The first layer is the input layer, it does not perform any operation and is limited to receiving input. The input is a vector of numbers representing the different variables. The vector is multiplied by a matrix of weights that transforms it and sends it to the next layer (or to the first hidden layer). An activation function is applied to the values resulting from the multiplication of the values of the previous layer with the weight matrix, the result becomes the values of the next layer, or activations (the value of each neuron is called activation).

A general formula for moving from the $l-1$ layer to the $l$ layer is given by the equation \ref{equation:forward-step}:

\begin{equation}
    a^{[l]} = g^{[l]}(W^{[l]}a^{[l-1]}+b^{[l]})
    \label{equation:forward-step}
\end{equation}

Where $g^{[l]}$ is the activation function of the $l$ layer, $W^{[l]}$ is the weights matrix that transforms the values (activations) from the $l-1$ layer to the $l$ layer and $a^{[l]}$ represents the activations of the $l$ layer. $b$ is the value of the bias (or intercept value), it is added to the multiplication between the activations and the weights matrix, it is a parameter that can be learned like the weights. The operation is repeated for each layer up to the output layer.
The first layer can be considered as layer 0 and the inputs can be designated by the vector $a^{[0]}$.

\section{Activation functions}
The activation functions are applied to the result of multiplying the inputs of the previous layer with the corresponding weights to determine the value of each neuron. There are different types of these functions.

The use of the non-linear activation function is very important for neural networks, they allow learning the complex non-linear mapping from input to output. If the network does not use non-linear activation (for example, linear activation or the identity function), then the whole network (regardless of its depth) is equivalent to a network with only one hidden layer.

There are a variety of activation functions that can be used for the hidden and output layers. Figure \ref{fig:activation-function} shows some examples:
\begin{figure}[h]
    \centering
	\input{figures/activation-functions.tex}
    \caption{Different activation functions}
    \label{fig:activation-function}
\end{figure}
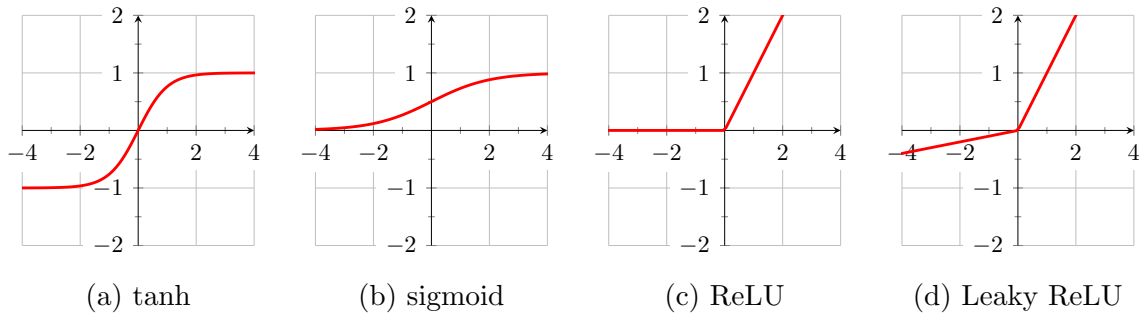

Table \ref{table:activation-functions} shows mathematical definition of several activation functions:

\begin{table}[h]
    \centering
    \begin{tabular}{c|c}
        \hline
        Activation function & Mathematical definition \\
        \hline
        Identity (no activation) & Id(x) = x \\
        Sigmoid & $\sigma(x)= \frac{1}{1+e^{-x}}$ \\
        tanh & $tanh(x)=\frac{(e^x-e^{-x})}{(e^x+e^{-x})}$\\
        Rectified Linear Unit (ReLU) & $ReLU(x)=max(0,x)$\\
        Leaky ReLU & $LeakyReLU(x)=max(0.1 x,x)$\\
    \hline
        
    \end{tabular}
    \caption{Mathematical definition of several activation functions}
    \label{table:activation-functions}
\end{table}

\section{Network training}
The process of training a neural network consists of determining the weights (coefficients) that map the neurons of each layer to the neurons of the next layer. This process can be formulated in more mathematical terms as an optimization problem: optimization of the network coefficients to find their values that minimize a cost function.

Neural network training uses training data that provide inputs and their corresponding outputs.

\subsection{Cost function}
The cost function is the function used to calculate the difference between the output of the neural network and the expected actual output, it quantifies the performance of the network. The goal of the training process is to minimize this function by using Gradient Descent (see next section) to find the best set of weights that gives the lowest difference between the training data and the network prediction.

There are many types of cost functions, each type corresponds to different neural networks tasks (e.g. regression, binary classification, …). The cost function usually used for regression problems is mean-squared error function (Equation \ref{equation:mse}) which calculates the sum of distances between the model predictions $\hat{y}_i$ and the actual output ($y_i$), $N$ is the number of data points available for training:
\begin{equation}
    MSE=\frac{1}{N}\sum_{i=1}^N(y_i-\hat{y}_i)
    \label{equation:mse}
\end{equation}

The other major task of neural networks and other types of models is classification. There are two main types of classification: binary classification, or classification of two classes and multiclass classification or classification of many classes. The first one uses binary cross-entropy loss function (Equation \ref{equation:logloss}) and the latter uses categorical cross-entropy.

\begin{equation}
    BCE = \sum_{i=1}^{N}\hat{y}_i log(y_i)+(1-\hat{y}_i)log(1-\hat{y}_i)
    \label{equation:logloss}
\end{equation}

The choice of activation function for the last layer of the network is directly related to the used cost function. For regression tasks, linear activation function is used. For binary classification it's sigmoid and for multiclass classification it's softmax function.

\subsection{Gradient Descent}
Gradient Descent is an iterative optimization algorithm used to optimize a differentiable function. Gradient Descent works by calculating the gradients of the objective function and then taking iterative steps in the negative direction of the gradients.

\subsection{Convex and non-convex functions}
\begin{figure}[h]
    \centering
    \includegraphics{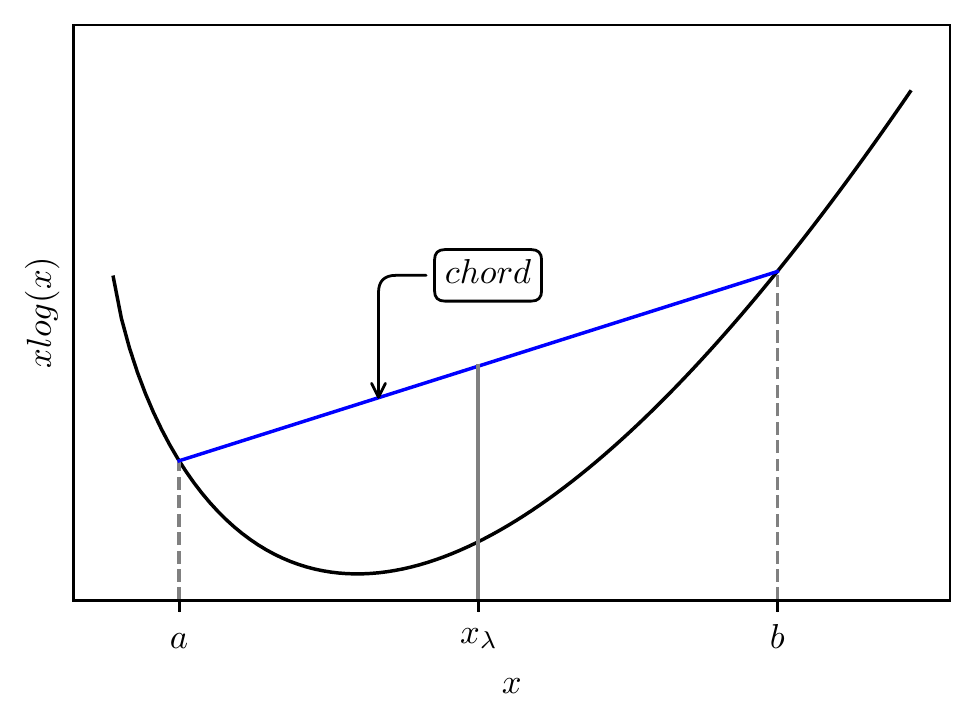}
    \caption{Condition of convexity}
    \label{fig:convexity}
\end{figure}

Gradient descent in neural networks is used to optimize (minimize) the cost function by finding the best set of weights and biases that give the lowest cost possible. Objective functions can be categorized into two types: convex and non-convex functions.
A convex function is said to be convex if it has the property that every chord lies on or above the function. Any value of $x$ in the interval from $x=a$ to $x=b$ can be written in the form $\lambda a+(1-\lambda)b$ where $0\leq\lambda\leq 1$. The corresponding point on the chord is given by $\lambda f(a)+(1-\lambda)f(b)$, and the corresponding value of the function is $f(\lambda a+(1-\lambda)b)$ (Figure \ref{fig:convexity}). Convexity then implies:
\begin{equation}
    f(\lambda a+(1-\lambda)b)\leq \lambda f(a)+(1-\lambda)f(b)
    \label{equation:convexity}
\end{equation}

This is equivalent to the requirement that the second derivative of the function be everywhere positive \cite{Bishop2006}. This condition of convexity can be extended to spaces with arbitrarily large number of dimensions.

\begin{figure}[h]
    \centering
    \includegraphics{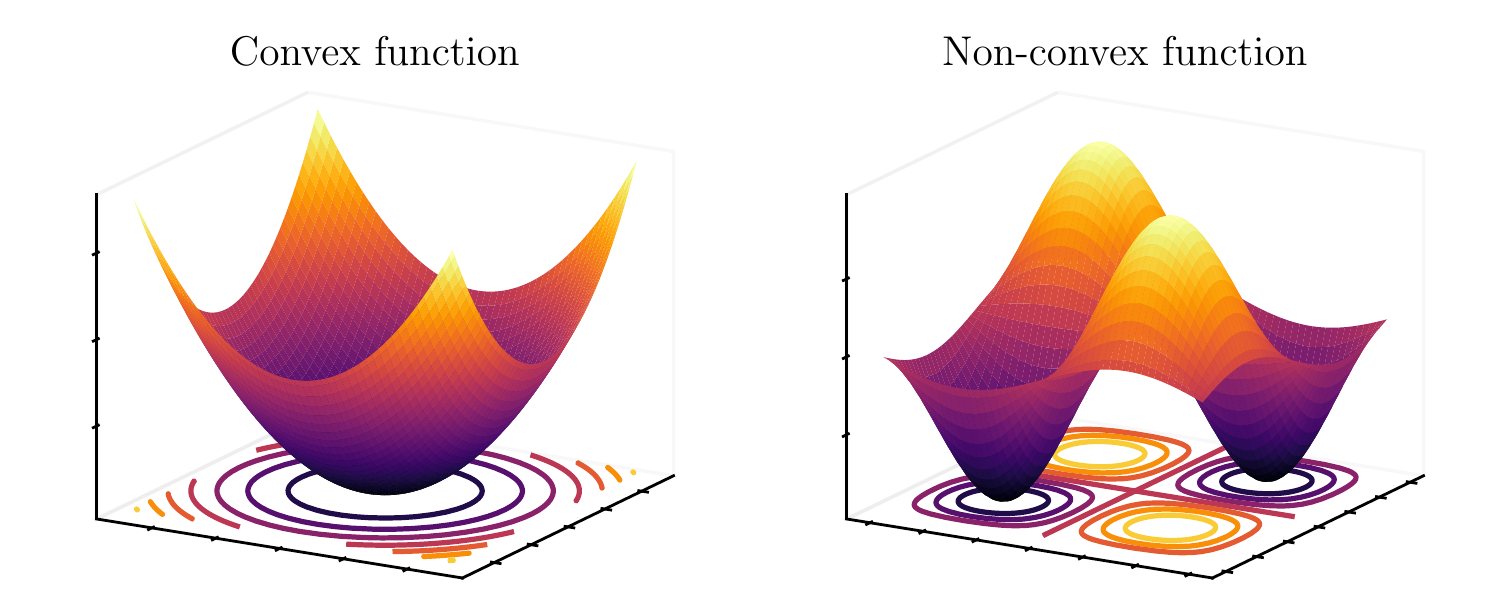}
    \caption{Example of convex and non-convex functions}
    \label{fig:convex-nonconvex-functions}
\end{figure}

A non-convex function is a function that doesn't satisfy the convexity condition. Figure \ref{fig:convex-nonconvex-functions} shows an example of a convex (left) and a non-convex function (right) where the function parameters are in two-dimensional space.

\subsection{Global and local minima}
Global minimum refers to the lowest possible value in a set (or of a function). Finding the global minimum refers to finding the set of parameters that correspond to this minimum value. When the function is convex, finding the global maximum is possible and easy, algorithms like gradient descent always converge to the global minimum in convex optimization. Examples of convex cost functions is linear regression cost function. For more complex models such as neural networks, cost function is highly non-convex with many local minima.

\begin{figure}[h]
    \centering
    \includegraphics{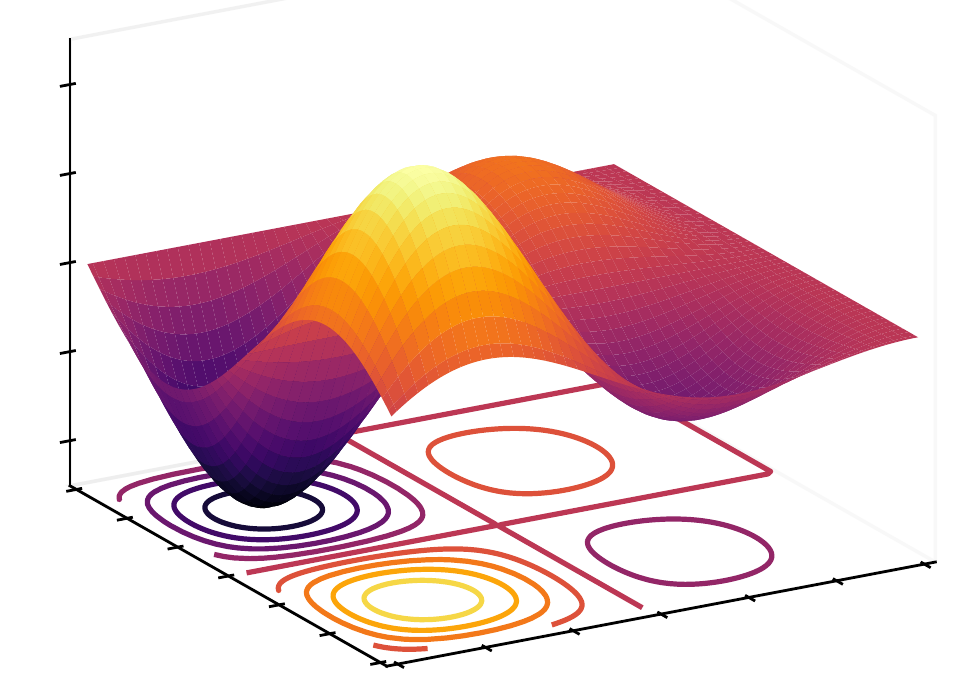}
    \caption{Non-convex function with global and local minima}
    \label{fig:global_local_minima}
\end{figure}

Figure \ref{fig:global_local_minima} shows an example of a non-convex function with a global and local minima. The problem with non-convex optimization is that the optimization algorithm can converge to the local minimum instead of the global one. The algorithm convergence is related to the random initialization of network weights.

\subsection{Saddle points}
\begin{figure}[h]
    \centering
    \includegraphics{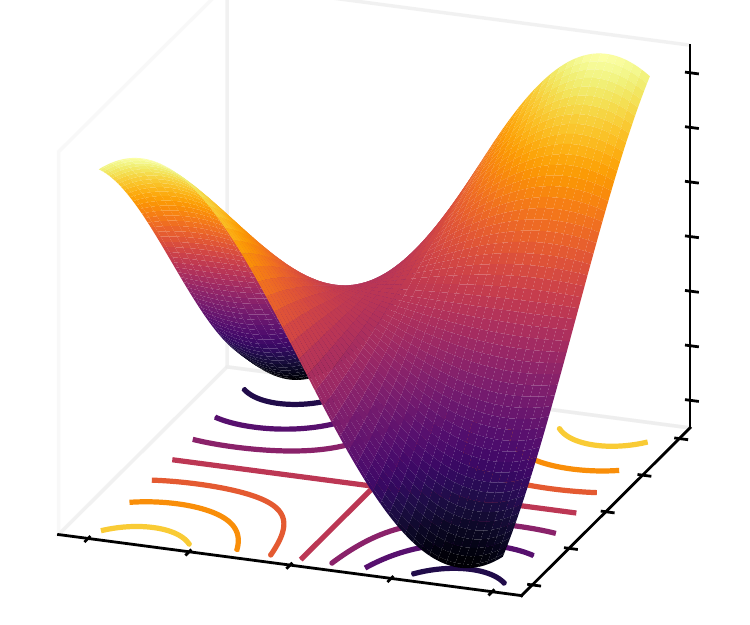}
    \caption{Saddle point}
    \label{fig:saddle-point}
\end{figure}

A saddle point or minimax point is a point on the surface of the graph of a function where the slopes (derivatives) in orthogonal directions are all zero (a critical point), but which is not a local extremum of the function (Figure \ref{fig:saddle-point}).

\subsection{Optimization of multilayer networks}
In  \cite{Choromanska2014} the authors showed that getting stuck in poor local minima is a major problem only for shallow networks but becomes gradually of less importance as the network size increases. This is mainly due to the fact that in large networks, local minima lie close to the global minimum so they yield similar good performance.

The following hypotheses were also verified empirically in the mentioned paper regarding learning with large-size networks:
\begin{itemize}
    \item When a network is large, there isn't a significant difference in performance on test data among most of local minima.
    \item Probability of finding a bad local minimum in small networks is higher than the probability of finding them in larger networks.
    \item Finding the global minimum on training set doesn't guarantee a better performance on testing data, rather it could cause overfitting and reduced performance.
\end{itemize}

\subsection{Backpropagation}
Backpropagation is the algorithm used to calculate the gradients of the cost function (the objective function) of a neural network. Since a neural network can be interpreted as functions composition, Backpropagation uses the derivation theorem of a composition of functions to find the gradients in relation to the network weights.

The use of Backpropagation for training neural networks was popularized by David E. Rumelhart, Geoffrey E. Hinton and Ronald J. Williams \cite{Rumelhart1986}, they described it as a procedure that repeatedly adjusts the weights of the network connections so as to minimize a measure of the difference between the actual output vector of the network and the desired output vector. As a result of the weight adjustments, "hidden" internal units that are not part of the input or output come to represent important characteristics of the task domain, and the regularities of the task are captured by the interactions of these units.

Figure \ref{fig:forward-backward-pass} represents the Forward Pass and the Backward Pass, the Forward Pass calculates the network output, the Backward Pass calculates the gradients of the cost function that measures the difference between this output and the actual desired output:

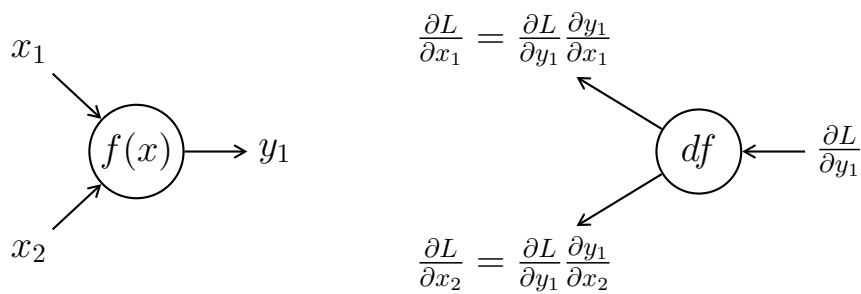
\begin{figure}[h]
    \centering
	\input{figures/forward-backward-pass.tex}
    \caption{Forward (left) and Backward (right) passes}
    \label{fig:forward-backward-pass}
\end{figure}

\section{Recurrent Neural Networks}
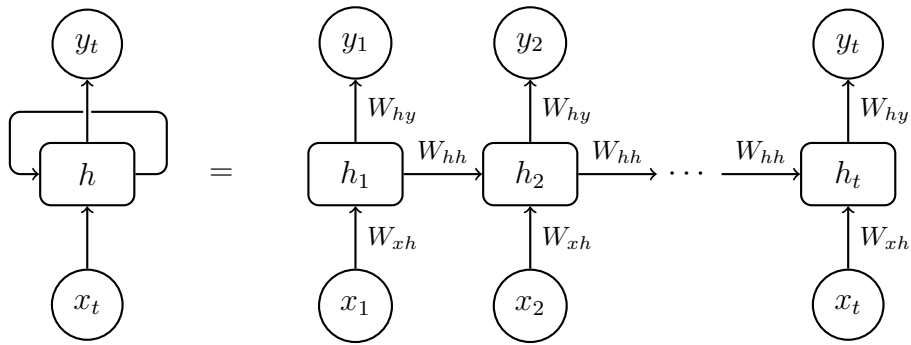
\begin{figure}[H]
        \centering
        \input{figures/rnn.tex}
        \caption{Recurrent neural networks architecture}
        \label{fig:rnn}
\end{figure}

\acrlong{rnn} (\acrshort{rnn}) is a special architecture that is more suitable for modeling sequential data. \acrshort{rnn}s process an input sequence one element at a time, maintaining in their hidden units a 'state vector' that implicitly contains information about the history of all the past elements of the sequence \cite{LeCun2015}. Figure \ref{fig:rnn} shows an example of a \acrshort{rnn} architecture. To the left is the unfolded version with a cyclic loop in the hidden layer, to the right is shows unfolding of the network which is easier to understand. $x_1$, $x_2$, …$x_t$ represents the input vector (sequence), $y_1$, $y_2$, …$y_t$ represents the output vector (can be a sequence with length equals to the input, of different length or a single element). $h_1$, $h_2$, …$h_t$ are neurons of the hidden layer. In \acrshort{rnn} each neuron passes a vector of its hidden state to the next neuron corresponding to the next time step.
    
\subsection{Long Short-Term Memory}
\label{section:lstm}
Simple \acrshort{rnn}s have problems with learning long-term dependencies (i.e. making predictions based on predictions made many steps previously in the past) and vanishing gradients. \acrlong{lstm} (\acrshort{lstm}) was first introduced by Jürgen Schmidhuber and Sepp Hochreiter \cite{Hochreiter1997}. \acrshort{lstm} overcomes problems with basic \acrshort{rnn} by using what is called a cell state which enables \acrshort{lstm} networks to learn long-term dependencies. \acrshort{lstm} cells also possess three types of gates which together control the flow of information inside the cell:
\begin{itemize}
    \item \textbf{Forget gate}: Controls which information are kept or discarded at time $t$
    \item \textbf{Input gate}: Controls which information to store in the cell state at time $t$
    \item \textbf{Output gate}: Controls the final output of the cell at time $t$
\end{itemize}

Input, forget and output gates values are calculated using Equations \ref{equation:lstm_input_gate}, \ref{equation:lstm_forget_gate} and \ref{equation:lstm_output_gate} respectively:

\begin{align}
i_t &= \sigma(W_{xi}x_t + W_{hi}h_{t-1}+W_{ci}c_{t-1}+b_i) \label{equation:lstm_input_gate}\\
f_t &= \sigma(W_{xf}x_t + W_{hf}h_{t-1}+W_{cf}c_{t-1}+b_f) \label{equation:lstm_forget_gate}\\
o_t &= \sigma(W_{xo}x_t + W_{ho}h_{t-1}+W_{co}c_t+b_o) \label{equation:lstm_output_gate}\\
\end{align}

Input, forget and output gates are calculated at time $t$ using sets of weights and biases ($W_{xi}$, $W_{hi}$, $W_{ci}$, $b_i$), ($W_{xf}$, $W_{hf}$, $W_{cf}$, $b_f$) and ($W_{xo}$, $W_{ho}$, $W_{co}$, $b_o$) that controls how each of $x_t$, $h_{t-1}$ and $c_{t-1}$ affects the value of the gate respectively. A sigmoid function is used to convert the values to the range of 0 to 1. 

At each time step $t$, a candidate cell state $\tilde{c}_t$ is calculated using weights and a bias term ($W_{xc}$, $W_{hc}$, $b_c$) that maps the values of input $x_t$ and previous hidden state $h_{t-1}$ to the candidate cell state (Equation \ref{equation:lstm_candidate_cell_state}). As the name indicates, $\tilde{c}_t$ serves as a candidate to replace current cell state $c_t$ at time $t$. Cell state at time $t$ is obtained by using the forget gate to control which information are kept from the previous cell state and input gate to control which information are kept from the candidate cell state according to Equation \ref{equation:lstm_cell_state}. Final cell output is calculated using current cell state $c_t$ and output gate according to Equation \ref{equation:lstm_hidden_state}.

\begin{align}
    \tilde{c}_t &= tanh(W_{xc}x_t+W_{hc}h_{t-1}+b_c) \label{equation:lstm_candidate_cell_state} \\
    c_t &= f_tc_{t-1}+i_t\tilde{c}_t \label{equation:lstm_cell_state}\\
    h_t &= o_ttanh(c_t) \label{equation:lstm_hidden_state}
\end{align}

Gates control the flow of information inside the \acrshort{lstm} cell, their values range from 0 to 1 (sigmoid function) and control which information to keep and which to discard when updating the cell state (forget and input gates) and when calculating the \acrshort{lstm} cell output (output gate).

Figure \ref{fig:lstm} illustrates the different operations happening in a single \acrshort{lstm} cell:

\begin{figure}[h]
    \centering
    \input{figures/lstm.tex}
    \caption{Long Short-Term Memory Cell}
    \label{fig:lstm}
\end{figure}
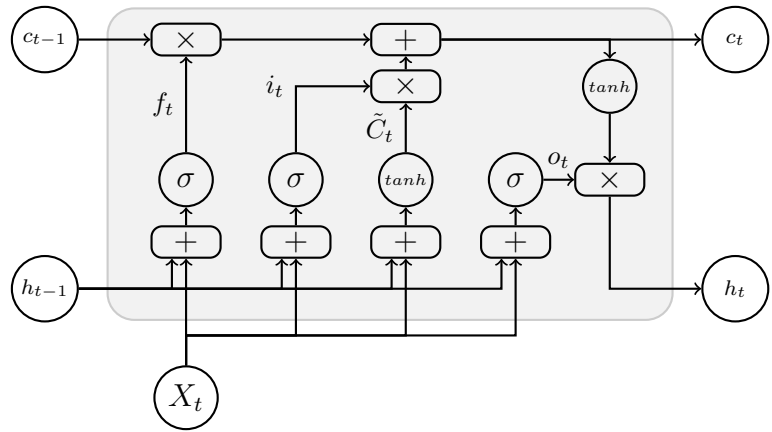

\section{Convolutional neural networks}
\label{section:cnn}
\acrlong{cnn} (\acrshort{cnn}) have what is called convolutional layers. A convolutional layer takes an input (usually 2D images, but it can be 1D or 3D either) and apply a mathematical operation called convolution\footnote{Mathematically, the operation happening in a \acrshort{cnn} is called cross-correlation which is a bit different than the mathematical definition of a convolution. In Machine Learning literature those operations are called convolutions, this is the terminology that will be used here.}. Convolutions act like filters that extract features from raw data by recursively multiplying what is called kernel with the input data. Different kernels serve different purposes and can extract a wide variety of features from raw data. After feature extraction done by the convolutional layer, they serve as an input for a feedforward neural network (as described in section \ref{section:feedforward-neural-network}) for classification or regression.

\begin{figure}[H]
    \centering
    \begin{subfigure}{0.22\linewidth}
        \centering
        \includegraphics[width=\linewidth]{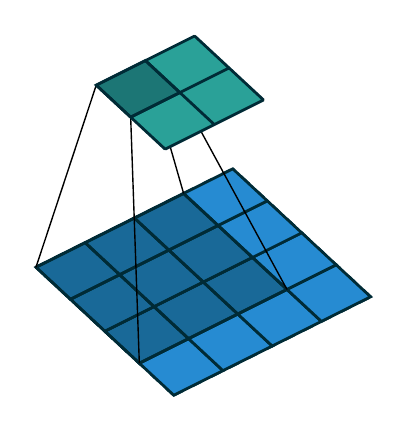}
        \subcaption*{step 01}
    \end{subfigure}\hfill%
    \begin{subfigure}{0.22\linewidth}
        \centering
        \includegraphics[width=\linewidth]{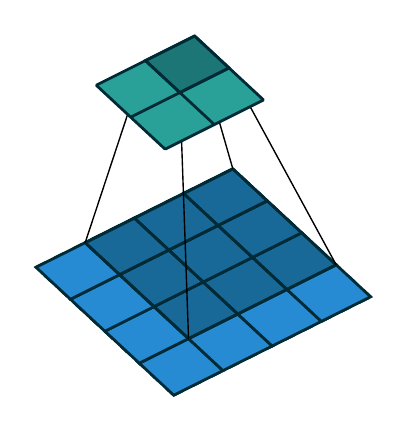}
        \subcaption*{step 02}
    \end{subfigure}\hfill%
    \begin{subfigure}{0.22\linewidth}
        \centering
        \includegraphics[width=\linewidth]{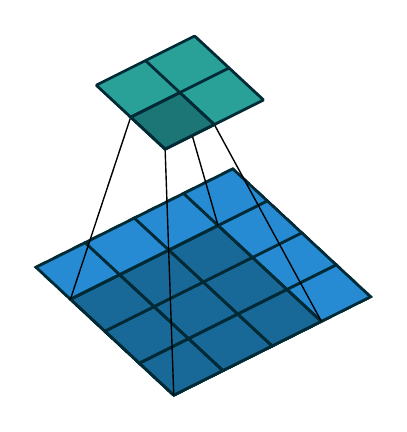}
        \subcaption*{step 03}
    \end{subfigure}\hfill%
    \begin{subfigure}{0.22\linewidth}
        \centering
        \includegraphics[width=\linewidth]{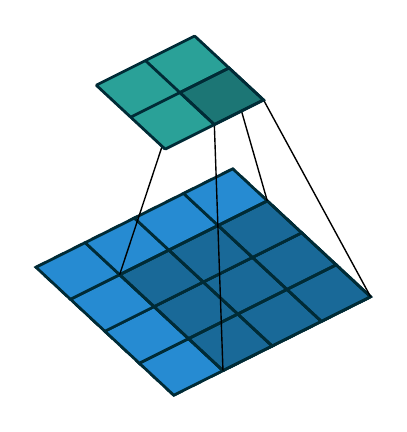}
        \subcaption*{step 04}
    \end{subfigure}
    \caption{Applying convolutions (by multiplying the input by a kernel) iteratively to different regions (dark blue) of the 2D input (whole blue square). Each iteration gives a numerical value (dark green). The result of all iterations is the green 2D output \cite{dumoulin2016guide}.}
    \label{fig:convolutions}
\end{figure}

\section{Conclusion}
Neural networks are a powerful tool to find the complex relationship between an input and an output (e.g. \acrlong{cm} data and machine degradation). They consist of different layers of interconnected neurons. the connection between each 2 layers is defined by a set of weights, which are multiplied by the values of the previous layer to find the values of the next layer, the process is repeated until the output layer is reached. the predicted output is compared to the actual output obtained from the training data, the weights are adjusted to minimize (using Backpropagation and Gradient Descent) the difference between the predictions and the reality.

%% file: figures/mlp-structure.tex
\begin{tikzpicture}[neuron/.style={circle,draw, thick,align=center, minimum size=2em,inner sep=1pt}, input/.style={->}]

\node[text width=2cm, align=center] at (0,.5*\nodedist) {\small Input Layer};

\node[text width=2cm, align=center] at (\layerdist,.5*\nodedist) {\small Hidden Layer};

\node[text width=2cm, align=center] at (2*\layerdist,.5*\nodedist) {\small Output Layer};

\foreach \y in {1,...,3} \node[] (in\y) at (-\layerdist*.5,-\y*\nodedist) {$x_\y^{(i)}$};  

\foreach \y in {1,...,3} \node[neuron]  (I\y) at (0,-\y*\nodedist) {$a_\y^{[1]}$};  

\foreach \in in {1,...,3} \draw[->, >=angle 60] (in\in) --  (I\in);

\foreach \y in {1,...,4} \node[neuron]  (H\y) at ($(\layerdist,-\y*\nodedist) +(0, .5*\nodedist)$) {$a_\y^{[2]}$};

\foreach \y in {1,...,1} \node[neuron] (O\y) at ($(I2) + (2*\layerdist, 0)$) {$a_\y^{[3]}$};

\node at ($(I2) + (2.5*\layerdist, 0)$) (y) {$\hat{y}^{(i)}$};

\foreach \dest in {1,...,4} \foreach \source in {1,...,3} \draw[->, >=angle 60] (I\source) -- (H\dest);

\foreach \dest in {1,...,1} \foreach \source in {1,...,4} \draw[->, >=angle 60] (H\source) -- (O\dest);

\draw[->, >=angle 60] (O1) -- (y);

\end{tikzpicture}

%% file: figures/activation-functions.tex
\begin{subfigure}[b]{0.24\textwidth}
	\resizebox{\linewidth}{!}{%
		\begin{tikzpicture}
		\begin{axis}[ 
		xmin=-4, xmax=4,ymin=-2, 
		ymax=2, grid=major,
		height=5cm, width=5cm,
		axis line style={latex-latex},
		axis lines=middle,
		ticklabel style={font=\footnotesize,fill=white},
		minor tick num=1,
		scaled ticks=false] 
		\addplot[samples=100,red,very thick] {tanh(x))};
		\end{axis}
	\end{tikzpicture} 
	}%
\subcaption{tanh}
\end{subfigure}
\hfill
\begin{subfigure}[b]{0.24\textwidth}
	\resizebox{\linewidth}{!}{%
	\begin{tikzpicture}
	\begin{axis}[ 
	xmin=-4, xmax=4,ymin=-2, 
	ymax=2, grid=major,
	height=5cm, width=5cm,
	axis line style={latex-latex},
	axis lines=middle,
	ticklabel style={font=\footnotesize,fill=white},
	minor tick num=1,
	scaled ticks=false] 
	\addplot[samples=100,red,very thick] {1/(1+exp(-x))};
	\end{axis}
	\end{tikzpicture}
	}%
	
\subcaption{sigmoid}
\end{subfigure}
\hfill
	\begin{subfigure}[b]{0.24\textwidth}
		\resizebox{\linewidth}{!}{%
		\begin{tikzpicture}
		\begin{axis}[ 
		xmin=-4, xmax=4, ymin=-2, ymax=2, grid=major,
		height=5cm, width=5cm,
		axis line style={latex-latex},
		axis lines=middle,
		ticklabel style={font=\footnotesize,fill=white},
		minor tick num=1,
		scaled ticks=false] 
		\addplot[samples=100,red,very thick] {max(0,x)};
		\end{axis}
		
		\end{tikzpicture}
}%
\subcaption{ReLU}
	\end{subfigure}
\hfill
		\begin{subfigure}[b]{0.24\textwidth}
			\resizebox{\linewidth}{!}{%
		\begin{tikzpicture}
		\begin{axis}[ 
		xmin=-4, xmax=4,ymin=-2, 
		ymax=2, grid=major,
		height=5cm, width=5cm,
		axis line style={latex-latex},
		axis lines=middle,
		ticklabel style={font=\footnotesize,fill=white},
		, xticklabel style={anchor=north},
		minor tick num=1,
		scaled ticks=false] 
		\addplot[samples=100,red,very thick] {max(0,x)+min(0,0.1*x)};
		\end{axis}
		
		\end{tikzpicture}
		}%
		\subcaption{Leaky ReLU}
		\end{subfigure}

%% file: figures/forward-backward-pass.tex
\begin{tikzpicture}[arrow/.style={thick, >=angle 60}]
	\node[draw, circle,thick,minimum width=3em, inner sep=0] (fp) {\large $f(x)$};
	\node[above left = 2em of fp] (x1) {\large $x_1$};
	\node[below left = 2em of fp] (x2) {\large $x_2$};
	\node[right = 2em of fp] (y1) {\large $y_1$};
	
	
	\draw[->,arrow] (x1) -- (fp);
	\draw[->,arrow] (x2) -- (fp);
	\draw[->,arrow] (fp) -- (y1);

\node[draw, circle,thick,minimum width=2.7em, inner sep=0, right = 15 em of fp] (bp) {\large $df$};
\node[above left = 2em of bp] (bx1) {\large $\frac{\partial L}{\partial x_1}=\frac{\partial L}{\partial y_1}\frac{\partial y_1}{\partial x_1}$};

\node[below left = 2em of bp] (bx2) {\large $\frac{\partial L}{\partial x_2}=\frac{\partial L}{\partial y_1}\frac{\partial y_1}{\partial x_2}$};

\node[right = 2em of bp] (by1) {\large $\frac{\partial L}{\partial y_1}$};


\draw[<-,arrow] (bx1) -- (bp);
\draw[<-,arrow] (bx2) -- (bp);
\draw[<-,arrow] (bp) -- (by1);
\end{tikzpicture}

%% file: figures/rnn.tex
	\begin{tikzpicture}[layer/.style={rectangle,draw, thick,align=center, minimum width=3em, minimum height=2em,inner sep=5pt,rounded corners=4},neuron/.style={circle,draw, thick,align=center,inner sep=5pt,rounded corners=4}, links/.style={->, thick}]
	
	\node[layer] (h-rolled) {$h$};
	\node[neuron, below = 2em of h-rolled] (input-rolled) {$x_t$};
	\node[neuron, above = 2em of h-rolled] (output-rolled) {$y_t$};
		\draw[links, rounded corners] (h-rolled.east) -| ++(1em,2em) -- ++(-5em,.0em) |- (h-rolled.west);
	\draw[links] (input-rolled) -- (h-rolled);
	\draw[line width=3pt,white] (h-rolled) -- (output-rolled);
	\draw[links] (h-rolled) -- (output-rolled);

	\node[right = 2em of h-rolled] (equals) {$=$};

	\node[layer, right = 2em of equals] (h-unrolled1) {$h_1$};
		\node[neuron, below = 2em of h-unrolled1] (input-unrolled1) {$x_1$};
	\node[neuron, above = 2em of h-unrolled1] (output-unrolled1) {$y_1$};
	\draw[links] (h-unrolled1) -- node[right] {\footnotesize $W_{hy}$} (output-unrolled1);
	\draw[links] (input-unrolled1) -- node[right] {\footnotesize $W_{xh}$} (h-unrolled1);

	\node[layer, right = 2.5em of h-unrolled1] (h-unrolled2) {$h_2$};
	\node[neuron, below = 2em of h-unrolled2] (input-unrolled2) {$x_2$};
	\node[neuron, above = 2em of h-unrolled2] (output-unrolled2) {$y_2$};
	
	\draw[links] (h-unrolled2) -- node[right] {\footnotesize $W_{hy}$} (output-unrolled2);
	\draw[links] (input-unrolled2) -- node[right] {\footnotesize $W_{xh}$} (h-unrolled2);
	
	\draw[links] (h-unrolled1) -- node[above] {\footnotesize $W_{hh}$} (h-unrolled2) ;
	
	\node[right = 2.5em of h-unrolled2] (dots) {$\cdots$}; 
	
	\draw[links] (h-unrolled2) -- node[above] {\footnotesize $W_{hh}$}(dots);
	
	\node[layer, right = 2.5em of dots] (h-unrolled3) {$h_t$};
	\node[neuron, below = 2em of h-unrolled3] (input-unrolled3) {$x_t$};
	\node[neuron, above = 2em of h-unrolled3] (output-unrolled3) {$y_t$};
	
	\draw[links] (h-unrolled3) -- node[right] {\footnotesize $W_{hy}$} (output-unrolled3);
	\draw[links] (input-unrolled3) -- node[right] {\footnotesize $W_{xh}$} (h-unrolled3);
	
	\draw[links] (dots) -- node[above] {\footnotesize $W_{hh}$}(h-unrolled3);
	
	\end{tikzpicture}

%% file: figures/lstm.tex
\begin{tikzpicture}

		\tikzstyle{operation} = [rectangle, thick,draw,align=center, minimum width=2.2em, minimum height=.5em,inner sep=2pt,rounded corners=4 ]
		
		\tikzstyle{links} = [->, thick]
		
		\tikzstyle{values} = [circle, draw, thick, inner sep = 2,minimum size=2.1em]
		
		\tikzstyle{functions} = [draw, circle, thick, inner sep=1, minimum size=20]
		
		\draw[rounded corners=10,black!20!white, thick, fill=black!5!white ] (0,0) rectangle (18em,-10em);
		
		\node[operation] at (2.5em, -7.5em) (addition1) {$+$};
		\node[operation] at (6em, -7.5em) (addition2) {$+$};
		\node[operation] at (9.5em, -7.5em) (addition3) {$+$};
		\node[operation] at (13em, -7.5em) (addition4) {$+$};

		\node[functions] at (2.5em, -5.5em) (sigma1) {$\sigma$};
		\node[functions] at (6em, -5.5em) (sigma2) {$\sigma$};
		\node[functions] at (9.5em, -5.5em) (tanh1) {\tiny $tanh$};
		\node[functions] at (13em, -5.5em) (sigma3) {$\sigma$};

		\node[operation] at (9.5em, -2.5em) (times1) {$\times$};
		\node[operation] at (9.5em, -1em) (addition5) {$+$};
		
		\node[operation] at (16em, -5.5em) (times2) {$\times$};
		\node[functions] at (16em, -2.5em) (tanh2) {\tiny $tanh$};
		
		\node[operation] at (2.5em, -1em) (times3) {$\times$}; 
		
		\node[circle, draw, thick, inner sep = 2.5] at (2.5em, -12.5em) (input) {$X_t$};

		
		\node[values] at (-2em,-1em) (previouscellstate) {\scriptsize $c_{t-1}$} ;
		
		\node[values] at (-2em,-9em) (previoushiddenstate) {\scriptsize $h_{t-1}$} ;
		
		\node[values] at (20em,-9em) (nexthiddenstate) {\scriptsize $h_{t}$} ;
		
		\node[values] at (20em, -1em) (nextcellstate) {\scriptsize $c_{t}$} ;
		
		\draw[links] (previouscellstate) -- (times3);
		\draw[links] (addition5) -- (nextcellstate);
		
		\draw[links] (input) -- (addition1);
		\draw[links] (input) |- ++(3em,2em) -| (addition2);
		\draw[links] (input) |- ++(5em,2em) -| (addition3);
		\draw[links] (input) |- ++(7em,2em) -| (addition4);
		
		\draw[links] (previoushiddenstate) -|  (addition1.230);
		\draw[links] (previoushiddenstate) -|  (addition2.230);
		\draw[links] (previoushiddenstate) -|  (addition3.230);
		\draw[links] (previoushiddenstate) -|  (addition4.230);
		
		\draw[links] (times2) |- (nexthiddenstate);
		
		\draw[links] (addition1) -- (sigma1);
		\draw[links] (addition2) -- (sigma2);
		\draw[links] (addition3) -- (tanh1);
		\draw[links] (addition4) -- (sigma3);
		
		\draw[links] (tanh1) -- node[left] {\footnotesize $\tilde{C}_t$} (times1);
		\draw[links] (times1) -- (addition5);
		\draw[links] (sigma2)    |-  node[left] {\footnotesize $i_t$} (times1) ;
		
		\draw[links]  (sigma1)  -- node[left] {\footnotesize $f_t$} (times3);
		
		\draw[links] (times3) -- (addition5);
		\draw[links] (addition5) -| (tanh2);
		\draw[links] (tanh2) -- (times2);		
		\draw[links] (sigma3) -- node[above] {\footnotesize $o_t$} (times2);
	\end{tikzpicture}

%% file: chapters/chapter04.tex
\chapter{Equipment Health Assessment using Artificial Neural Networks}%
\label{chapter:equipment_health_assessment_using_artificial_neural_networks}

\chapterintrobox{This chapter demonstrates the use of an artificial neural network to estimate the health state of a set of turbofan engines from NASA C-MAPSS dataset and predict the remaining useful life. Also different metrics to assess the performance of the network will be introduced.}

\section{Introduction to NASA C-MAPSS dataset}

C-MAPSS is a tool for simulating a realistic large commercial turbofan engine. The software is coded in the MATLAB\textsuperscript{\textregistered} and Simulink\textsuperscript{\textregistered} environment.

\begin{wrapfigure}{r}{0.5\textwidth}
    \centering
    \includegraphics[width=.48\textwidth]{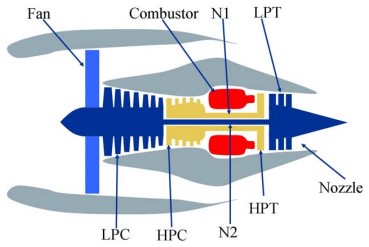}
    \caption{Simplified diagram of engine simulated in C-MAPSS \cite{Saxena2008}}
    \label{figure:c-mapss-engine-diagram}    
\end{wrapfigure}

C-MAPSS software consists of many editable input parameters that controls the simulation, these inputs are specified by the user and control many aspects of the simulation such as operational profile, closed-loop controllers, environmental conditions, etc. \cite{Saxena2008}. 

Figure \ref{figure:c-mapss-engine-diagram} is a simplified diagram of the simulated engine showing its main elements, like low pressure compressor section (LPC), high pressure compressor section (HPC), fan and combustor. The dataset released by NASA Ames Research Center contains resulting data from simulating many turbofan engines, from beginning of operation until failure. The dataset was originally released for Prognostics and Health Management 2008 Data Competition, Table \ref{table:c-mapss-sensors} shows different variables, the output of the simulation and their units, that were provided for the participants in the competition:

\begin{table}[ht]
    \centering
    \begin{tabu}{lll}
		\tabucline[1.5pt]{-} 
        \textbf{Symbol} & \textbf{Description} & \textbf{Units}\\
        \hline
        \textbf{T2} & Total temperature at fan inlet & R \\
        \textbf{T24} & Total temperature at LPC outlet & R \\
        \textbf{T30} & Total temperature at HPC outlet & R  \\
        \textbf{T50} &Total temperature at LPT & R\\
        \textbf{P2} & Pressure at fan inlet& psia\\
        \textbf{P15}& Total pressure in bypass-duct& psia\\
        \textbf{P30}& Total pressure at HPC outlet& psia\\
        \textbf{Nf}& Physical fan speed& rpm\\
        \textbf{Nc} & Physical core speed &rpm\\
        \textbf{epr}& Engine pressure ratio (P50/P2)& --\\
        \textbf{Ps30}& Static pressure at HPC outlet& psia\\
        \textbf{phi}& Ratio of fuel flow to Ps30& pps/psi\\
        \textbf{NRf}& Corrected fan speed &rpm\\
        \textbf{NRc}& Corrected core speed& rpm\\
        \textbf{BPR}& Bypass Ratio& --\\
        \textbf{farB}& Burner fuel-air ratio &--\\
        \textbf{htBleed}& Bleed Enthalpy &-- \\
        \textbf{Nf\_dmd} &Demanded fan speed& rpm\\
        \textbf{PCNfR\_dmd}& Demanded corrected fan speed &rpm\\
        \textbf{W31} & HPT coolant bleed & lbm/s \\
        \textbf{W32} & LPT coolant bleed & lbm/ \\
		\tabucline[1.5pt]{-} 
    \end{tabu}
    \caption{C-MAPSS outputs to measure system response.}
    \label{table:c-mapss-sensors}
\end{table}

C-MAPSS data contains 4 datasets: FD001, FD002, FD003 and FD004. Each dataset has different working conditions and fault modes. Table \ref{table:c-mapss-statistics} contains statistics of the different datasets:

\begin{table}[ht]
    \centering
    \begin{tabu}{ccccc}
        
		\tabucline[1.5pt]{2-5} 
                    & units number  & max length    & average length    & min length    \\
       \hline
            FD001   & 100           & 362           & 206.31            & 128           \\
            FD002   & 260           & 378           & 206.77            & 128           \\
            FD003   & 100           & 525           & 247.2             & 145           \\
            FD004   & 249           & 543           & 245.95            & 128           \\
		\tabucline[1.5pt]{-} 
    \end{tabu}
    \caption{Units number and cycles length statistics in C-MAPSS data}
    \label{table:c-mapss-statistics}
\end{table}

\section{Visualization of equipment degradation}
Visualizing the simulation output can give a sense of how these variables change during the life of the engine, Figure \ref{fig:sensors-plot} shows four different sensors from one of the engines (values are normalized):

\begin{figure}[h]
    \centering
    \includegraphics[width=\linewidth]{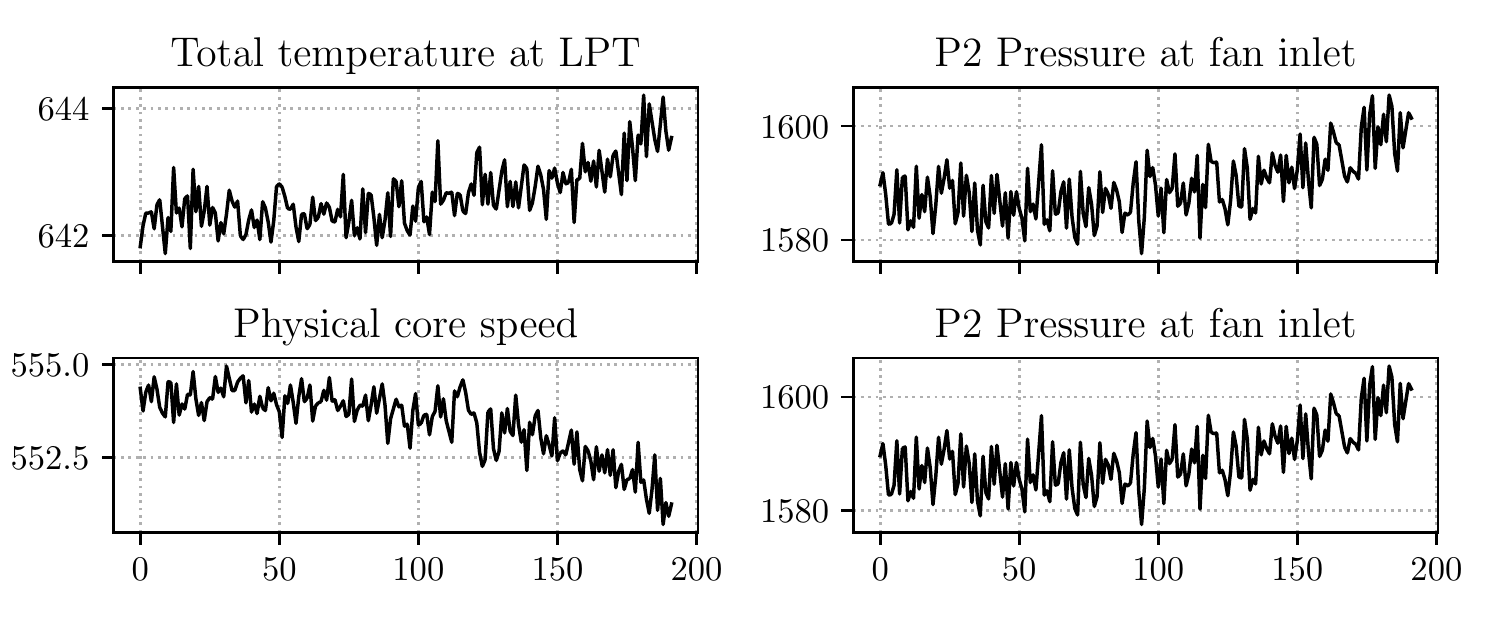}
    \caption{Development of 6 sensors outputs from one of the engines (normalized)}
    \label{fig:sensors-plot}
\end{figure}
\begin{figure}[H]
    \centering
    \includegraphics[width=.9\linewidth]{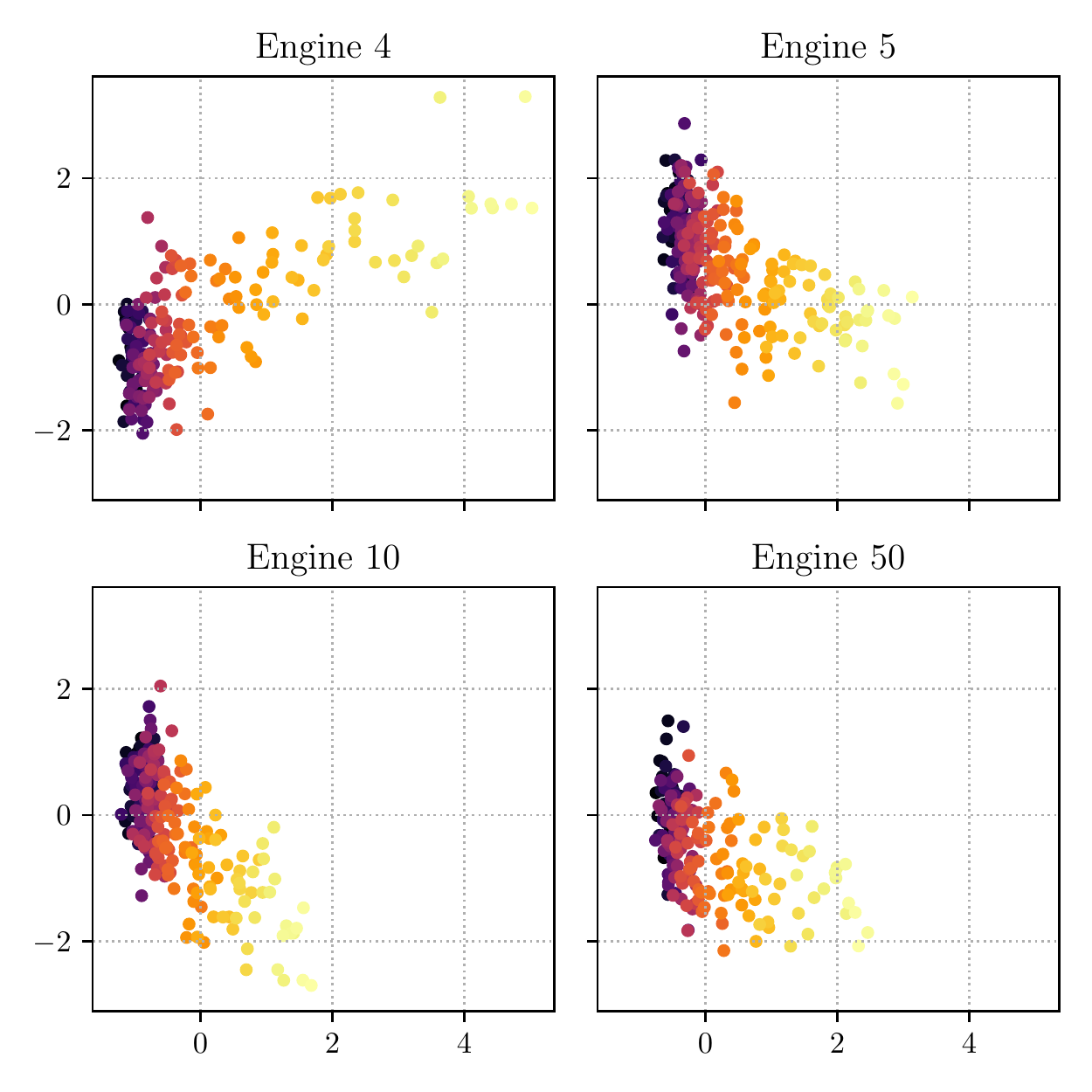}
    \caption{Equipment health degradation (lighter colors indicate advancing of health degradation) of four turbofan engines from C-MAPSS dataset}
    \label{fig:pca-degradation}
\end{figure}

It is apparent that sensors output follow a specific pattern (increasing or decreasing) from beginning of operation until breakdown, this is very useful and can increase the robustness of the predictive model.

Alternatively, all sensors values can be combined and visualized altogether using Principal Component Analysis (section \ref{section:dimensionality-reduction}) to reveal the general trend in the data, if condition monitoring data is directly indicative of the equipment health state, visualization of principal components can show apparent visual degradation patterns.

Sensors values from 4 different engines are combined using PCA and the two first principal components are represented on Figure \ref{fig:pca-degradation}. 

There is absolutely an apparent pattern for health state degradation across the different engines from left (where darker colors indicate normal working state) to the right (where the lighter colors indicate fault development).

\section{Engines health classification}
Before proceeding with a complicated task such as \acrshort{rul} estimation, a simpler task like engines health classification can show the complexity of the problem. This section describes using neural networks to classify engines states as healthy or faulty. The first and last 25 cycles from each unit are considered to be healthy and faulty respectively.

A neural network that uses all 24 inputs (all operational settings and sensors) with two hidden layers is used to carry out this classification. Table \ref{table:c-mapss-classifier-architecture} summarizes the network architecture:

\begin{table}[ht]
    \centering
    \begin{tabu}{lll}
		\tabucline[1.5pt]{-}
		\textbf{Layer (type)}   & \textbf{Output shape} &   \textbf{Param \#} \\
		\tabucline[1pt]{-}
		Dense1 (Dense) 			&   (None, 8)   &   200\\
		Dense2 (Dense) 	        &   (None, 4)   &   36       \\
		Dense3 (Dense)			&   (None, 1)   &   5   \\
		\tabucline[1pt]{-}
		Total params: 241       &                   &           \\
		Trainable params: 241   &                   &           \\
		Non-trainable params: 0     &                   &           \\
	\tabucline[1.5pt]{-}
    \end{tabu}
    \caption{C-MAPSS classifier architecture}
    \label{table:c-mapss-classifier-architecture}
\end{table}

\begin{figure}[H]
    \centering
    \includegraphics{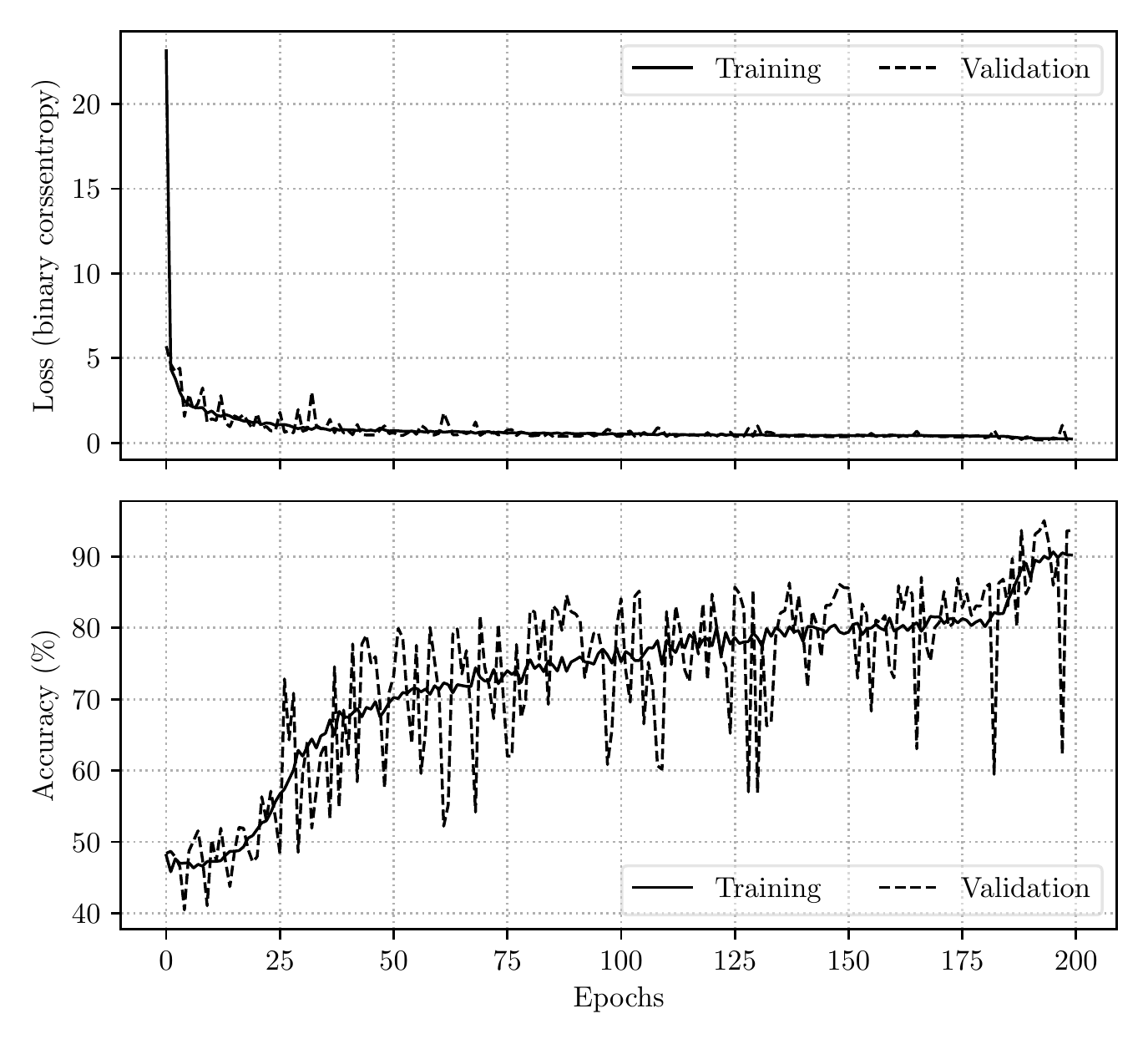}
    \caption{Training process of C-MAPSS classifier}
    \label{fig:cmapss-classifier-training}
\end{figure}

The model is trained for 200 epochs with batch size of 32 samples, the classifier achieves \textbf{93.46\% accuracy} on the test set. Figure \ref{fig:cmapss-classifier-training} shows the training process of the network. In the upper plot, the y-axis corresponds to the train and validation losses (binary crossentropy). The y-axis in the lower plot corresponds to the network train and validation accuracies, the x-axis shared between the two plots indicates the training epochs. Although the validation accuracy varies a lot, the training accuracy keeps improving with each epoch.

\begin{figure}[H]
    \centering
    \includegraphics{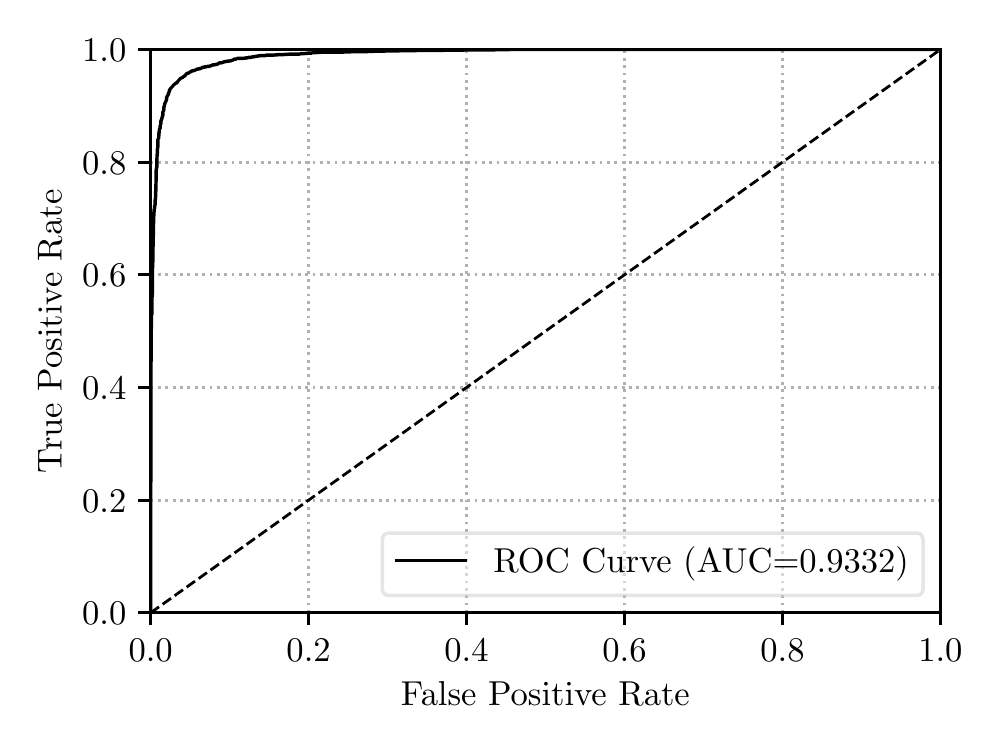}
    \caption{C-MAPSS classifier ROC curve on test set}
    \label{fig:cmapss-classifier-roc}
\end{figure}

Figure \ref{fig:cmapss-classifier-roc} shows \acrlong{roc} (\acrshort{roc}) Curve of the classifier with a high \acrlong{auc} (\acrshort{auc}). Other classification metrics are shown in Table \ref{table:cmapss-classifier-metrics}.

\begin{table}[H]
    \centering
    \begin{tabu}{cccccc}
        
    \tabucline[1.5pt]{-}
    \textbf{Metric} &  \textbf{Accuracy} &  \textbf{Precision} &  \textbf{Recall} &  \textbf{F-1} &  \textbf{ROC AUC}  \\
    \hline
    \textbf{Score} & 93.43\% & 0.90 & 0.98 & 0.94 & 0.9332 \\
	\tabucline[1.5pt]{-}
    \end{tabu}
    \caption{C-MAPSS classifier metrics on test set}
    \label{table:cmapss-classifier-metrics}
\end{table}

\section{Remaining Useful Life prediction}
\subsection{Remaining Useful Life modeling}
In order to train a neural network to estimate the \acrlong{rul} (\acrshort{rul}) of a new unseen data from C-MAPSS dataset, an appropriate \acrshort{rul} that corresponds to the training data must be constructed. \acrshort{rul} was defined in Section \ref{section:rul}, from this definition several approaches to construct an appropriate \acrshort{rul} can be developed for units in the train data (where the total number of cycles before failure is known). The simplest approach is to use an always-decreasing \acrshort{rul}, this implies that the equipment state is always decreasing and moving towards failure. The problem with this approach is that degradation process isn't linear and in real life, machines don't start to degrade as soon as they are start working. A second approach is to use a piecewise function where the equipment state is constant at first then at a specific point it starts to degrade linearly. This approach is much better than the first one but it is not very representative of the real world behavior of degradation process. In the context of this thesis, \acrshort{rul} is approached as a nonlinear polynomial function where the degradation is slow at first then accelerates towards the end of life. Figure \ref{fig:rul-models} shows the different choices for modeling \acrshort{rul}:

\begin{figure}[H]
    \centering
    \includegraphics{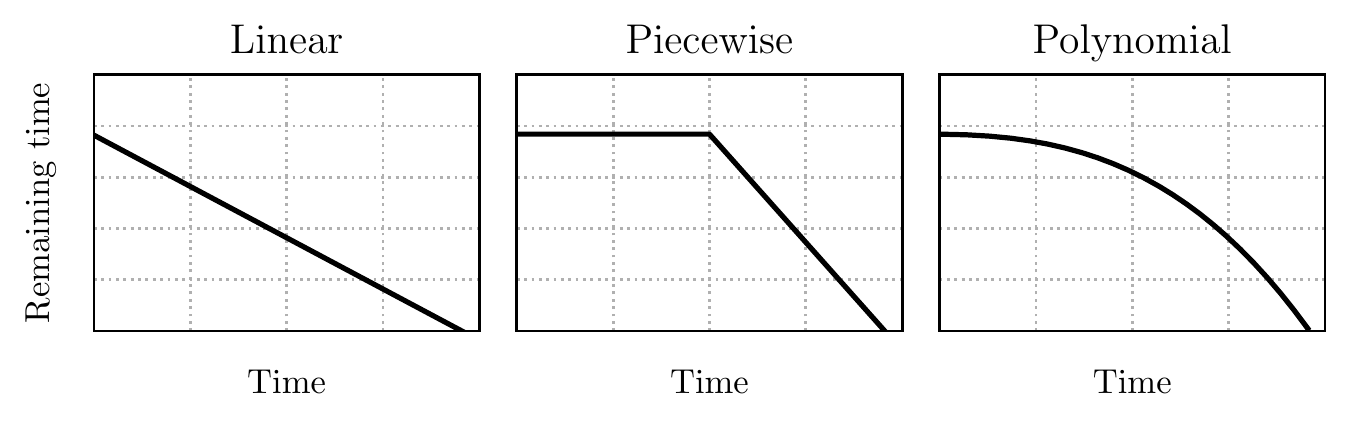}
    \caption{Different \acrshort{rul} modeling choices}
    \label{fig:rul-models}
\end{figure}

It must be noted that not any of these approaches can be claimed to be representative of the real world degradation process, but choosing the most intuitive representation of \acrshort{rul} can help the learning algorithm learn the implicit degradation trends in the data.

\subsection{\acrshort{rul} prediction using feed-forward network}
The last section dealt with classifying units health state as healthy or faulty. This section instead presents a neural network for estimating the \acrshort{rul}. In this section, a neural network architecture with 3 hidden layers is used. Since this is a regression problem, mean squared error is used as the loss function, mean absolute error is used as a metric for the model. Table \ref{table:cmapss-regression-architecture} presents the architecture details:

\begin{table}[h]
    \centering
    \begin{tabu}{lll}
		\tabucline[1.5pt]{-}
		\textbf{Layer (type)}   & \textbf{Output shape} &   \textbf{Param \#} \\
		\tabucline[1pt]{-}
		Dense1 (Dense) 			&   (None, 32)  &       800     \\
		Dense2 (Dense)          &   (None, 16)  &       528     \\
		Dense3 (Dense)          &   (None, 8)   &       136     \\
		Dense4 (Dense)          &   (None, 1)   &       9       \\

		\tabucline[1pt]{-}
		Total params: 1,473       &                   &           \\
		Trainable params: 1,473   &                   &           \\
		Non-trainable params: 0   &                   &           \\
	\tabucline[1.5pt]{-}
    \end{tabu}
    \caption{Fully-connected network architecture for \acrshort{rul} prediction}
    \label{table:cmapss-regression-architecture}
\end{table}

\begin{figure}[H]
    \centering
    \includegraphics{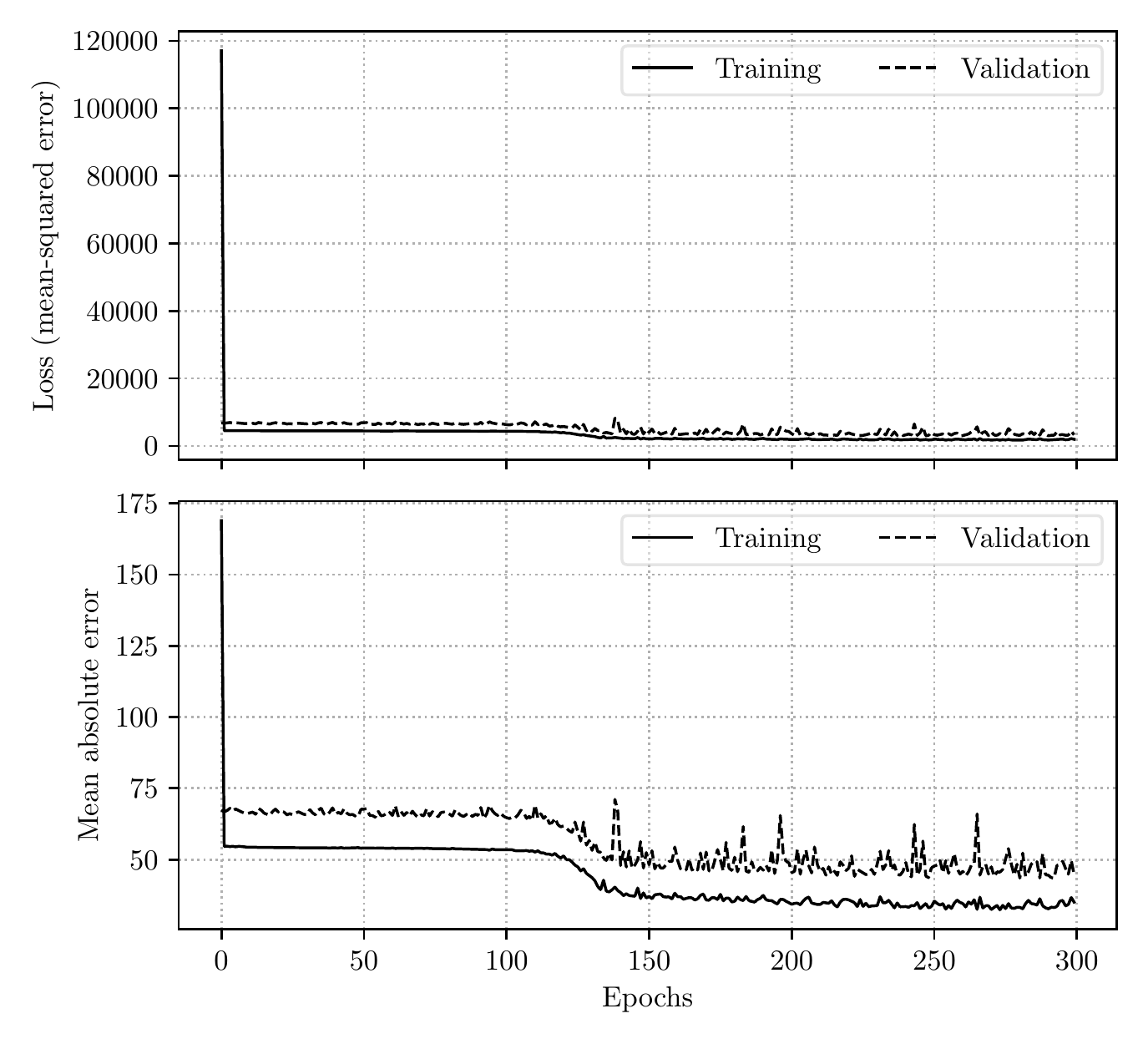}
    \caption{Training process of \acrshort{rul} predictor}
    \label{fig:cmapss-regression-training}
\end{figure}

The network was trained for 300 epochs, batch size of 128 samples. Figure \ref{fig:cmapss-regression-training} shows the training process of the network. The figures show the development of training and validation losses (mean squared error) and metrics (mean absolute error) respectively as a function of training epochs.

\begin{figure}[H]
    \centering
    \includegraphics{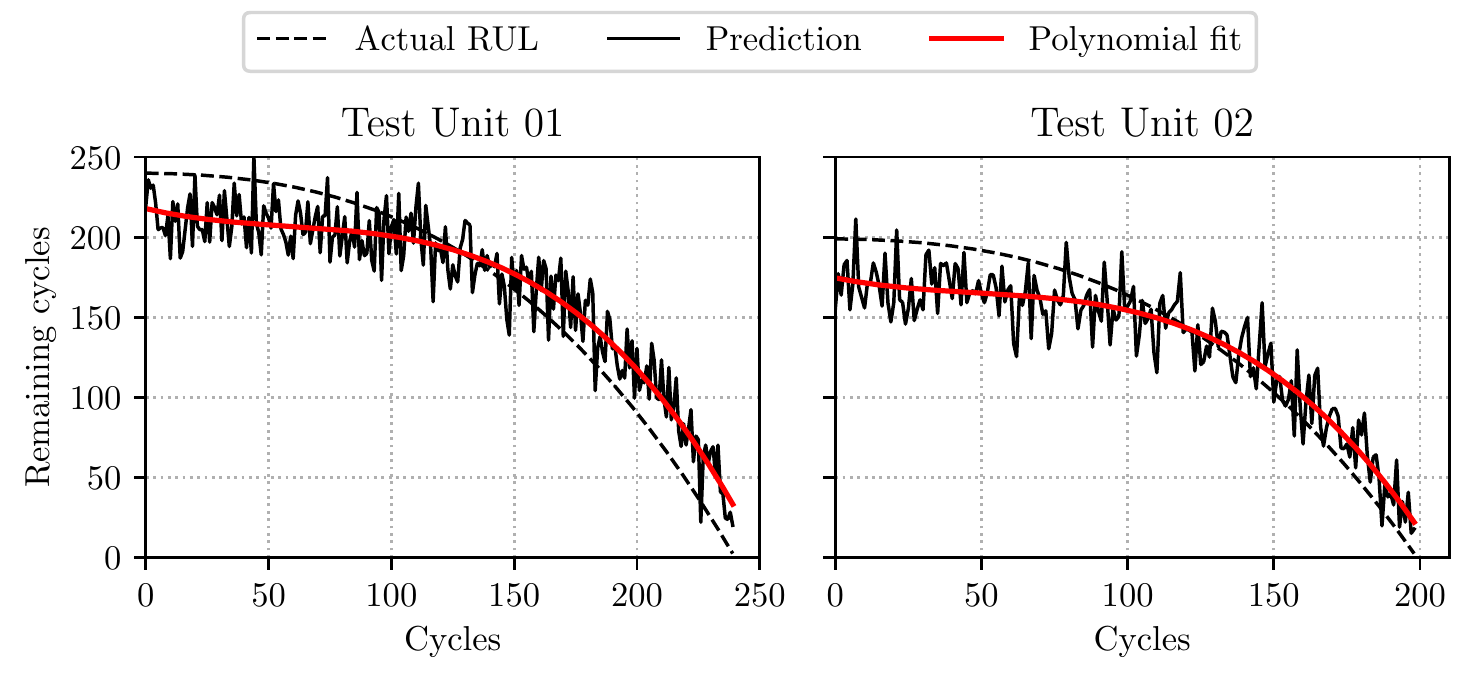}
    \caption{Prediction results on two test units}
    \label{fig:cmapss-regression-prediction}
\end{figure}

After training, the model is evaluated on two units from the test set. Figure \ref{fig:cmapss-regression-prediction} shows the actual \acrshort{rul} (dashed line), model prediction at each cycle and a 3rd degree polynomial fit of model's predictions.

Although the model's prediction are close the the actual \acrshort{rul}, but they are noisy and there is so much fluctuation. The noise can be reduced by smoothing the output for example using moving average or fitting the points to a polynomial. But the reason of the noise in the first place is that fully-connected architecture doesn't take into consideration the previous predictions (i.e. acyclic architecture). Since \acrshort{rul} at each instant of time depends on the \acrshort{rul} from the previous one, taking previous steps into consideration while making new predictions can effectively reduce the noise and also improve prediction accuracy. To achieve that, a cyclic neural architecture like recurrent neural networks should be used.

\subsection{Improving \acrshort{rul} prediction using LSTM networks}
Fully-connected neural networks are powerful tool for modeling a wide range of problems, but using the fully-connected architecture for time series prediction, such as \acrshort{rul} prediction, can yield very noisy output because the network is acyclic and each instant in time is evaluated separately and doesn't take into account the previous predictions. \acrlong{lstm} (Section \ref{section:lstm}) are a powerful tool for modeling such problems and to make more robust and less noisy predictions. In this section, fully-connected neural network architecture from previous section is replaced with an \acrshort{lstm} network to predict \acrshort{rul} in C-MAPSS dataset.

The architecture used here is described in Table \ref{table:cmapss-lstm-architecture}:

\begin{table}[h]
    \centering
    \begin{tabu}{lll}
		\tabucline[1.5pt]{-}
		\textbf{Layer (type)}   & \textbf{Output shape} &   \textbf{Param \#} \\
		\tabucline[1pt]{-}
		LSTM1 (LSTM) 			&   (None, 100, 100)    &       50000   \\
		LSTM2 (LSTM)           &   (None, 100, 100)    &       80400   \\
		LSTM3 (LSTM)           &   (None, 75)          &       52800   \\
        Dense1 (Dense)         &   (None, 120)         &       9120    \\
        Dense2 (Dense)         &   (None, 110)         &       13310   \\
        Dense3 (Dense)         &   (None, 100)         &       11100   \\
		\tabucline[1pt]{-}
		Total params: 216,730       &                   &               \\
		Trainable params: 216,730   &                   &               \\
		Non-trainable params: 0     &                   &               \\
	\tabucline[1.5pt]{-}
    \end{tabu}
    \caption{LSTM network architecture for \acrshort{rul} prediction}
    \label{table:cmapss-lstm-architecture}
\end{table}

It is apparent that the \acrshort{lstm} architecture has much more parameters (216,730 parameters) than fully-connected architecture (1,473). This is because of the more complicated design of \acrshort{lstm} cells and their possession of different gates. This results in much longer training time.

\acrshort{lstm} layers take 3 dimensional tensor as an input of shape \textit{(samples, sequence length, features)}. Sequence length was set to 100 in this architecture, that's why every training sample must have a shape of \textit{(sequence length, features)}. All units have the same number of features but different lengths, since every sample must have a sequence length of 100, samples with cycles less than 100 are padded with -10 where \acrshort{lstm} layers are set to ignore any time steps with this value.

The network was trained for 50 epochs with batch size of 16 and validation split of 0.2 using Adam optimizer. Training process is visualized in Figure \ref{fig:cmapss-lstm-training}:

\begin{figure}[H]
    \centering
    \includegraphics{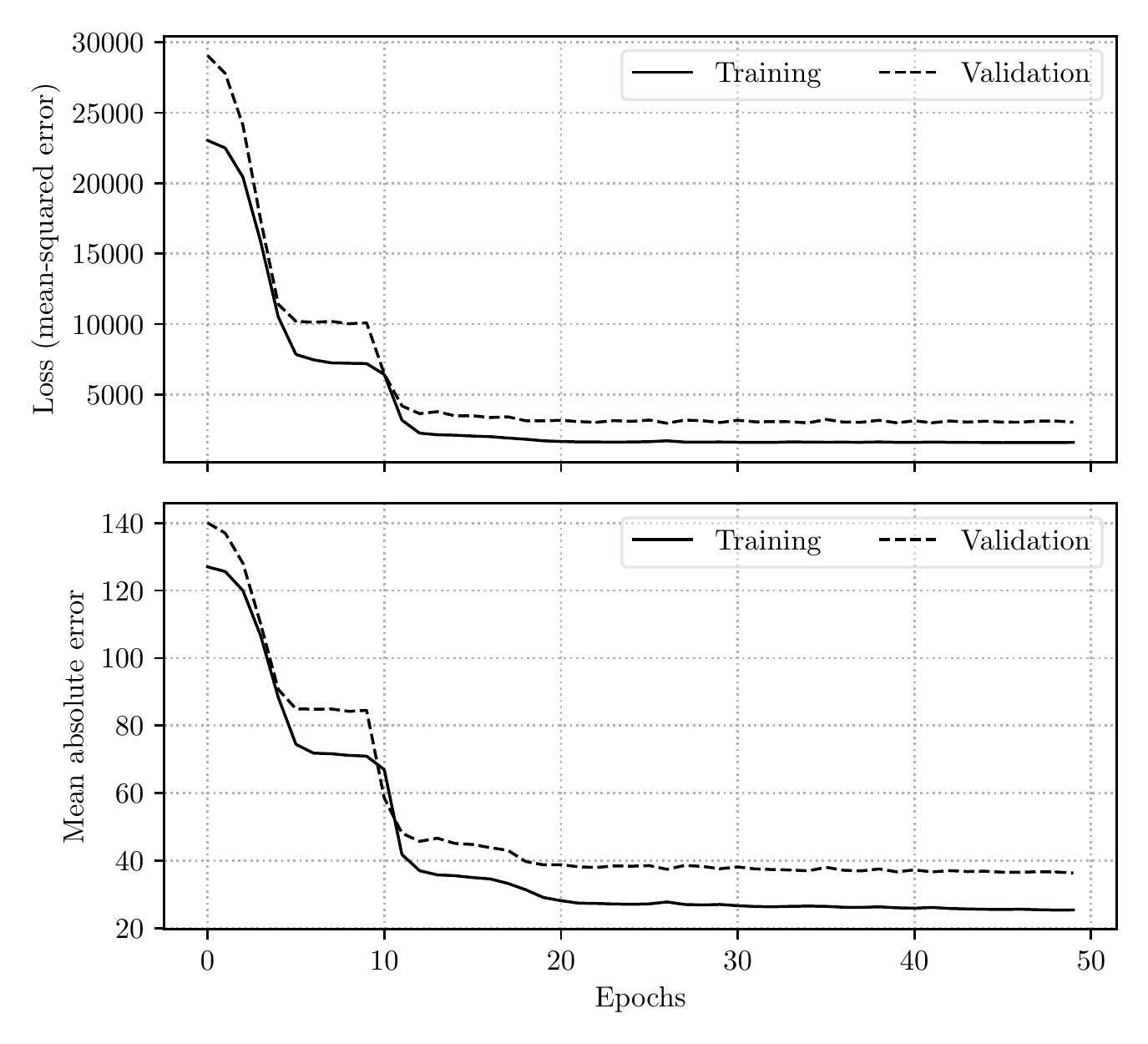}
    \caption{LSTM training process}
    \label{fig:cmapss-lstm-training}
\end{figure}

Table \ref{table:cmapss-lstm-results} shows the loss and metric (i.e. mean absolute error) on the different train, validation and test sets:

\begin{table}[H]
	\centering
	\begin{tabu}{lcc}
		\tabucline[1.5pt]{2-3} 
						&	\textbf{Loss}	&	\textbf{Mean Absolute Error}	\\
	   \tabucline[1pt]{-}
		Train set 		&	1600.60			    &	25.43				\\
		Validation set 	&	3039.68 			&	36.37					\\
		Test set		&	1379.67 			&	23.27					\\
   \tabucline[1.5pt]{-}
   \end{tabu}
   \caption{\acrshort{lstm} training results}
   \label{table:cmapss-lstm-results}
\end{table}

\begin{figure}[h]
    \centering
    \includegraphics{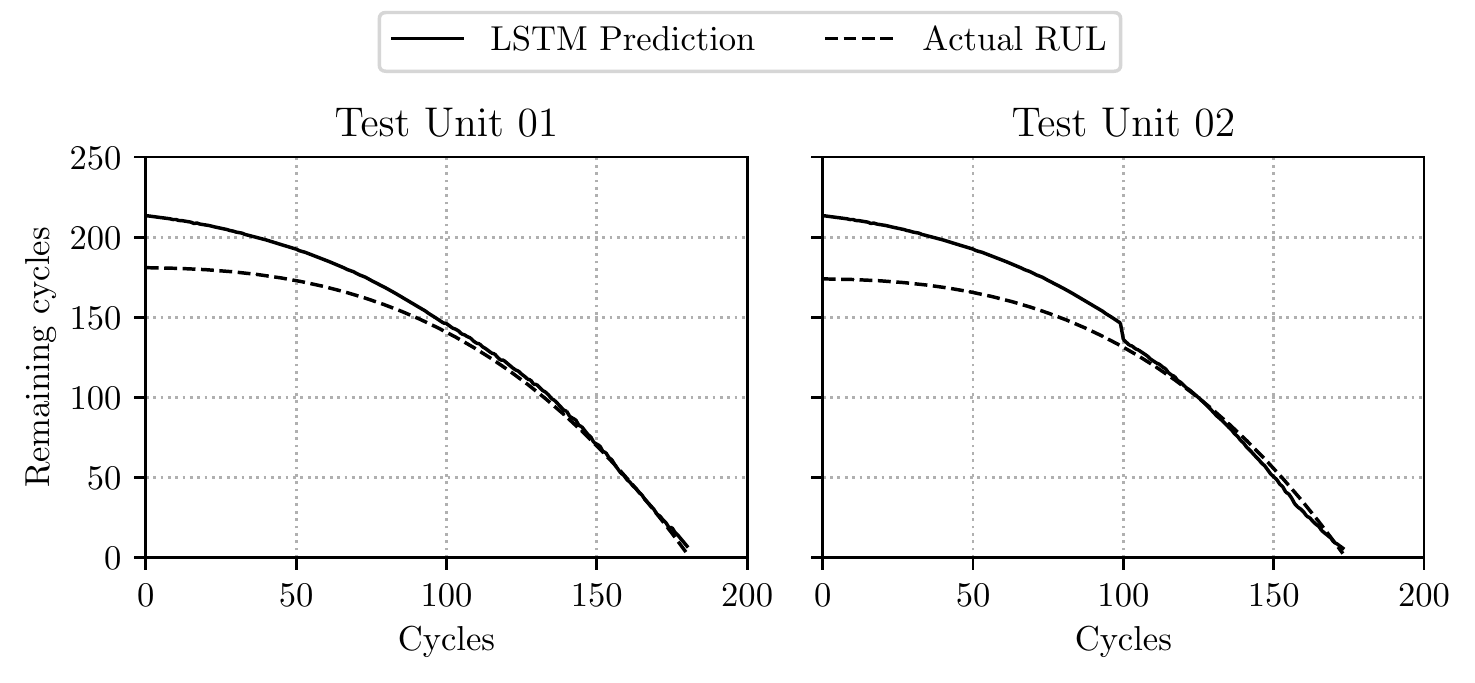}
    \caption{LSTM prediction results on two test units}
    \label{fig:cmapss-lstm-prediction}
\end{figure}

Four different units were reserved as testing units, after training the network is used to predict \acrshort{rul} on test units. Figure \ref{fig:cmapss-lstm-prediction} shows prediction results and actual \acrshort{rul} of two different units.

It's apparent that \acrshort{lstm} network predictions are much better and almost free of noise compared to those made by the fully-connected network shown in Figure \ref{fig:cmapss-regression-prediction}. Predicted \acrshort{rul} is almost identical to actual \acrshort{rul} towards the end of life of the unit.

\section{Application to oilfield equipment}%
\label{sec:application_to_oilfield_equipment_1}
This chapter presented a predictive maintenance approach for \acrshort{rul} estimation based on condition monitoring data provided by different sensors that can measure different physical variables that are related to the equipment's health state. The data used in the previous sections consisted of C-MAPSS turbofan engines dataset, but the same approach can be extended to other applications like oil rigs.

Modern oil rigs contain a large number of sensors distributed across almost every critical equipment. These sensors measure different physical variables such as temperature, flow, pressure…. But the majority of the data provided by these sensors isn't properly exploited into an appropriate framework of prognostics and predictive maintenance, there are enormous amounts of historic data in oil rigs that if exploited can greatly enhance maintenance programs in such critical applications where reducing downtime is very important and equipment unavailability can have cause significant production losses.

\begin{figure}[p!]
	\centering
	\includegraphics[width=0.9\linewidth]{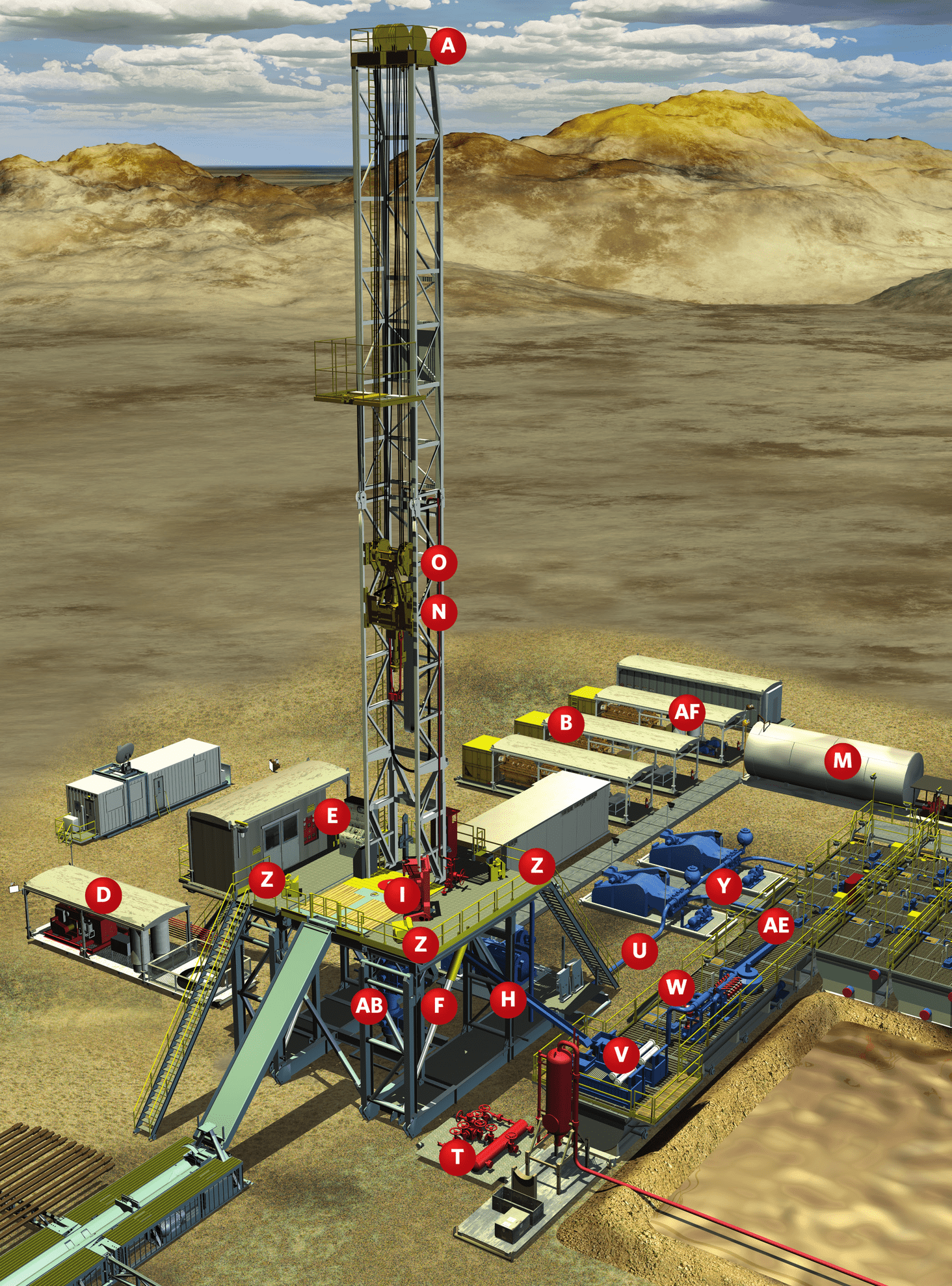}
	\caption{List of sensors installed on oil rigs for monitoring}%
	\label{fig:honeywell-oilrig-sensors}
\end{figure}

Figure \ref{fig:honeywell-oilrig-sensors} shows different spots where sensors are installed on most modern oil rigs. Table \ref{table:honeywell-oilrig-sensors} lists types of different sensors that can be used for prognostics, the units to which they are attached and which physical variables they can measure. A neural network can be trained on the rig historic data and be used to monitor different equipment, their interaction, detect anomalies and report them in time in order to carry on the appropriate maintenance actions.

\begin{table}[p!]
	\centering
	\begin{adjustbox}{angle=90}
	\begin{tabu}{clp{30mm}p{120mm}}
		\tabucline[1.5pt]{-} 
		\textbf{Sensor} & \textbf{Equipment} & \textbf{Sensor Type}  & \textbf{Measured quantity}\\
        \hline
A&	Crown Block &	Load cell	&	Weight on drill line via cable tension\\
B& Power Generation Unit	&	Pressure&	Oil, water, and hydraulic fluid pressure\\
D&Accumulator Unit	&	Pressure&	Inlet/outlet pressure with high accuracy\\
F&Rig Hydraulic Lift	&	Pressure&	Hydraulic pressure, weight, force/strain, or movement, monitor raising or lowering deck for directional drilling\\
H&Drawworks	&	Load cells	&	Torque, load/weight/position while guiding pipe into position\\
M&Water/Storage Tank	&	Pressure& Tank liquid levels\\
N&Top Drive	&	Torque/Pressure&	Monitor torque/twisting movement to ensure right amount of force is applied. Weight on drill bit. Hydraulic pressure and feed information into control system.\\
O&Traveling Block	&	Load cells	&	Weight on the drill line via cable tension\\
R&Deadline anchor	&	Load cells	&	Tension on deadline/drilling line cable\\
U&Mud Return Line	&	Pressure&	Drilling mud pressure to monitor and control mud flow\\
Y&Mud Pump	&	Pressure&	Pressure and flow of mud media\\
AB&Blowout Preventor &	Pressure&	Monitor RAM position via hydraulic volumetric or pressure behind the piston\\
AD&Drill Bit	&	Pressure&	Pressure or differential pressure at high temperature and pressure ranges\\
AE	&	Fluid manifold	&	Pressure	&	Drilling fluid pressure\\
AF&Mud Tank/Reservoir	&	Pressure&	Tank liquid levels\\
		\tabucline[1.5pt]{-} 
    \end{tabu}
\end{adjustbox}
    \caption{Honeywell oil rig sensors list}
    \label{table:honeywell-oilrig-sensors}

\end{table}

\section{Conclusion}
Fully-connected neural networks are a powerful tool for quantifying health state of complex systems using condition monitoring data. C-MAPSS dataset is an example of a system with many interacting parts that provide a variety of condition-monitoring data (e.g. temperature, pressure, rpm, …) where the degradation isn't directly related to one component but is indicated by the sum of the trends in all the monitoring data. Neural networks are able to capture such trends and predict the \acrshort{rul} of the system before failure. Fully-connected networks were first used for this task, but a cyclic architecture like \acrshort{lstm} can produce better results with improved accuracy and much less noise.

%% file: chapters/chapter05.tex
\chapter{Bearings Faults Diagnostics and Prognostics}%
\label{chapter:bearings_faults_diagnostics_and_prognostics}

\chapterintrobox{Vibration condition monitoring are vital for many industrial systems, vibration data contains very useful information about health state of the equipment and the fault type. Nevertheless, gaining insights from vibration signals in real-world applications turns out to be a complex—and in many times, unfruitful—process. This is mainly due to the complexity of the problem. This chapter introduces several approaches where neural networks are used for diagnostics and prognostics of bearings elements}

\section{Case Western Reserve University Bearings Data}

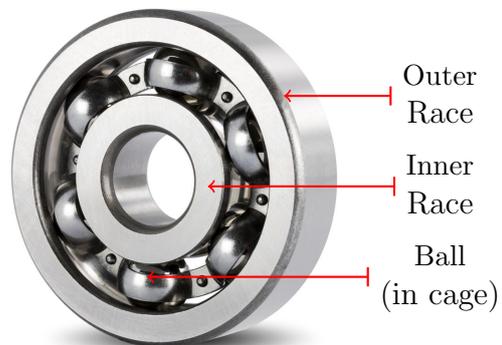
\begin{wrapfigure}{r}{0.5\textwidth}
    \centering
	\input{figures/skf.tex}
	\caption{Bearing's components}
    \label{figure:skf-bearing-components}    
\end{wrapfigure}
\vspace{-1em}

The dataset used in this section is bearings vibration dataset provided by Case Western Reserve University (CWRU). The bearings used in the test are SKF ball bearings. Figure \ref{figure:skf-bearing-components} shows the different components of a standard ball bearing.The test was conducted where the bearings support the shaft of a 2hp motor at different loading conditions. 

The test bearings have single point faults which were introduced using electro-discharge matching with fault diameters of 0.18mm, 0.36mm, 0.53mm, 0.71mm and 1.02mm. These faults were introduced in the bearing's ball, inner and outer raceways. SKF bearings were used for the 0.18mm, 0.36mm and 0.53mm faults, and NTN equivalent were used for the 0.71mm and 1.02mm faults. Table \ref{table:cwru-bearings-specification} contains the used SKF bearings' model dimensions and the corresponding frequencies (as multiples of RPM) associated with with different faults types.

\begin{table}[H]
	\centering
	\begin{tabu}{cc|[1.5pt]cc}
		\tabucline[1.5pt]{-} 
		Dimension		&	Size (mm)	&	Defect 			& Frequency ($\times$RPM Hz)	\\
		\hline
		Inner diameter	&	25.00		& Inner Ring 		& 5.4152\\
		Outside diameter&	52.00		& Outer Ring 		& 3.5848 \\
		Thickness 		&	15.00		& Cage Train		& 0.3983 \\
		Pitch diameter	&	08.03		& Rolling Element	& 4.7135\\
		\tabucline[1.5pt]{-} 
	\end{tabu}
	\caption{CWRU bearings dimensions and faults frequencies}
	\label{table:cwru-bearings-specification}
\end{table}

\section{Generating data from raw vibration signals}
\begin{figure}[h]
	\centering
	\includegraphics{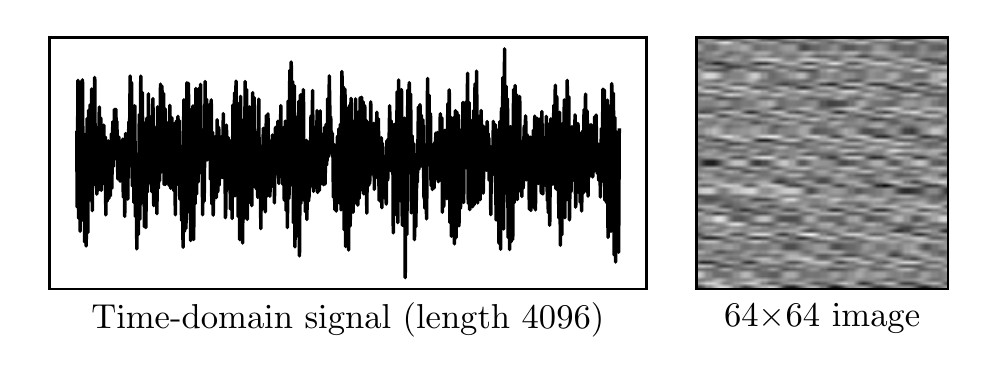}
	\caption{Data generation by converting the signal into 64$\times$64 images}
	\label{fig:cw_bearings_data_generation}
\end{figure}

Raw vibration data can't be used directly as an input to a neural network. This chapter uses the approach proposed in \cite{Wen2018} to convert raw vibration data into images. Figure \ref{fig:cw_bearings_data_generation} shows data generation principal where 1-dimensional vibration signals are converted into 2-dimensional arrays (images) by reshaping chunks of length 4096 into matrices of 64$\times$64.

As mentioned before, several tests have been conducted with different types of bearings faults (i.e. ball, inner and outer raceways faults) with different fault diameters. Signals from the different tests are transformed into images to serve as an input for a convolutional neural network which will be trained to classify the different signals into the corresponding fault types and their diameters. Signals are distributed across 10 classes: three different fault types and for each fault type there are three different fault diameters, plus a normal baseline signal that belongs to healthy bearing. Figure \ref{fig:bearings_faults_samples} shows some samples of the transformed signals of the different nine fault types/diameters: 

\begin{figure}[h]
    \centering
	\includegraphics{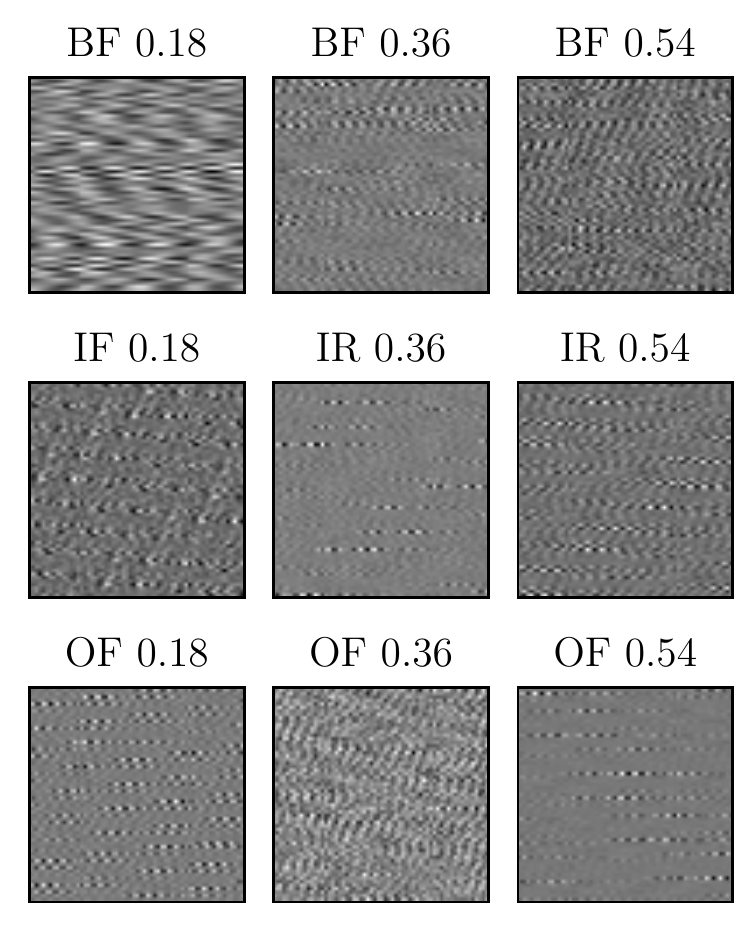}
    \caption{Converted signals of different faults types}
    \label{fig:bearings_faults_samples}
\end{figure}

Signals of the different fault types have varying lengths which results in a different number of synthesized images per fault type. Number of images per class is given by equation \ref{equation:labels-per-class}:

\begin{equation}
	N=floor \left(\frac{\text{signal length}}{64\times64}\right)
	\label{equation:labels-per-class}
\end{equation}

Number of images corresponding to each class is shown in in Table \ref{table:cw-classes-count}:

\begin{table}[h]
	\centering
	\begin{tabu}{lc}
		\tabucline[1.5pt]{-} 
	   Class 					&	Samples count	\\
	   \hline 
	   Normal bearing 			&	295				\\
	   Roller element 0.18mm 	&	146				\\
	   Roller element 0.36mm 	&	116				\\
	   Roller element 0.54mm	&	116				\\
	   Inner race 0.18mm		&	295				\\
	   Inner race 0.36mm		&	116				\\
	   Inner race 0.54mm		&	116				\\
	   Outer race 0.18mm		&	116				\\
	   Outer race 0.36mm		&	116				\\
	   Outer race 0.54mm		&	116				\\
   \tabucline[1.5pt]{-}
   \end{tabu}
   \caption{}
   \label{table:cw-classes-count}
\end{table}

\section{Case study: Bearings faults diagnostics using neural networks}%
\label{sec:case_study_bearings_faults_diagnostics_using_neural_networks}

After generating data by converting raw vibration signals into images, these images serve as an input for a convolutional neural network.

\subsection{Network architecture}
\acrlong{cnn} (\acrshort{cnn}) describes a type of neural networks that is suitable for image processing. This section presents the use of \acrshort{cnn} to classify bearings vibration signals that have been transformed into images into the corresponding fault types. \acrshort{cnn} were described in details in section \ref{section:cnn}.

To perform this classification task, a \acrshort{cnn} architecture is used, the network consists of three convolutional layers with max-pooling layers in between, followed by three fully-connected layers. All the details of this architecture from are mentioned in Table \ref{table:bearings-faults-cnn-classifier-architecture}.

\begin{table}[h]
    \centering
    \begin{tabu}{lll}
		\tabucline[1.5pt]{-}
		\textbf{Layer (type)}   & \textbf{Output shape} &   \textbf{Param \#} \\
		\tabucline[1pt]{-}
		Conv1 (Conv2D) 			&   (None, 1, 64 ,32)   &   18464   \\
		MaxPool1 (MaxPool2D) 	&   (None, 1, 32, 32)   &   0       \\
		Conv2 (Conv2D)			&   (None, 1, 32, 64)   &   18496   \\
		MaxPool2 (MaxPooling2D) &   (None, 1, 16, 64)   &   0       \\
		Conv3 (Conv2D)          &   (None, 1, 16, 128)  &   73856   \\
		MaxPool3 (MaxPooling2D) &   (None, 1, 8, 128)   &   0       \\       
		Flatten1 (Flatten)      &   (None, 1024)        &   0       \\     
		Dense1 (Dense)          &   (None, 128)         &   131200  \\   
		Dense2 (Dense)          &   (None, 64)          &   8256    \\     
		Dense3 (Dense)          &   (None, 10)          &   650     \\
		\tabucline[1pt]{-}
		Total params: 250,922       &                   &           \\
		Trainable params: 250,922   &                   &           \\
		Non-trainable params: 0     &                   &           \\
	\tabucline[1.5pt]{-}
    \end{tabu}
    \caption{\acrshort{cnn} architecture}
    \label{table:bearings-faults-cnn-classifier-architecture}
\end{table}

\subsection{Training process}
\begin{figure}[h]
    \centering
    \includegraphics{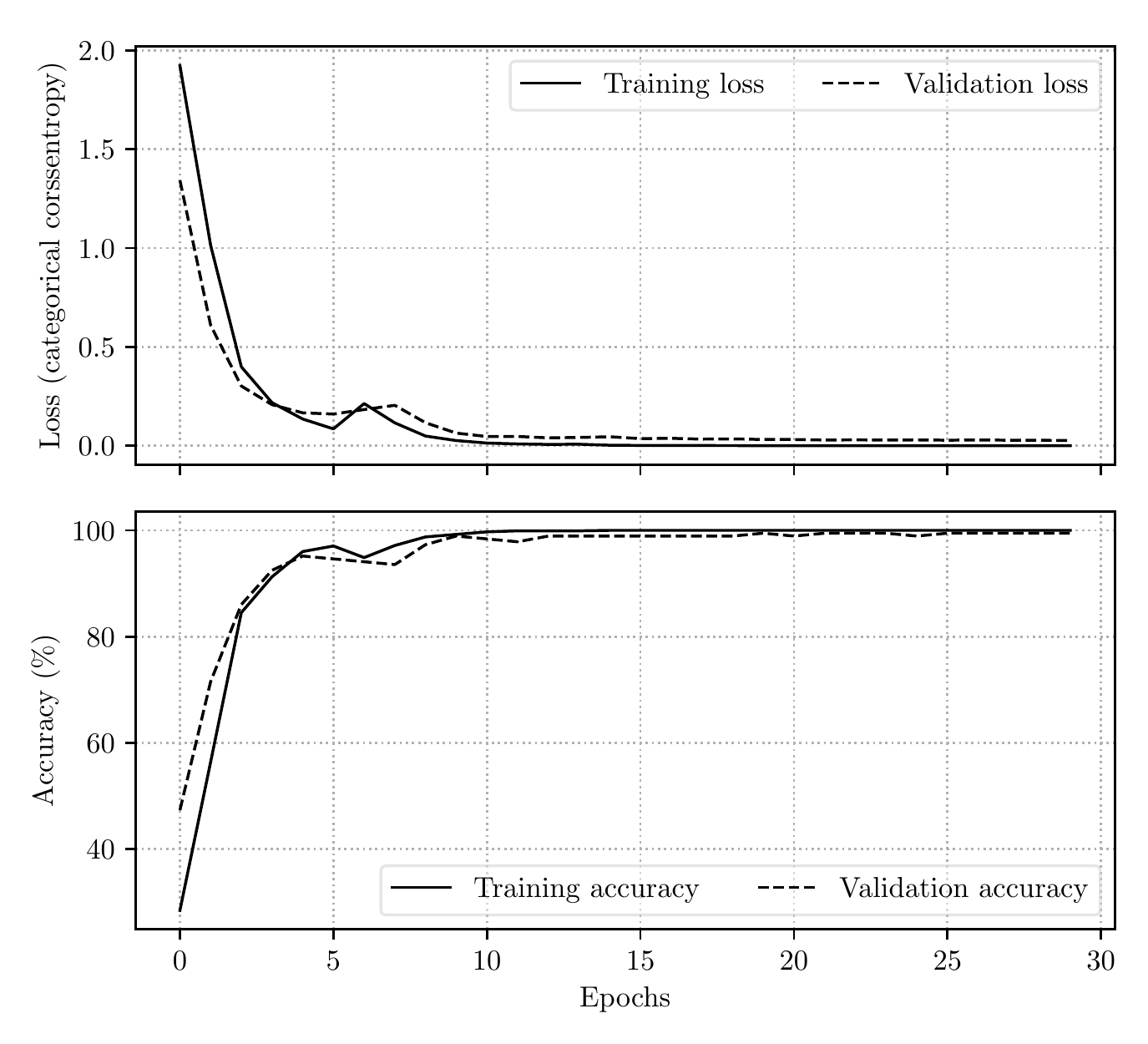}
    \caption{Classifier training}
    \label{fig:bearings_faults_classification_training}
\end{figure}

The dataset contains in total 1548 samples, 20\% is used as a test set while the remaining 80\% is the training set. From the train set, 15\% is used as a validation split during the training process. The network was trained for 30 epochs and a batch size of 32.
Figure \ref{fig:bearings_faults_classification_training} shows the evolution of the accuracy (\%) and loss (categorical cross-entropy) as a function of training epochs. The network converges around 10th epoch with minor increase in accuracy and decrease in loss afterwards. 

\subsection{Network results discussion}
The model achieves perfect 100\% accuracy on training set and a close accuracy on validation set. After training the model is evaluated on the test set and achieves a near perfect accuracy of 98.72\% accuracy. Train, validation and test loss and accuracy are summarized in Table \ref{table:cw-cnn-results}.

\begin{table}[H]
	\centering
	\begin{tabu}{lcc}
		\tabucline[1.5pt]{2-3} 
						&	\textbf{Loss}	&	\textbf{Accuracy}	\\
	   \tabucline[1pt]{-}
		Train set 		&	0.0003			&	100.00\%				\\
		Validation set 	&	0.0269 			&	99.46\%					\\
		Test set		&	0.0586 			&	98.71\%					\\
   \tabucline[1.5pt]{-}
   \end{tabu}
   \caption{\acrshort{cnn} training results}
   \label{table:cw-cnn-results}
\end{table}

\begin{figure}[h]
    \centering
    \includegraphics{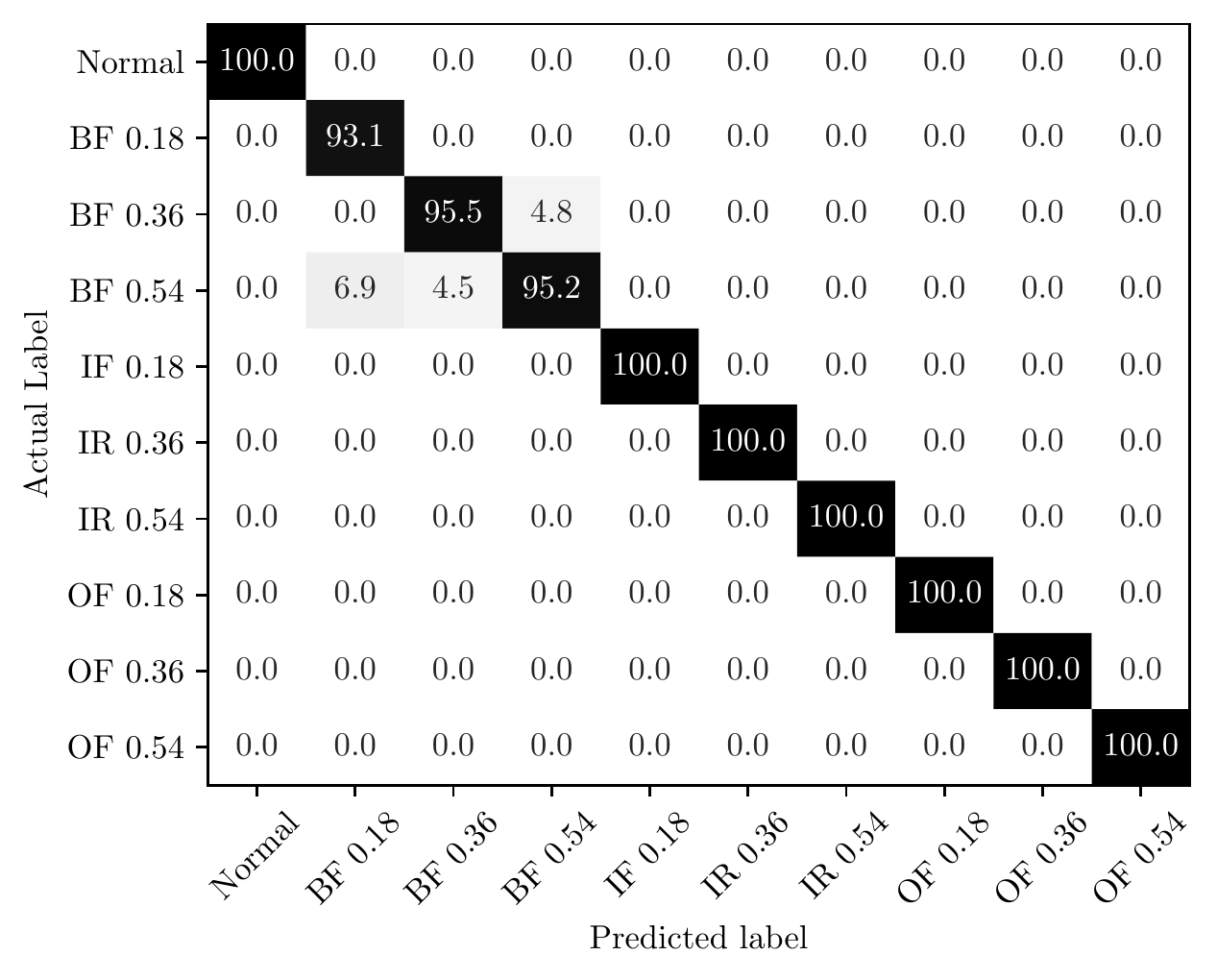}
    \caption{Confusion matrix of bearings faults classification using \acrshort{cnn}}
    \label{fig:bearings_faults_classification_confusion_matrix}
\end{figure}

To further understand the model's results, confusion matrix is constructed. The confusion matrix (Figure \ref{fig:bearings_faults_classification_confusion_matrix}) shows the results of model predictions on the test set where y-axis shows the actual (real) labels in the test set and x-axis the model predictions. Diagonal of confusion matrix is where actual labels intersect with model's predicted labels for each sample and it is apparent that the model achieves near-perfect classification.

\section{Case study: Bearings faults prognostics using neural networks}%
\label{sec:case_study_bearings_faults_prognostics_using_neural_networks}

Section \ref{sec:case_study_bearings_faults_diagnostics_using_neural_networks} presented an approach for diagnosing bearings faults using vibration data and neural networks. Diagnosis is important aspect of maintenance in general, but it's not enough alone, in a predictive maintenance program and in the context of \acrshort{phm}, it is also important to monitor the equipment, estimate its health state and project it into the future to predict the \acrshort{rul} and that is what is covered in this section. In this section, FEMTO bearings dataset is used.

\subsection{Introduction to FEMTO database}%
\label{sub:introduction_to_femto_database}

FEMTO Bearings dataset was provided by FEMTO-ST Institute (Besançon - France) as a part of PHM 2012 Data Challenge. The dataset consists of vibration and temperature measurements from bearings accelerated aging (run to failure) experiments which were conducted using PRONOSTIA platform.

\begin{figure}[h]
	\centering
	\includegraphics[width=.8\linewidth]{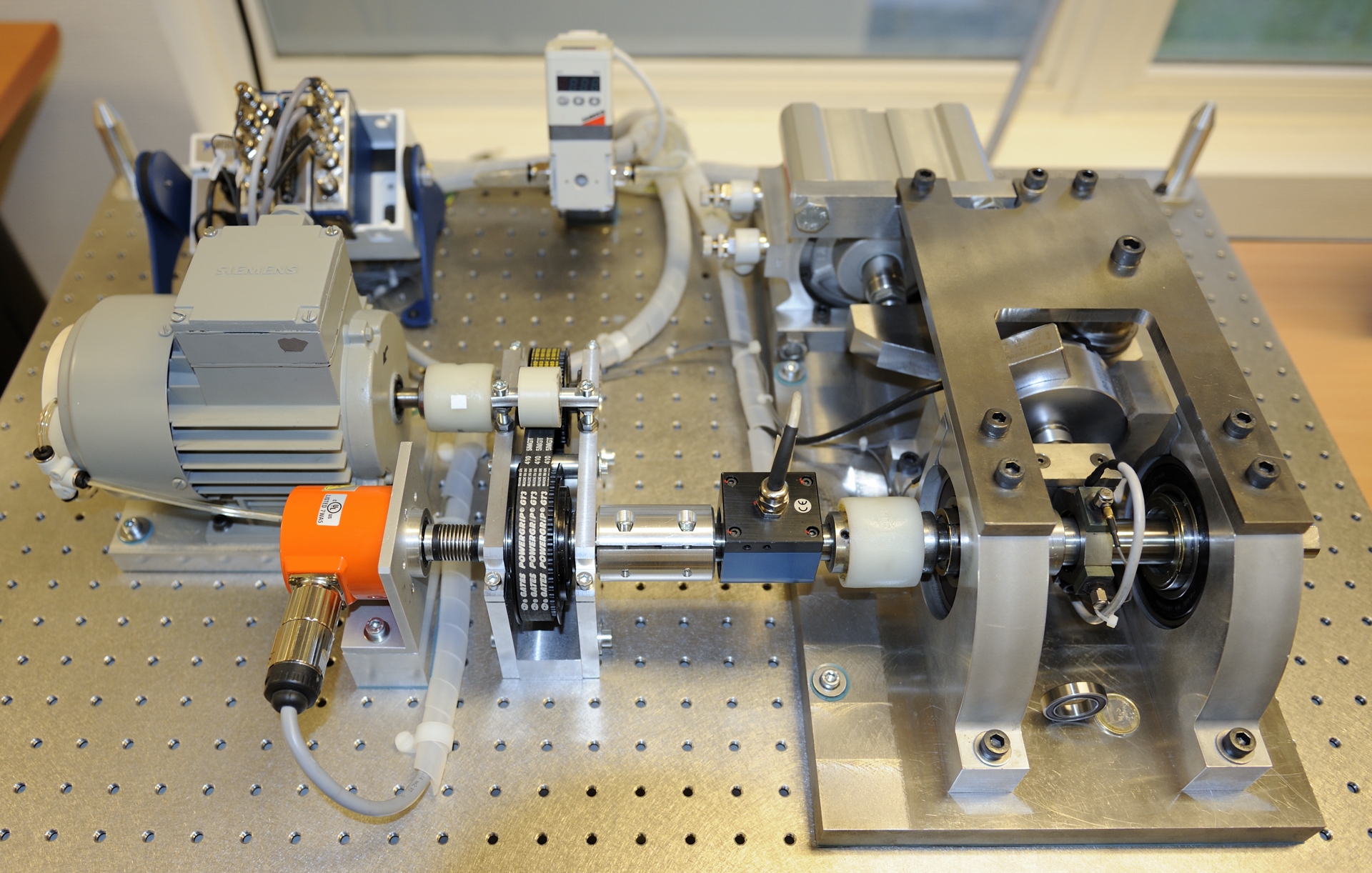}
	\caption{PRONOSTIA platform \cite{pronostia}}%
	\label{fig:pronostia-platform}
\end{figure}

PRONOSTIA (Figure \ref{fig:pronostia-platform}) is an experimental platform used for accelerated bearings life tests. PRONOSTIA makes it possible to conduct these tests in few hours. Data provided by the platform corresponds to naturally degraded bearings rather than using bearings with artificially-induced faults (e.g. CWRU bearings data).

PRONOSTIA consists of three main parts:

\begin{itemize}
	\item \textbf{Rotating part}: Asynchronous motor with a power of 250W that transmits rotation motion through a gearbox which allows a rotation speed up to 2830 rpm. The motor's speed and direction of rotation are controlled by the user through a human machine interface (HMI).
	\item \textbf{Load part}: Which is used to apply a radial force to induce faults in bearings. By applying a load that exceeds the bearing's maximum dynamic load (4000 N), this part allows to drastically reduce the bearing's life and conduct the accelerated aging test. The load is generated using a force actuator that consists of a pneumatic jack powered by a pressure delivered through a digital electro-pneumatic regulator.
	\item \textbf{Measurement part}: Operation conditions consist of three variables: a) the applied radial force, b) shaft speed and c) torque inflicted on the bearings, all three which are sampled at a frequency of 100Hz. Bearings health is measured by two variables: a) vibrations (both horizontal and vertical) which are measured by 2 accelerometers placed on the bearing's outer race and b) temperature measured by an RTD (Resistance Temperature Detector) probe placed close to the outer race. Vibration is sampled at 26.5kHz and temperature at 10Hz. Vibration measurements consist of snapshots of 2560 samples taken every 10s while temperature measurements are 600 samples taken each minute.
\end{itemize}

The bearings in the datasets are grouped under three operation conditions:
\begin{itemize}
	\item First operating conditions: rotation speed of 1800 rpm and load of 4000N.
	\item Second operating conditions: rotation speed of 1650 rpm and load of 4200N.
	\item Third operating conditions: rotation speed of 1500 rpm and load of 5000N.
\end{itemize}

Table \ref{table:femto-bearings-dataset} shows different operating conditions and their corresponding bearings in both learning and testing datasets:

\begin{table}[ht]
\centering
\begin{tabu}{cccc}
\tabucline[1.5pt]{-}
Datasets & \multicolumn{3}{c}{Operation conditions} \\
\cline{2-4}
 & Conditions 1 & Conditions 2 & Conditions 3 \\
\hline
Learning set & \begin{tabular}[c]{@{}c@{}}Bearing1\_1\\ Bearing1\_2\end{tabular} & \begin{tabular}[c]{@{}c@{}}Bearing2\_1\\ Bearing2\_2\end{tabular} & \begin{tabular}[c]{@{}c@{}}Bearing3\_1\\ Bearing3\_2\end{tabular} \\
\hline
Testing set & \begin{tabular}[c]{@{}c@{}}Bearing1\_3\\ Bearing1\_4\\ Bearing1\_5\\ Bearing1\_6\\ Bearing1\_7\end{tabular} & \begin{tabular}[c]{@{}c@{}}Bearing2\_3\\ Bearing2\_4\\ Bearing2\_5\\ Bearing2\_6\\ Bearing2\_7\end{tabular} & Bearing3\_3 \\
\tabucline[1.5pt]{-}
\end{tabu}
\caption{Different operating conditions and corresponding bearings in FEMTO dataset}
\label{table:femto-bearings-dataset}
\end{table}

\subsection{Generating data from scaleograms}%
\label{sub:generating_data_from_scaleograms}

It is important to detect bearing degradation before the actual failure happens. In this section, a convolutional neural network architecture will be used to classify healthy and faulty bearings using scaleograms from vibration data of FEMTO bearings dataset.

\begin{figure}[h]
	\centering
	\includegraphics[]{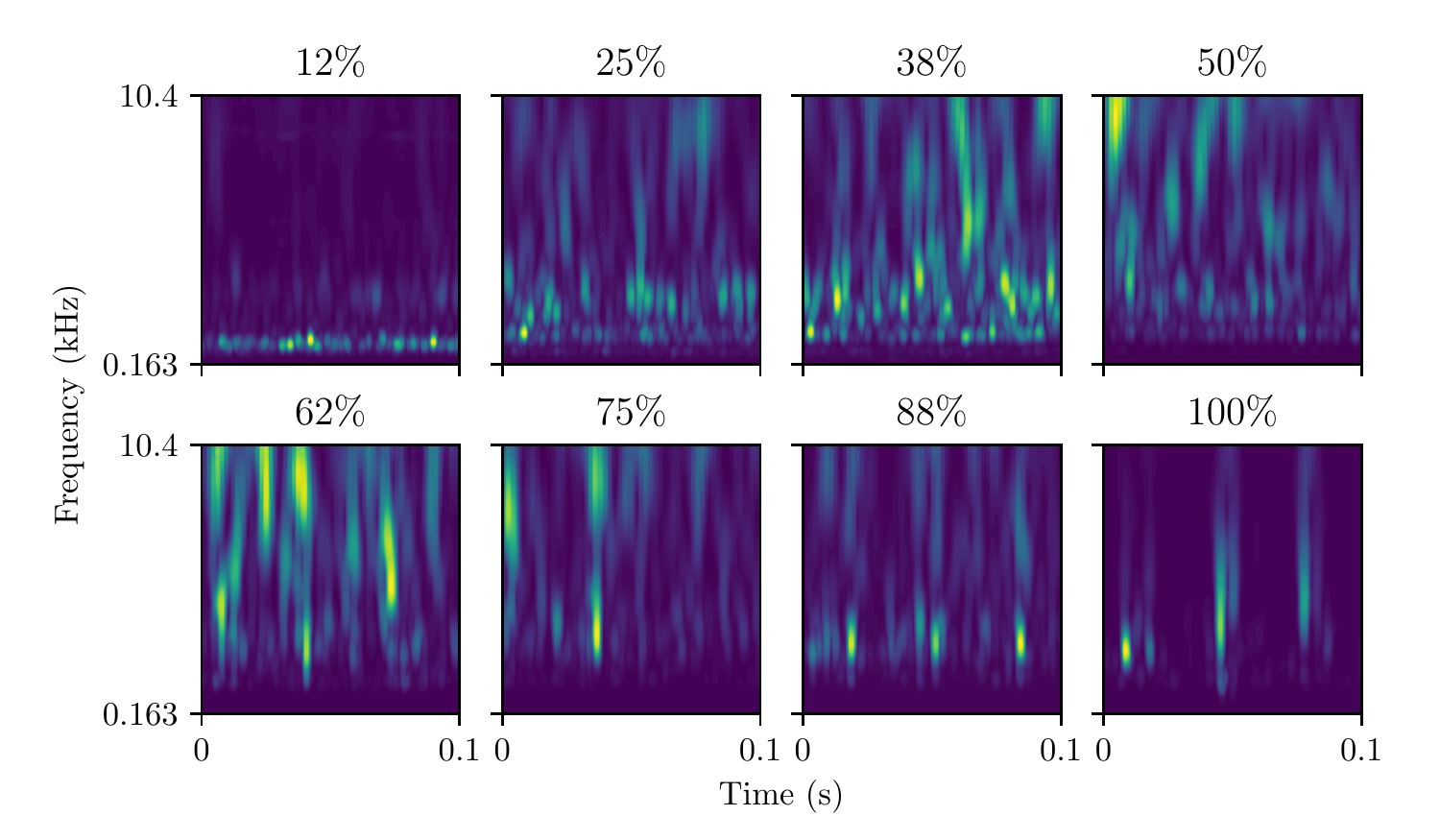}
	\caption{Scaleograms of different stages of Bearing1\_1 life (horizontal vibrations)}%
	\label{fig:bearings_fault_progress_scaleograms_h}
\end{figure}

\begin{figure}[h]
	\centering
	\includegraphics[]{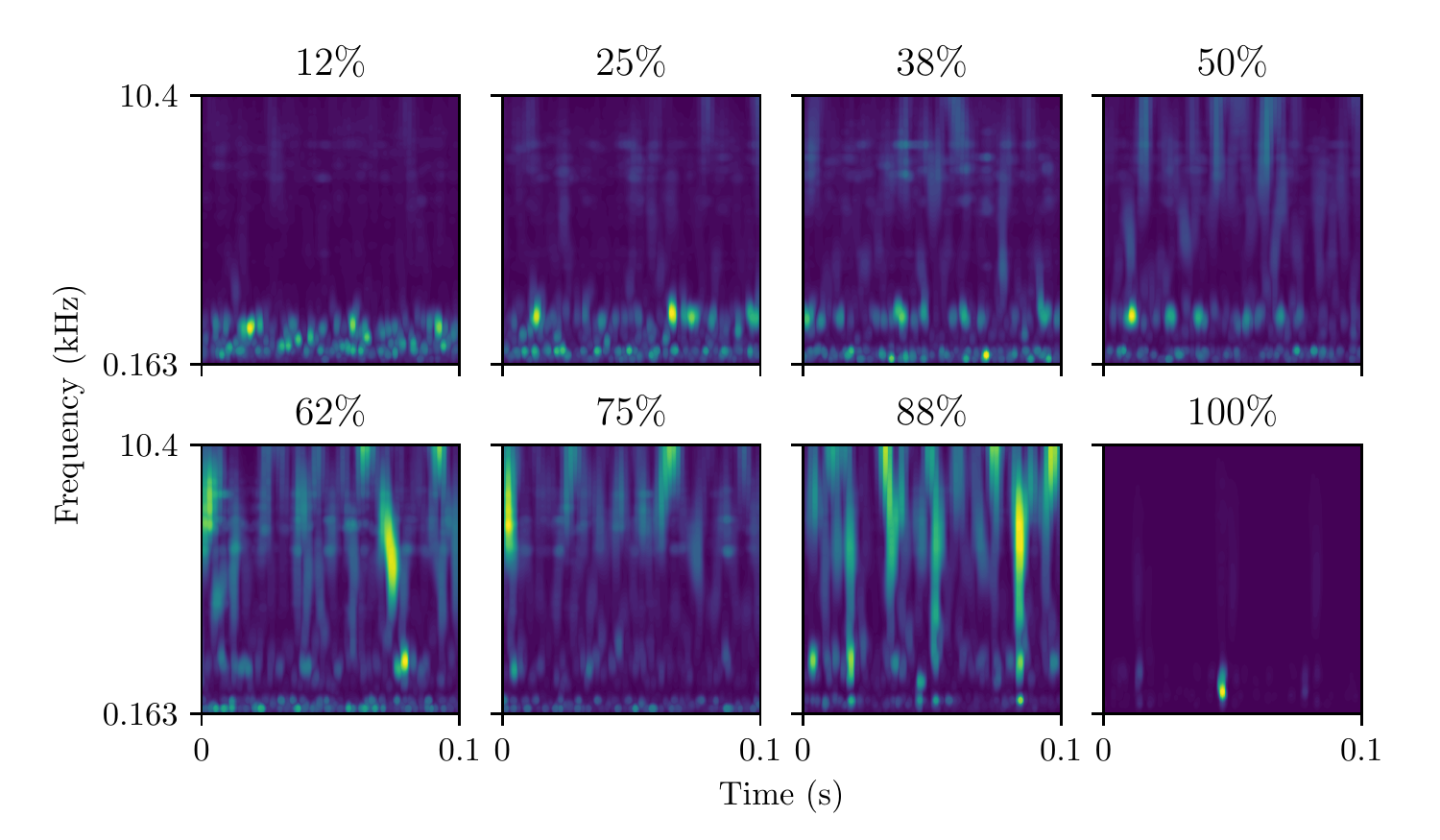}
	\caption{Scaleograms of different stages of Bearing1\_1 life (vertical vibrations)}%
	\label{fig:bearings_fault_progress_scaleograms_v}
\end{figure}

Figures \ref{fig:bearings_fault_progress_scaleograms_h} and \ref{fig:bearings_fault_progress_scaleograms_v} show different scaleograms of Bearing1\_1 horizontal and vertical vibration data (respectively) taken at different stages of the bearing's life.

Scaleograms can show important information about the bearing's health state, but different bearings can have different degradation patterns. For this reason, a dataset of scaleograms extracted from vibration data is constructed. The first 80 vibration snapshots from each bearing are considered to represent a healthy bearing, while the last 80 vibration snapshots are considered faulty. This is considered as a binary classification task, where a convolutional neural network will be used to automatically extract features from scaleograms (both horizontal and vertical vibration data scaleograms) dataset and classify the bearing's health state. To avoid complexity and the cost of training a large model, the scaleograms are resized to a shape of (128$\times$128), so the input to the network is of shape (2$\times$128$\times$128) where the two channels correspond to both horizontal and vertical vibration scaleograms. This data represents bearings from the first operation condition (from Bearing1\_1 to Bearing1\_7).

\subsection{Detecting bearings failure using convolutional neural networks}%
\label{sub:detecting_bearings_failure_using_convolutional_neural_networks}

\subsubsection{Network architecture}%
\label{subsub:network_architecture}
For this task, a convolutional neural network (\acrshort{cnn}) with 3 convolutional layers, 3 maxpooling layers and 3 fully-connected layers is used. The complete network architecture with the number of parameters in the network is represented in Table \ref{table:scaleograms-classifier-architecture}.

\begin{table}[ht]
    \centering
    \begin{tabu}{lll}
\tabucline[1.5pt]{-}
\textbf{Layer (type)}   & \textbf{Output shape} &   \textbf{Param \#} \\
\tabucline[1pt]{-}
Conv1 (Conv2D)		&	(None, 2, 128, 16)	&	18448\\
MaxPool1 (MaxPooling2D) &    (None, 1, 64, 16)		&	0\\
Conv2 (Conv2D)          &    (None, 1, 64, 32)         	&	4640\\
MaxPool2 (MaxPooling2D) &    (None, 1, 32, 64)         	&	0\\
Conv3 (Conv2D)          &    (None, 1, 32, 128)        	&	18496\\
MaxPool3 (MaxPooling2D) &    (None, 1, 16, 128)        	&	0\\
Flatten1 (Flatten)      &   (None, 2048)              	&	0\\
Dense1 (Dense)          &    (None, 1024)              	&	524800\\
Dropout1 (Dropout)	&	(None, 512)		&	0\\
Dense2 (Dense)          &    (None, 64)                	&	32832\\
Dropout2 (Dropout)	&	(None, 64)		&	0\\
Dense3 (Dense)          &    (None, 2)                 	&	66\\
\tabucline[1pt]{-}
Total params: 559,346 &				&	\\
Trainable params: 599,346				&	\\
Non-trainable params: 0&				&	\\
	\tabucline[1.5pt]{-}
    \end{tabu}
    \caption{Bearings health state classifier architecture}
    \label{table:scaleograms-classifier-architecture}
\end{table}

\subsubsection{Training process}%
\begin{figure}[H]
	\centering
	\includegraphics{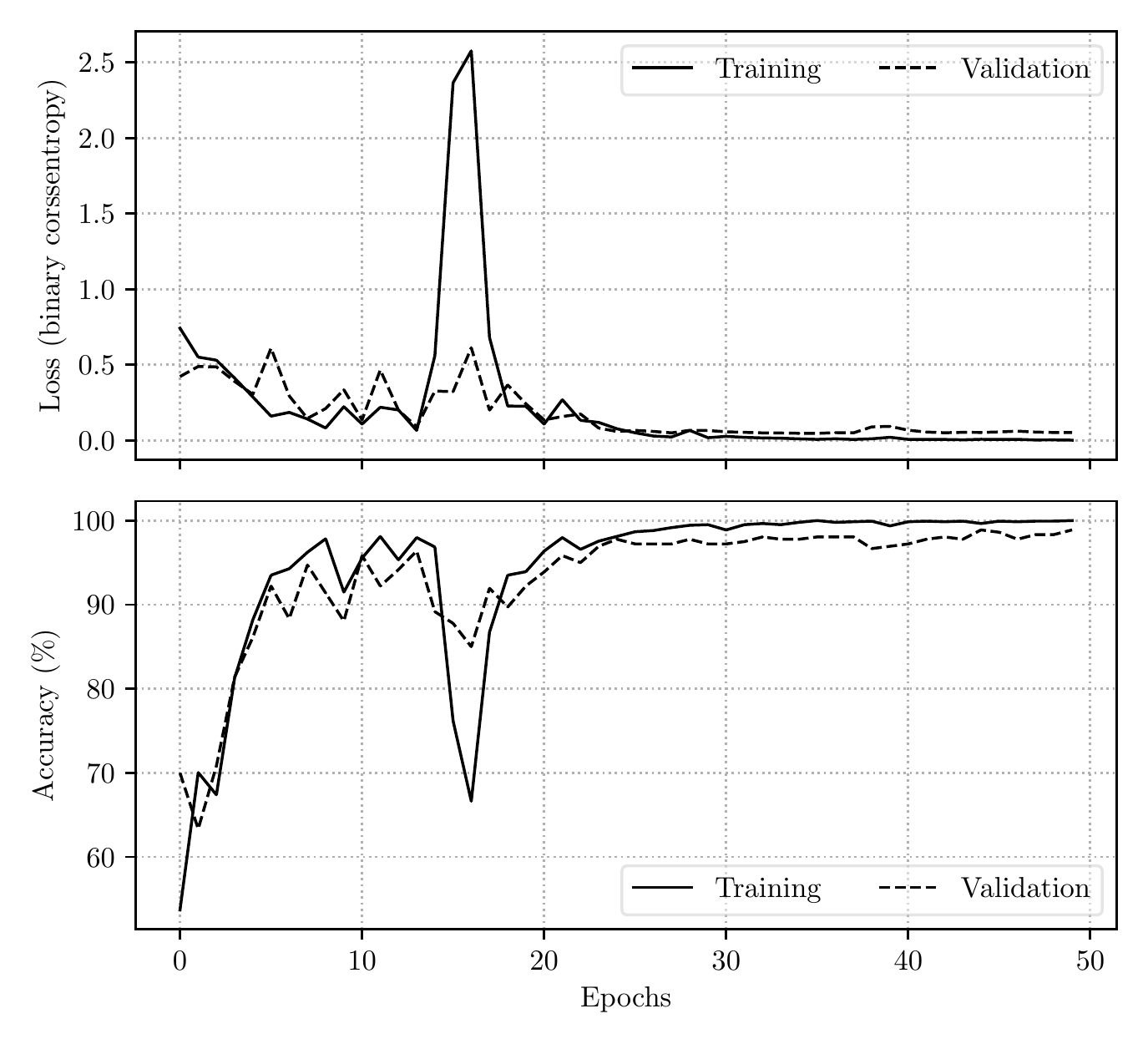}
	\caption{Bearings health state classifier training}%
	\label{fig:scaleogram-classifier-training}
\end{figure}

\label{subsub:training_process}
Generated dataset contained a total of 1120 samples, 716 samples were used for training, 180 for validation and 224 for test. The network was trained for 50 epochs with batch size of 64 samples. Figure \ref{fig:scaleogram-classifier-training} shows the training process.

\subsubsection{Results discussion}%
\label{subsub:results-discussion}
The network achieved a perfect training accuracy of 100\% and similar validation and test accuracy of around 98\%. Table \ref{table:femto-cwt-results} shows loss and accuracy on train, validation and test data. Table \ref{table:femto-cwt-metrics} shows the additional precision, recall and F-1 score metrics.

\begin{table}[H]
	\centering
	\begin{tabu}{lcc}
		\tabucline[1.5pt]{2-3} 
		&			\textbf{Loss}	&	\textbf{Precision}	\\
	   \tabucline[1pt]{-}
		Train dataset	&	0.0017	&	100.00\%		\\
		Validation dataset 	&	0.0528 	&	98.89\%			\\
		Test dataset	&	0.0527 	&	98.44\%			\\
   \tabucline[1.5pt]{-}
   \end{tabu}
   \caption{Training result}
   \label{table:femto-cwt-results}
\end{table}

\begin{table}[H]
	\centering
	\begin{tabu}{cc}
		\tabucline[1.5pt]{1-2} 
		\textbf{Metric} & \textbf{Value}	\\
	   \tabucline[1pt]{-}
		Precision	&	0.9896		\\
		Recall	 	&	0.9694		\\
		F1 Score	&	0.9794		\\
   \tabucline[1.5pt]{-}
   \end{tabu}
   \caption{Additional metrics for network performance}
   \label{table:femto-cwt-metrics}
\end{table}

Figure \ref{fig:bearings_health_state_classifier_roc} shows the ROC curve of the classifier:

\begin{figure}[H]
	\centering
	\includegraphics[]{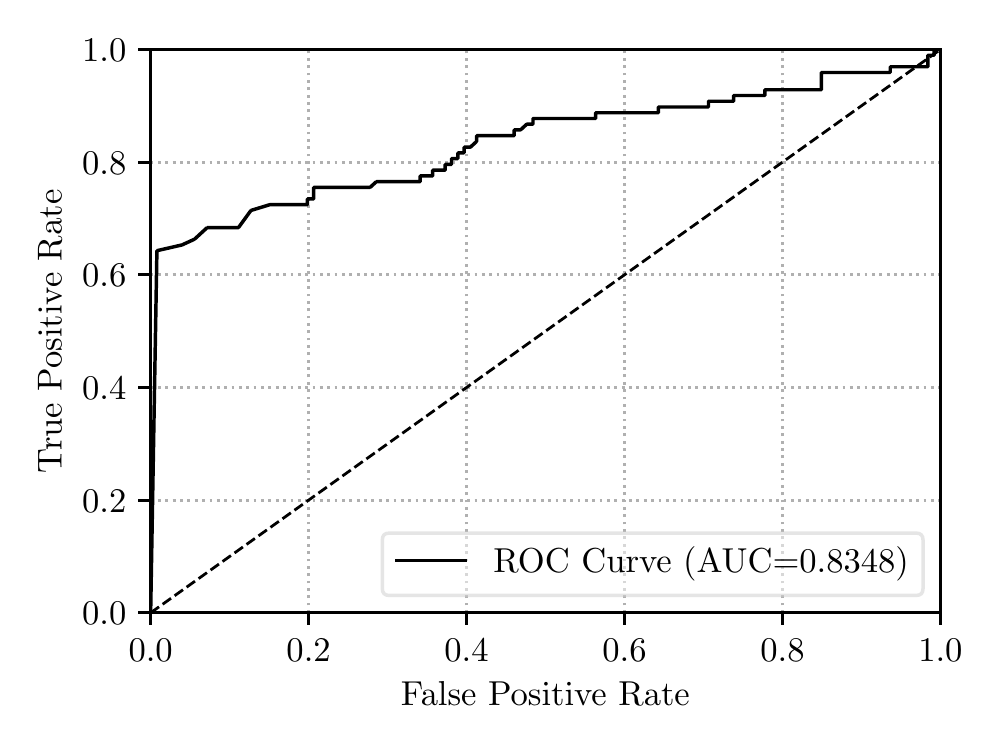}
	\caption{Bearings health state classifier ROC curve on test set}%
	\label{fig:bearings_health_state_classifier_roc}
\end{figure}

\subsection{The need for new prognostics features}%
\label{sub:the_need_for_appropriate_features}
The prognostic model's input is vital to its predictions accuracy \cite{coble2009} and since that's the case, a feature selection step should be carried out while developing the model, which aims to select a set of appropriate features that can make the prediction more accurate \cite{javed2012}. That's why it's important to use concrete metrics which can quantify the quality of our selected prognostics features and help comparing them directly and select the most suitable ones.

\subsubsection{Trendability and monotonicity}%
\label{subsub:trendability_and_monotonicity}
In \cite{coble2009} the authors proposed a set of metrics which can be used to quantify the quality of prognostics features and directly compare their suitability for use in predictive models, two of these metrics are \textbf{monotonicity} and \textbf{trendability}. According to their definition, monotonicity refers to the nature of the variable whether it is increasing or decreasing and it has a value between 0 and 1 where a variable that is always increasing or decreasing will have monotonicity of 1. Monotonicity is defined by equation \ref{equation:monotonicity}: 
\begin{equation}
	M=\frac{\text{no. of }\frac{d}{dx} > 0}{n-1} - \frac{\text{no. of }\frac{d}{dx} < 0}{n-1}
\label{equation:monotonicity}
\end{equation}

Trendability on the other hand was defined as whether a given feature follows the same trend (increasing or decreasing) across a population of systems. A feature that is always decreasing or increasing have a higher trendability than another one that doesn't follow a specific trend across different cases. Accordingly, trendability is defined by equation \ref{equation:trendability}:  

\begin{equation}
	\begin{aligned}
		t_i&= \frac{\text{no. of }\frac{d}{dx}>0}{n-1}+\frac{\text{no of } \frac{d^2}{dx^2}>0}{n-2}\\
Trendability&=1-std(t_i)
	\end{aligned}
	\label{equation:trendability}
\end{equation}

\subsection{Trigonometric features and cumulative descriptors}%
\label{sub:trigonometric_features}
In \cite{javed2013} the authors proposed a new feature extraction procedure from vibration data based on discrete wavelet transform (Section \ref{subsub:discrete_wavelet_transform}) and trigonometric functions. First, the raw vibration signal is decomposed using discrete wavelet transform with db4 wavelet and 4th level of decomposition, then the decomposition coefficients are scaled using trigonometric functions (e.g. asinh, atan), finally standard deviation of the scaled coefficients is calculated to obtain the final value. According to the authors, these features are less sensitive to noise and variability of raw vibration signals. 

Table \ref{table:trigonometric-classic_features} shows mathematical definition of different classic and trigonometric features used in literature:

\begin{table}[ht]
    \centering
    \begin{tabu}{ll}
		\tabucline[1.5pt]{-}
		\textbf{Trigonometric features}   & \textbf{Formula} \\
		\tabucline[1pt]{-}
		Standard deviation of asinh &   $\sigma\left(log\left[x_j+\sqrt(x_j^2+1)\right]\right)$  \\
		Standard deviation of atan  &   $\sigma\left(\frac{i}{2}log\left(\frac{i+x_j}{i-x_j}\right)\right)$ \\
					    &  \\
		\textbf{Classic features} & \textbf{Formula}\\
		\tabucline[1pt]{-}
		Entropy & $E(x)=\sum_jE(x_j)$ \\
		Energy & $e=\sum_{j=0}^nx_j^2$\\
		Root mean square & $RMS=\sqrt{\frac{1}{n}(x_1^2+\ldots+x_n^2)}$\\
		Skewness &  $\frac{\sum_{j=1}^n(x_j-\bar{X})^3}{(n-1)\sigma^3}$\\
		Kurtosis &  $\frac{\sum_{j=1}^n(x_j-\bar{X})^4}{(n-1)\sigma^4}$\\
		Upper bound & $max(x)+\frac{1}{2}\frac{max(X)-min(X)}{n-1}$\\
	\tabucline[1.5pt]{-}
    \end{tabu}
    \caption{Prognostics trigonometric and classic features \cite{javed2013}}
    \label{table:trigonometric-classic_features}
\end{table}

\begin{figure}[h]
	\centering
	\includegraphics[width=0.8\linewidth]{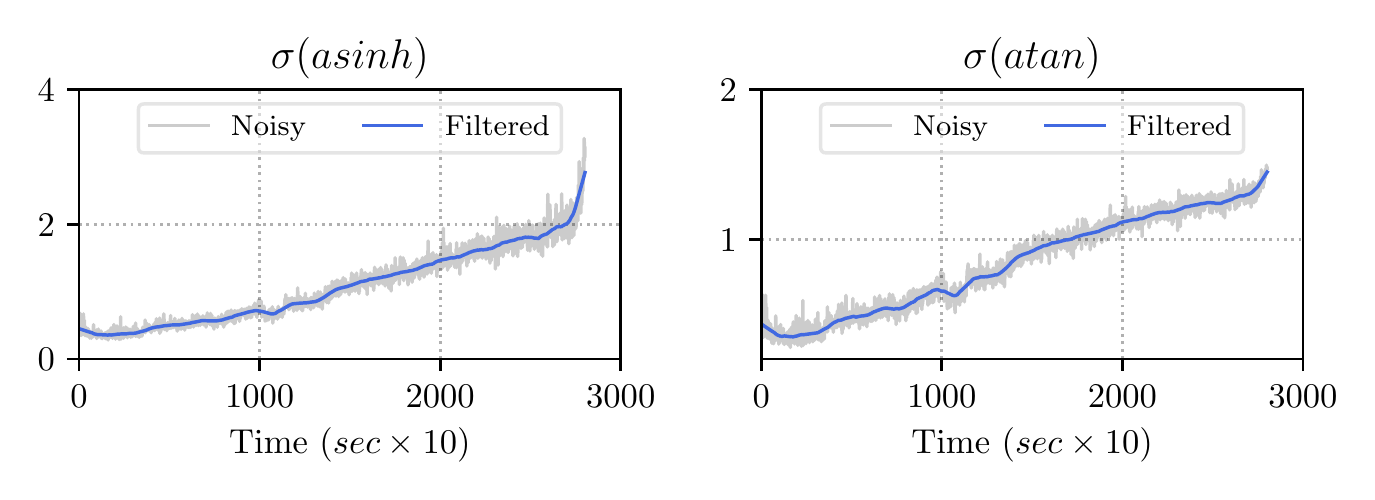}
	\caption{Trigonometric features of Bearing1\_1}%
	\label{fig:trigonometric_features_bearing1_1}
\end{figure}

Figure \ref{fig:trigonometric_features_bearing1_1} shows two trigonometric features (asinh and atan) for Bearing1\_1 from FEMTO dataset. They have been filtered using Savitsky--Golay filter to reduce noise and variability.

Even that trigonometric features are less susceptible to noise compared to classical features (e.g. rms, skewness, kurtosis, etc.), that's not always the case. The authors also proposed the use of cumulative descriptors (i.e. running total) in order to assure monotonicity of the features. The cumulative feature associated with each feature is calculated according to equation \ref{equation:cumulative_features} :

\begin{equation}
Cf_{nk} = \frac{\sum_{i=1}^n f_{ik}} {\sqrt{abs\left(\sum_{i=1}^nf_{ik}\right)}}
\label{equation:cumulative_features}
\end{equation}

\begin{figure}[H]
	\centering
	\includegraphics[width=0.8\linewidth]{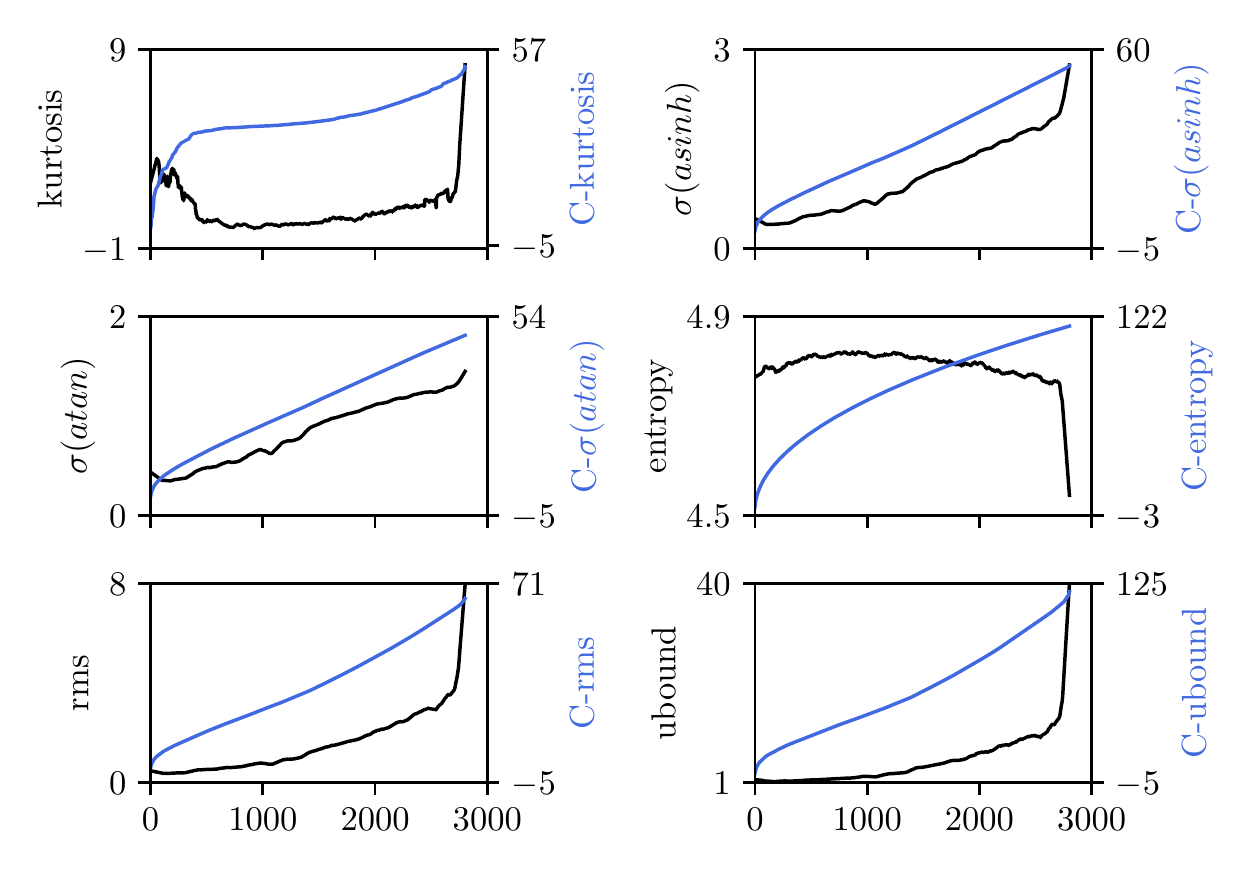}
	\caption{Classic and trigonometric cumulative features}%
	\label{fig:trig_classic_cumulative_features}
\end{figure}

In Figure \ref{fig:trig_classic_cumulative_features} cumulative classic and trigonometric features are compared to their non-cumulative counterparts. Even with only visual inspection it is apparent that using cumulative descriptors greatly enhances the quality of prognostics features by reducing noise and fluctuations. 

To quantify the enhancement caused by using cumulative descriptors on prognostics features, monotonicity and trendability for each feature and its cumulative counterpart is calculated and summarized in Tables \ref{table:trigonometric-classic-monotonicity} and \ref{table:trigonometric-classic-trendability} respectively.

\begin{table}[ht]
\centering
\begin{tabu}{cc|cc}
\tabucline[1.5pt]{-}
Feature & Monotonicity & Cumulative Feature & Monotonicity \\
\hline
$\sigma(atan)$ & 0.486 & C-$\sigma(atan)$ & 1 \\
$\sigma(asinh)$ & 0.481 & C-$\sigma(asinh)$ & 1 \\
kurtosis & 0.059 & C-kurtosis & 0.998 \\
entropy & 0.035 & C-entropy & 1 \\
rms & 0.481 & C-rms & 1 \\
ubound & 0.287 & C-ubound & 1\\
\tabucline[1.5pt]{-}
\end{tabu}
\caption{Monotonicity difference between trigonometric and classic features and their cumulative descriptors}
\label{table:trigonometric-classic-monotonicity}
\end{table}

\begin{table}[ht]
\centering
\begin{tabu}{cc|cc}
\tabucline[1.5pt]{-}
Feature & Trendability & Cumulative Feature & Trendability \\
\hline
$\sigma(atan)$ & 0.987 & C-$\sigma(atan)$ & 0.993 \\
$\sigma(asinh)$ & 0.989 & C-$\sigma(asinh)$ & 0.995 \\
kurtosis & 0.985 & C-kurtosis & 0.890 \\
entropy & 0.994 & C-entropy & 0.976 \\
rms & 0.988 & C-rms & 0.993 \\
ubound & 0.990 & C-ubound & 0.996\\
\tabucline[1.5pt]{-}
\end{tabu}
\caption{Trendability difference between trigonometric and classic features and their cumulative descriptors}
\label{table:trigonometric-classic-trendability}
\end{table}

It is very obvious that features monotonicity increased significantly for cumulative features compared to non-cumulative ones. But for trendability the difference was less significant. Figure \ref{fig:features_fitness} plots both monotonicity (x-axis) and trendability (y-axis) of both types of features:

\begin{figure}[H]
	\centering
	\includegraphics{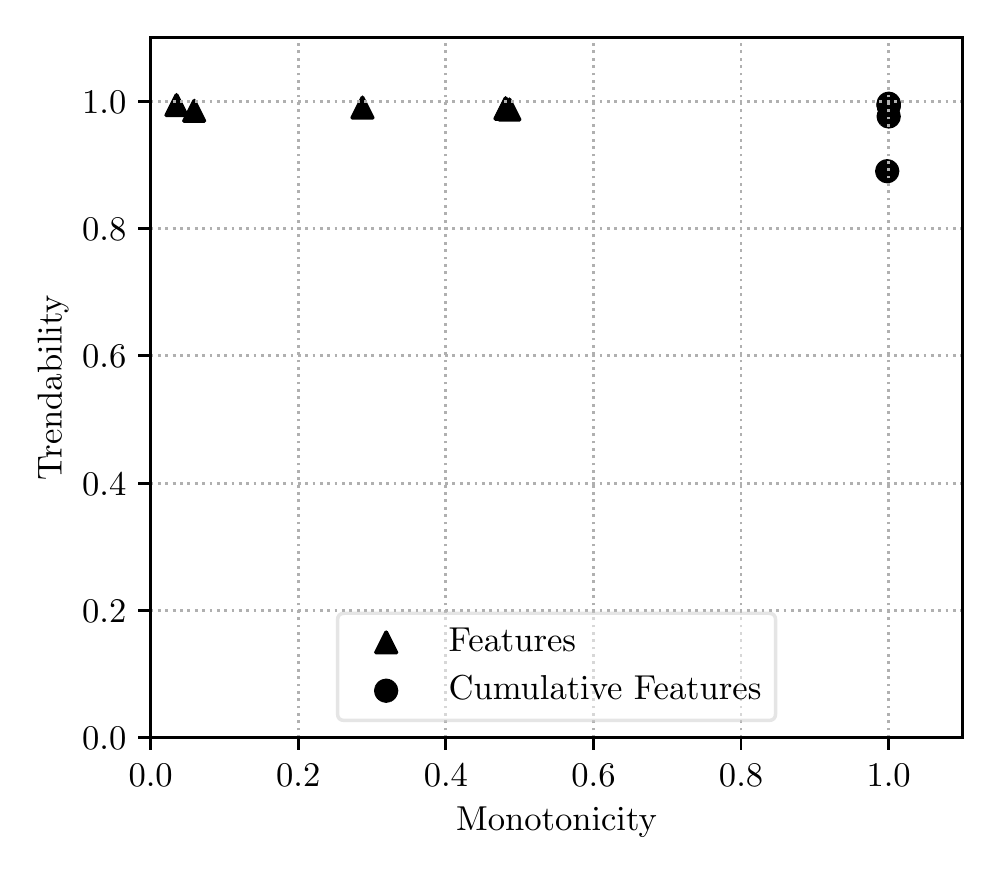}
	\caption{Features fitness}%
	\label{fig:features_fitness}
\end{figure}

\section{Application to oilfield equipment}%
\label{sec:application_to_oilfield_equipment}

\subsection{Top Drive}%
\label{sub:top_drive}

In oil drilling rigs, Top Drive is a device that is used to turn the drill string and replace the conventional rotary table and kelly. The main advantage of Top Drives over conventional solutions is the ability to drill using three joints of pipes instead of just one, thus reducing drastically drilling time. Top Drives also helps drillers to minimize the cost and frequency of stuck pipe incidents \cite{slbtopdrive}.

\begin{figure}[h]
	\centering
	\includegraphics[width=0.9\linewidth]{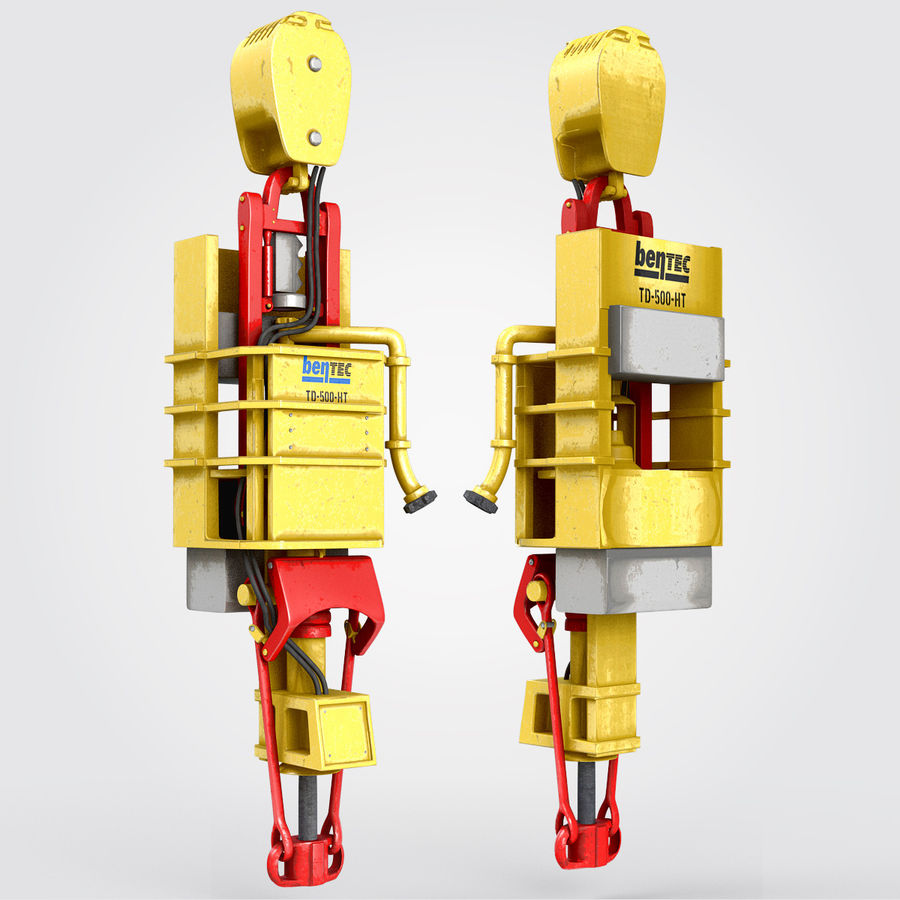}
	\caption{Topdrive Bentec 500-HT}%
	\label{fig:figures/bentec_500_ht}
\end{figure}

Figure \ref{fig:figures/bentec_500_ht} shows Bentec 500-HT Top Drive.

From an engineering perspective, Top Drives are much more complicated systems than conventional rotary table and kelly, thus they need more rigorous maintenance programs to ensure their availability considering their major role in drilling operations. Every year the oil and gas technology is pushed harder, to drill in harder conditions and to drill deeper and more challenging oil wells, this poses stricter requirements on used equipment: to be able under and sustain harsher conditions while maximizing its availability which is of great importance in the field, for example  Top Drives downtime cost can reach 1m\$ per day and cause significant delay to drilling operations \cite{skfbrochure}. 

In \cite{Pournazari2016} the authors reported an industry survey that was conducted to assess the impression of different groups of people (operators, contractors, rig builders…) on the usage of Top Drives. The survey showed an average 60\% satisfaction level across all the groups in the survey, which is an indicative that Top Drives aren't really meeting the expectations in the industry. They were also asked which are the features they would like to see in Top Drives, the most wanted things are less downtime and better ability to detect failures before the breakdown of the system. This means that a more reliable preventive maintenance and prognostics approach is the most wanted improvement to these systems in the field.

\subsection{Top Drive components}%
\label{sub:top_drive_components}

Top Drives are made of many sub-assemblies which are shown in Figure \ref{fig:topdrive-subassemblies}\footnote{All the following figures of Top Drive components were taken from Bentec technical bulletin for TD-500-HT and TD-350-HT Top Drives}.

\begin{figure}[H]
	\centering
	\includegraphics[width=\linewidth]{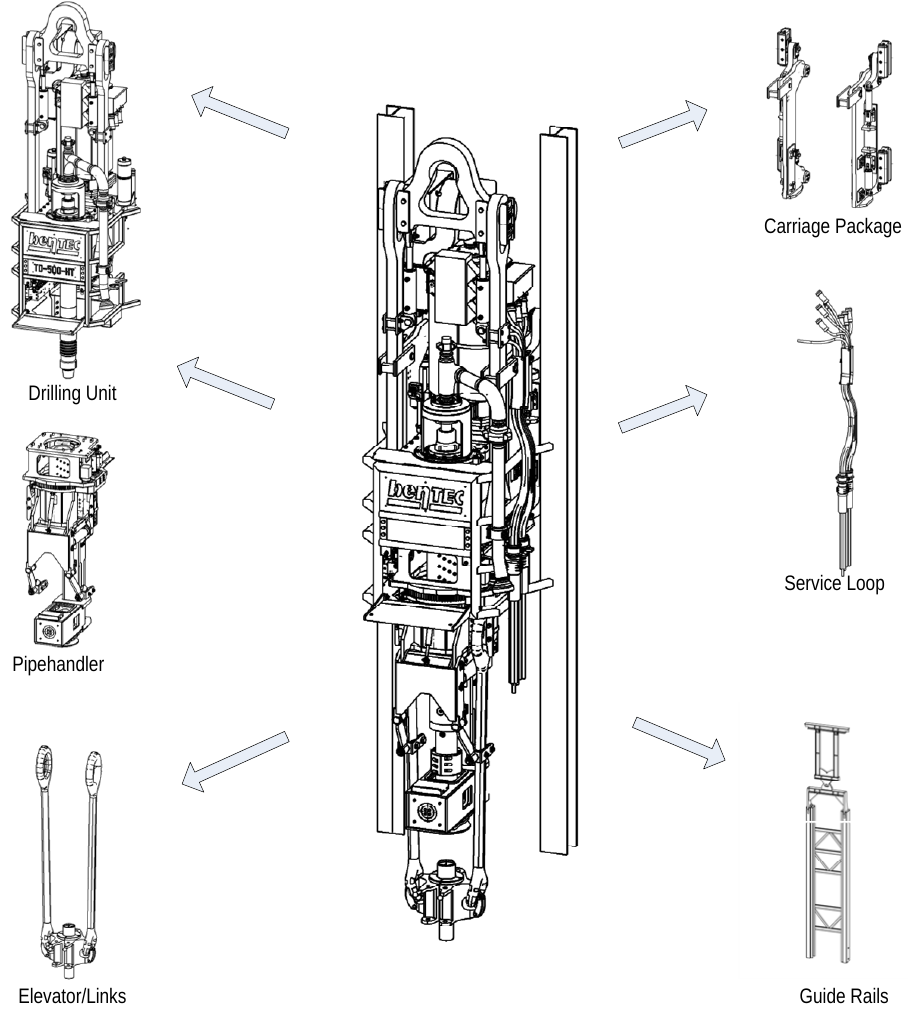}
	\caption{Bentec 500-HT Top Drive sub-assemblies}%
	\label{fig:topdrive-subassemblies}
\end{figure}

\begin{figure}[H]
	\centering
	\includegraphics[width=.7\linewidth]{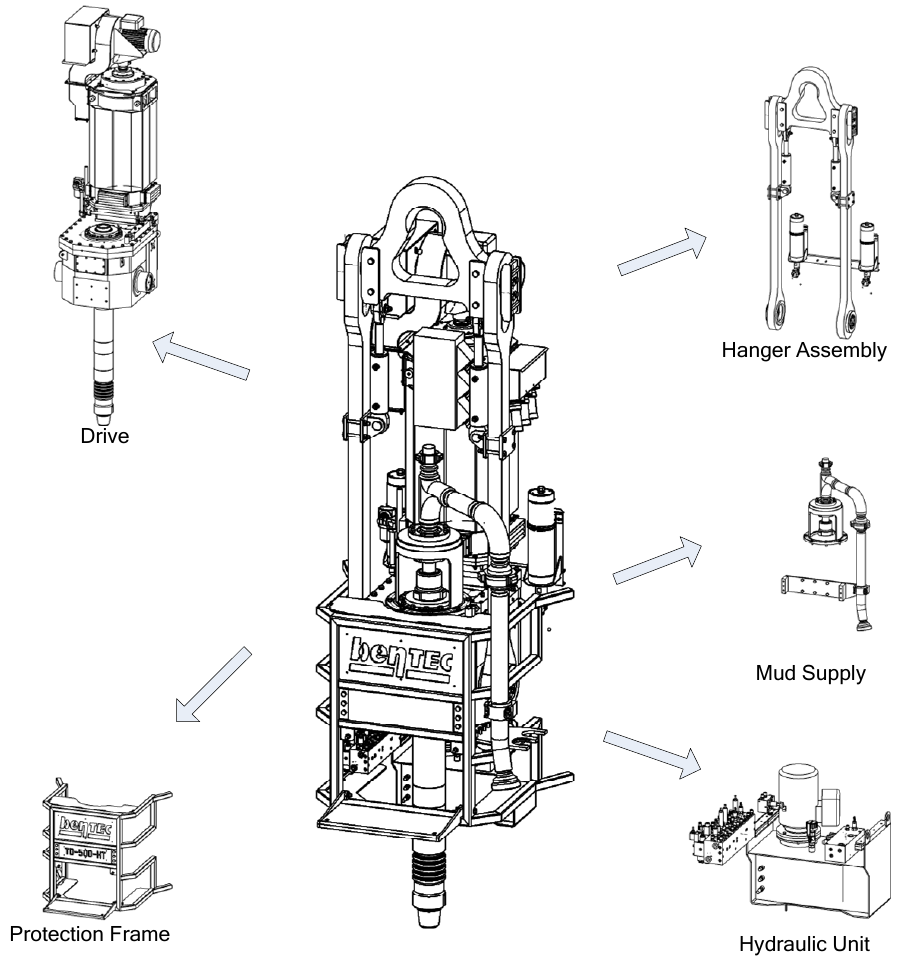}
	\caption{Bentec 500-HT Top Drive drilling unit}%
	\label{fig:topdrive-drillingunit}
\end{figure}

The Top Drive sub-assembly of interest here is the drilling unit (Figure \ref{fig:topdrive-drillingunit}) which is responsible for generating the rotation motion and transferring it to the drill string. Drilling unit is composed of protection frame, hydraulic unit, mud supply, hanger assembly and\textemdash the most important for the current discussion\textemdash a drive.

\begin{figure}[H]
	\centering
	\begin{subfigure}[t]{0.4\textwidth}
         \centering
	 \includegraphics[width=.7\textwidth]{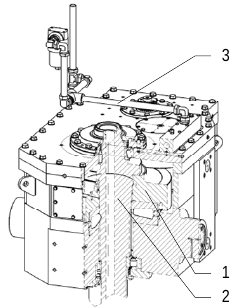}
	 \label{fig:topdrive-drillingunit-drive-gearbox}
	 \caption{Drive gearbox}
     \end{subfigure}%
     \begin{subfigure}[t]{0.4\textwidth}
         \centering
 \includegraphics[width=0.7\textwidth]{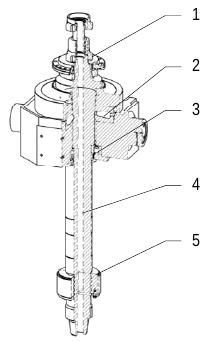}
	 \label{fig:topdrive-drillingunit-drive-mainshaft}
	 \caption{Drive main shaft}
     \end{subfigure}
	\caption{Main components of the drive}%
	\label{fig:topdrive-drillingunit-drive-components}
\end{figure}

The drive itself is composed of:

\begin{itemize}
	\item Engine cooling system
	\item Brake
	\item AC Motor
	\item Gearbox (Figure \ref{fig:topdrive-drillingunit-drive-components}(a))
	\item Main shaft (Figure \ref{fig:topdrive-drillingunit-drive-components}b)
\end{itemize}

The gearbox has two stages with transmission ratio of 14:1. The speed of the engine is reduced twice then transferred to the bull-gear (1) of the gearbox. The bull-gear is located on the thrust bearing. Lubrication of the gearbox is done with combined splash/pressure lubrication (3). The main shaft (4) is powered by the bull-gear of the gearbox and located in the gearbox with the thrust bearing (2) and it is conducted via lower bearing (3). A wash pipe (1) is connected to execute drilling fluid. The main shaft also holds the load collar (5) which bears the link adapter to the drill pipe. 

\begin{figure}[H]
	\centering
	\includegraphics[width=0.6\linewidth]{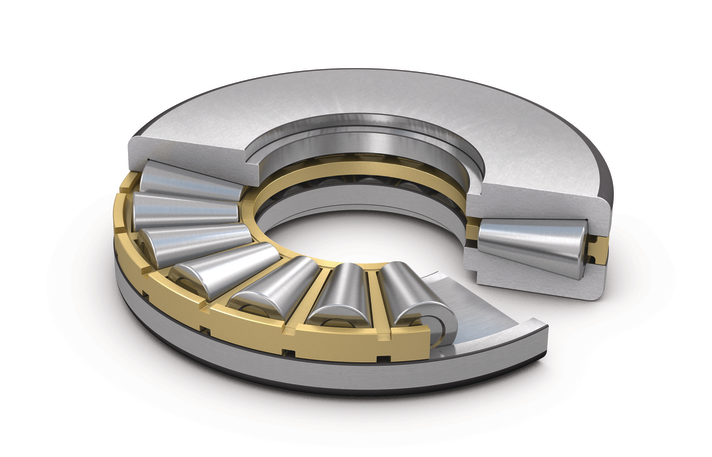}
	\caption{SKF tapered roller thrust bearing \cite{skf_tapered_roller_thurst_bearing}}%
	\label{fig:skf_tapered_roller_thrust_bearing}
\end{figure}

\begin{figure}[H]
	\centering
	\includegraphics[width=0.7\linewidth]{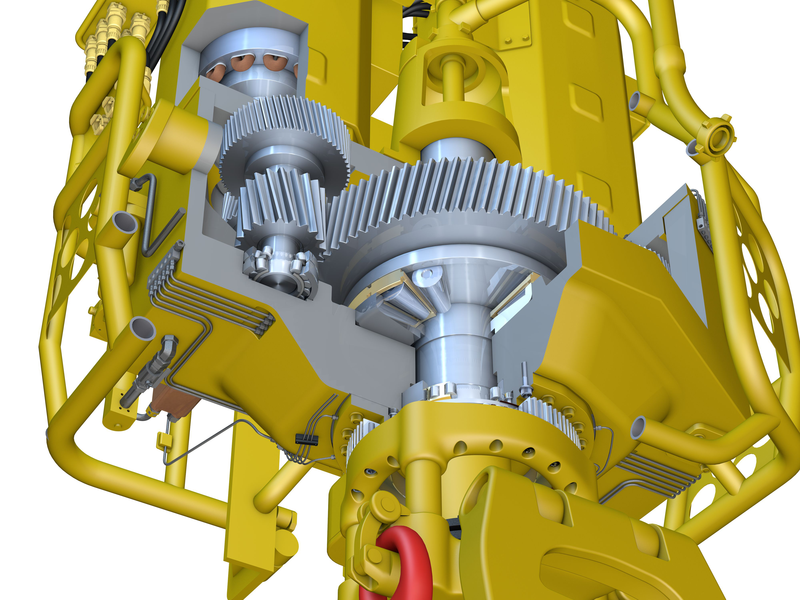}
	\caption{Bearing position within the Top Drive \cite{skf_bearing}}%
	\label{fig:skf_topdrive}
\end{figure}

Figure \ref{fig:skf_tapered_roller_thrust_bearing} shows an example of a tapered bearing usually used in Top Drives. While Figure \ref{fig:skf_topdrive} shows the bearing within a Top Drive.

\subsection{Proposed approach for Top Drive monitoring using neural networks}%
\label{sub:proposed_approach_for_top_drive_monitoring_using_neural_networks}

Top Drives are already equipped with many sensors to monitor their state like temperature, pressure and flow sensors, yet information that these sensors is only used in a simplified manner by operators \cite{Pournazari2016}. In order to improve maintenance programs and adopt preventive maintenance and prognostics approaches, a more sophisticated and methodological approach to process data provided by these sensors need to be employed, along with introduction of new vibration sensors which are the main monitoring sensors used for bearings and gears, the elements which are present in the top drive and are work under high and varying loads.

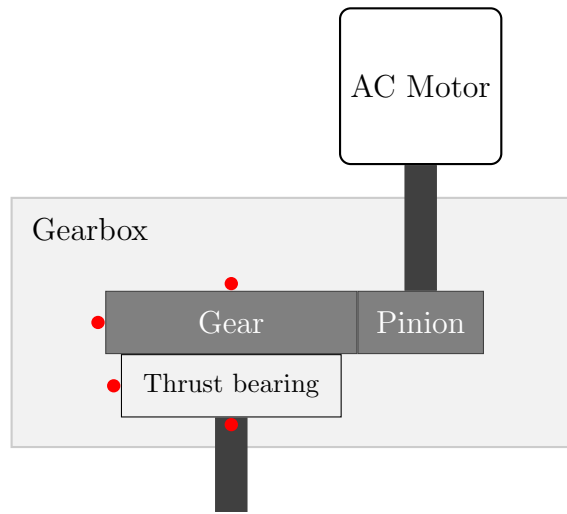
\begin{figure}[H]
	\centering
	\input{figures/topdrive_drive_sensors.tex}
	\caption{Proposed additional vibration sensors to monitor the drive}%
	\label{fig:topdrive-drive-sensors}
\end{figure}

Figure \ref{fig:topdrive-drive-sensors} is a simplified schematic representation of the proposed procedures to employ additional sensors (red dots) for vibrations in the drive to monitor different elements like gears and bearings. Enough sensors should be installed to capture vibrations in the three axes: vertical, horizontal and axial vibrations. When the Top Drive is in use, data collected from these sensors can be used to develop a neural network architecture to detect bearings degradation (Section \ref{sec:case_study_bearings_faults_prognostics_using_neural_networks}) or---if enough historic data is available---classifying different degradation patterns (Section \ref{sec:case_study_bearings_faults_diagnostics_using_neural_networks}). Correctly detecting degradation or even classifying degradation patterns can be used to plan and prepare the appropriate maintenance actions and execute them in the appropriate time before any serious degradation happens which may cause serious non-productive time and high corrective maintenance costs. 

\section{Conclusion}
This chapter adopted a data generation approach from literature that converts vibration signals into images. Vibration signals used here corresponds to bearings with different types of faults and fault diameters. After converting raw signals into images, a \acrshort{cnn} is used to classify transformed signals (i.e. images) into their corresponding faults types and diameters. The model achieves near-perfect classification accuracy on the test set.

%% file: figures/skf.tex
\begin{tikzpicture}
	\node (outer) at (3.5,1.2) {\makecell{\small Outer\\Race}};
	\node (inner) at (3.5,0) {\makecell{\small Inner\\Race}};
	\node (ball) at (3.5,-1.2) {\makecell{\small Ball\\(in cage)}};
	
	\node (outer2) at (1.3,1.2) {};
	\node (inner2) at (.3,0) {};
	\node (ball2) at (-.5,-1.2) {};

	\node[inner sep=0] (image) at (0,0) {\includegraphics[width=0.35\textwidth]{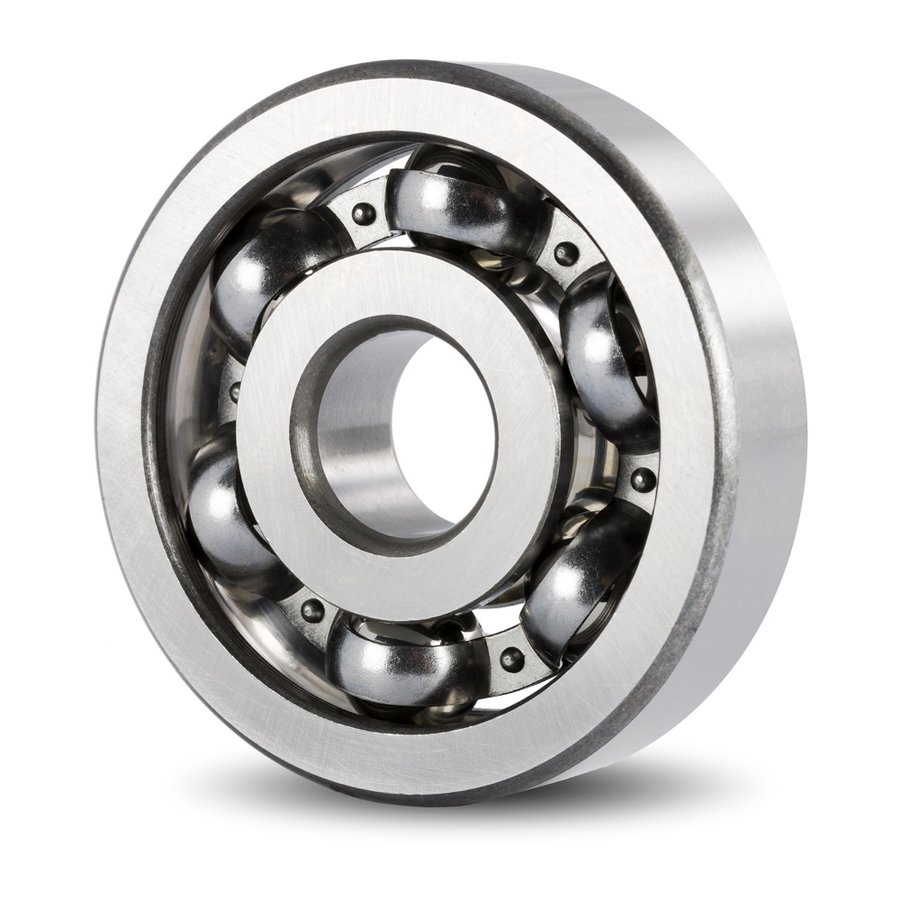}};

	\draw [|->,  thick, red] (outer.west) -- (outer2);
	\draw [|->,  thick, red] (ball) -- (ball2);
	\draw [|->,  thick, red] (inner) -- (inner2);
\end{tikzpicture}

%% file: figures/topdrive_drive_sensors.tex
\begin{tikzpicture}
	\tikzstyle{gear} = [rectangle,  draw, align=center, darkgray, fill=gray, text=white]
	\tikzstyle{bearing} = [rectangle, draw, align=center]
	\tikzstyle{shaft} = [rectangle, draw, minimum width = 1em, fill=darkgray, darkgray]
	\tikzstyle{sensor}  = [circle, fill,red, minimum size=5,inner sep=0]
	\draw[black!20!white, thick, fill=black!5!white ] (0,0) rectangle (18em,-8em);
	\node[] at (2.5em,-1em) {Gearbox};

	\node[gear, minimum width=8em, minimum height=2em] at (7em,-4em) (gear) {Gear};
	\node[gear, minimum width=4em, minimum height=2em, right = 0em  of gear] (pinion) {Pinion};

	\node[bearing, minimum width=7em, minimum height=2em, below = 0em of gear] (thrustbearing) {\footnotesize Thrust bearing};

	\node[shaft, minimum height=4em, above = 0em of pinion] (shaft) {};
	\node[draw, thick, minimum width=3em, rounded corners, radius=3,minimum height=5em, above = 0em of shaft] {AC Motor};

	\node[shaft, minimum height=3em, below = 0em of thrustbearing] {};

	\node[sensor, below = 0em of thrustbearing] {};
	\node[sensor, left = 0em of thrustbearing] {};
	\node[sensor, above = 0em of gear] {};
	\node[sensor, left = 0em of gear] {};
	
\end{tikzpicture}	

%% file: chapters/conclusion.tex
\chapter*{Conclusion}
\addcontentsline{toc}{chapter}{Conclusion}
Maintenance is an important process in all industrial facilities that ensures equipment availability and reduces equipment downtime. We can distinguish different types of maintenance actions where each type has a specific area of application. For critical equipment where repair cost is elevated and downtime can result in large losses, predictive maintenance was the preferred maintenance policy for many decades, yet with the technological advance in the last couple of decades where computational power become cheaper and sensors are ubiquitous, the industry started to shift towards what's called predictive maintenance and prognostics, which aim at providing predictions about the system's future behavior and potential breakdowns. There are many approaches to predictive maintenance and prognostics. This thesis focused on using data-driven methods, mainly machine learning and deep learning techniques and equipment monitoring data to develop predictive models that are able to learn degradation patterns of specific equipment and use it to predict the performance on new equipment. Chapter \ref{chapter:equipment_health_assessment_using_artificial_neural_networks} used NASA C-MAPSS dataset to showcase the use of artificial neural networks (both fully-connected neural networks and LSTM networks) to model equipment degradation using data from sensors that monitor and measure different physical variables. The developed models showed great performance at predicting equipment remaining useful life (\acrshort{rul}). Later, the chapter presented the potential adoption of this approach in oilfields on oil equipment.

Chapter \ref{chapter:bearings_faults_diagnostics_and_prognostics} focused on more specialized application of neural networks: monitoring bearings using vibration data for diagnostic and prognostic. The chapter used more sophisticated data processing techniques to extract suitable features from raw vibration data. Two different convolutional neural networks (\acrshort{cnn}) were  used, the first was used to diagnose bearings and classify fault type from vibration data, and the second used signal processing techniques and continuous wavelet transform to construct a dataset of scaleograms from vibration data which then was used as an input to the neural network to classify healthy and faulty bearings. The chapter also presented new feature extraction techniques from literature using trigonometric features and cumulative descriptors along with metrics used to quantify the suitability of these features to prognostics. Finally, a procedure was described to adopt solutions introduced in this chapter to Top Drives in oilfields.